\documentclass[accept,9pt,times,1p]{elsarticle_mod}

\usepackage{amsmath,amsfonts,bm}
\usepackage{amsthm}

\def\eqref#1{equation~\ref{#1}}

\def\beqref#1{(\ref{#1})}

\def\1{\bm{1}}

\DeclareMathAlphabet{\mathsfit}{\encodingdefault}{\sfdefault}{m}{sl}
\SetMathAlphabet{\mathsfit}{bold}{\encodingdefault}{\sfdefault}{bx}{n}

\newcommand{\pdata}{p_{\rm{data}}}

\newcommand{\E}{\mathbb{E}}
\newcommand{\Ls}{\mathcal{L}}
\newcommand{\Js}{\mathcal{J}}

\newcommand{\KL}{D_{\mathrm{KL}}\hspace{-1pt}}

\newcommand{\diag}{\mathrm{diag}}

\DeclareMathOperator*{\argmin}{arg\,min}

\usepackage{amssymb}

\usepackage[super]{nth}
\usepackage{threeparttable}
\usepackage{graphicx,subfig,array}
\usepackage{comment}
\usepackage{multirow}
\usepackage{url}            
\usepackage{hyperref}
\newtheorem{theorem}{Theorem}
\theoremstyle{definition}
\newtheorem{definition}{Definition}
\theoremstyle{lemma}
\newtheorem{lemma}{Lemma}

\newtheorem*{hypothesis*}{Main Hypothesis}

\newtheorem*{theorem*}{Property}
\newcommand{\bhline}[1]{\noalign{\hrule height #1}}

\newcounter{num}

\newcommand{\trace}{\mathrm{tr}}

\usepackage{colortbl}
\definecolor{tb}{rgb}{0.678,0.839,1.0}
\definecolor{tr}{rgb}{1.0,1.0,1.0}
\definecolor{crevise}{rgb}{0.0,0.0,0.0}

\usepackage{color,soul}

\usepackage{algorithm, algorithmic}

\begin{document}

\begin{frontmatter}



\title{Preventing Oversmoothing in VAE via Generalized Variance Parameterization}

\author{Yuhta Takida, Wei-Hsiang Liao, Chieh-Hsin Lai, Toshimitsu Uesaka, Shusuke Takahashi, Yuki Mitsufuji}
\address{Sony Group Corporation, 1-7-1 K\={o}nan, Minato-ku, Tokyo 108-0075, Japan}


\begin{abstract}
Variational autoencoders (VAEs) often suffer from posterior collapse, which is a phenomenon in which the learned latent space becomes uninformative. This is often related to the hyperparameter resembling the data variance. It can be shown that an inappropriate choice of this hyperparameter causes the oversmoothness in the linearly approximated case and can be empirically verified for the general cases.
{\color{crevise}Moreover, determining such appropriate choice becomes infeasible if the data variance is non-uniform or conditional.}
Therefore, we propose VAE extensions with generalized parameterizations of the data variance and incorporate maximum likelihood estimation into the objective function to adaptively regularize the decoder smoothness. The images generated from proposed VAE extensions show improved Fr\'{e}chet inception distance~(FID) on MNIST and CelebA datasets.
\end{abstract}



\end{frontmatter}






\vspace{0.0pt}
\section{Introduction}
\label{sec:motivation}
The variational autoencoder~(VAE) framework~\cite{kingma2013auto, higgins2017beta, zhao2019infovae} is a popular approach to achieve generative modeling in the field of machine learning.
In this framework, a model that approximates the true posterior of observation data, is learned by a joint training of encoder and decoder, which creates a stochastic mapping between the observation data and the learned deep latent space.
The latent space is assumed to follow a prior distribution. The generation of a new data sample can be done by sampling the latent space and passing the sample through the decoder. It is common to assume that both the prior on the latent space and the posterior of the observation data follow some parameteric probability distribution, such as the Gaussian distribution. In this case, the distribution of the output of decoder is characterized as $(\boldsymbol\mu_x, \mathbf{\Sigma}_x)$, where $\mathbf{\Sigma}_x$ is usually modeled as an isotropic matrix $\sigma_x^2\mathbf{I}$ with a scalar parameter $\sigma_x^2\geq0$.
Furthermore, in order to deal with the intractable log-likelihood of the true posterior, the evidence lower bound~(ELBO)~\cite{jordan1999introduction} is adopted as the objective function instead.
{\color{crevise}Recently there are variants of VAE such as NVAE~\cite{vahdat2020nvae} and Very deep VAE~\cite{child2021very}, in which their decoders are modeled with mixture of logistics (MoL)~\cite{salimans2017pixelcnn++}. However, in this work, we would like to focus on Gaussian-based decoder.}

While VAE-based generative models are usually considered to be more stable and easier to train than generative adversarial networks~\cite{goodfellow2014generative}, they often suffer from the problem of \textit{posterior collapse}~\cite{bowman2015generating,sonderby2016ladder,alemi2017fixing,xu2018spherical,he2019lagging,razavi2019preventing,ma2019mae}, in which the latent space has little information of the input data.
The phenomenon is generally referred as ``the posterior collapses to the prior in the latent space'' ~\cite{razavi2019preventing}.
Recently, several works have suggested that the variance parameter $\mathbf{\Sigma}_x$ is strongly related to posterior collapse.
For example, Lucas \textit{et al.} performed analysis on a linear VAE with $\mathbf{\Sigma}_x=\sigma_x^2\mathbf{I}$~\cite{lucas2019don}. It revealed that an inappropriate choice of $\sigma_x^2$ will introduce sub-optimal local optima and cause posterior collapse.
Moreover, they revealed that contrary to the popular belief~\cite{bowman2015generating,kingma2016improved,sonderby2016ladder}, these local optima are not introduced by replacing the log-likelihood with the ELBO, but by an excessively large $\sigma_x^2$.
On the other hand, it can be shown that fixing $\sigma_x^2$ to an excessively small value leads to under-regularization of the decoder, which can cause overfitting. 
In another work, Dai \textit{et al.} proposed a two-stage VAE and treated $\sigma_x^2$ as a training parameter~\cite{dai2019diagnosing}.  
Despite of these issues related to the setup of $\mathbf{\Sigma}_x$, many of the existing VAE implementations follow the $\mathbf{\Sigma}_x=\sigma_x^2\mathbf{I}$ setup and $\sigma_x^2=1.0$ is a fixed constant independent of data distribution. 

Besides the inappropriate choice of the variance parameter, posterior collapse can also induced by other causes. For example, small nonlinear perturbation introduced in the network architecture can also result into extra sub-optimal local minima~\cite{dai2020usual}. 
However, in this work we will keep our focus on the variance parameter.
In addition, although determining the appropriate value of the variance parameter is crucial, since the true data variance is possibly conditional or not spatially uniform within the dataset, modeling the data variance with only a scalar value is likely to be sub-optimal.
This motivates us to derive more generalized parameterizations of the data variance and find the mechanisms that determine the appropriate values for these parameterizations.   
%

%

In this paper, we would like to suggest that $\mathbf{\Sigma}_x$ affects the strength of regulation over the gradient magnitude of the decoder. We call the expected gradient magnitude as \textit{smoothness}. The smaller the gradient magnitude, the smoother the model. In particular, we focus on the \textit{local smoothness} of the model, which is the smoothness evaluated within the neighborhood of the encoded observation data in the latent space.
As the first step, we depict the relation between $\mathbf{\Sigma}_x$ and the oversmoothing phenomenon with the following hypothesis:
\begin{hypothesis*}
  \label{th:hypothesis}
  The value of $\mathbf{\Sigma}_x$ affects the regularization strength of the smoothness of the decoder. Consequentially, $\mathbf{\Sigma}_x$ with excessively large values causes oversmoothness, which results in posterior collapse.
\end{hypothesis*}

Following the hypothesis, we will start with analyzing how $\mathbf{\Sigma}_x$ regularizes the local smoothness of the stochastic decoder. Then, we will propose several parameterizations of $\mathbf{\Sigma}_x$ and their corresponding objective functions which determine $\mathbf{\Sigma}_x$ via maximum likelihood estimation (MLE) to achieve adaptive regularization strength control. In other words, we extended the conventional VAE by introducing alternative variance parameterizations. These parameterizations are capable of model non-uniform or conditional data variances correctly and thus be able to prevent posterior collapse induced by oversmoothness.


Our main contributions are listed as follows:
\begin{enumerate}
  \setlength{\parskip}{0cm}
  \setlength{\itemsep}{0cm}
  \item
  We show that our main hypothesis holds for linear approximated ELBO and empirically holds in the general case in Section~\ref{sec:major_problem}. This also verifies that the variance parameter should be estimated from data observation.
  \item
  We propose VAE extensions with alternative data variance parameterizations to handle non-uniform or conditional data variance. The proposed approach can adaptively regularizes the smoothness of the decoder by MLE of the variance parameter $\mathbf{\Sigma}_x$. The correctness of estimated variances are empirically verified in Section~\ref{sec:variance_models}.  
  \item
  The proposed approach not only prevents the posterior collapse induced by oversmoothing, but also improves the quality of generation, as shown in Section~\ref{sec:experiment_comparison}.
  
 
\end{enumerate}

The organization of this paper is as follows. In Section~\ref{sec:preliminaries}, we begin with introducing fundamental mathematical definitions. In addition, for the sake of clarity,  we introduce a customized definition for the posterior collapse. In Section~\ref{sec:major_problem}, the theoretical analysis and empirical support of the main hypothesis are given. In Section~\ref{sec:variance_models}, we propose several parameterizations on posterior variance and derive corresponding objective functions. In Section~\ref{sec:experiment_comparison}, we evaluate the quality of generation on the MNIST and CelebA datasets. 

Throughout this paper, we use $a$, $\mathbf{a}$ and $\mathbf{A}$ for a scalar, a column vector and a matrix, respectively.
$\ln$ and $\log$ denote the natural logarithm and common logarithm. 

\vspace{0.0pt}
\section{{\color{crevise}Related works}}
\vspace{0.0pt}
\label{sec:related_works}
{\color{crevise}In this section, we will introduce several related works. Some of them will be included in Section~\ref{sec:experiment_comparison}.}

To the best of our knowledge, \cite{lucas2019don} was among the first to suggest that posterior collapse may be caused by a sub-optimal variance parameter.
In the past, one of the common approaches for dealing with posterior collapse was to anneal the weight of the KL term in the ELBO. The first such attempt was KL annealing~\cite{bowman2015generating}.
{\color{crevise}In \cite{bowman2015generating}, a weighting coefficient on the KL term was introduced into the cost function.} The weighting scheduling is determined in advance, e.g., increases monotonically~\cite{bowman2015generating,sonderby2016ladder} or changes cyclically~\cite{fu2019cyclical} as the training progresses.

In \cite{higgins2017beta}, the weighting coefficient is interpreted as a hyperparameter that controls the information capacity of the latent space and a value larger than 1 is recommended.
Therefore it enforces a stronger smoothness in exchange for better latent space disentanglement.
{\color{crevise}To be mentioned, our proposed method differs from \cite{higgins2017beta} in several ways: (i) The coefficient in proposed method is characterized as a matrix $\bm{\Sigma}_x$; (ii) $\bm{\Sigma}_x$ updates every minibatch; and (iii) our work seeks a dynamic balance of the regularization strength between the decoder smoothness and the latent space disentanglement.}

{\color{crevise}There are other works that also apply the dynamic weighting control.} ControlVAE~\cite{shao2020controllable} incorporated the control theory and applied PI/PID control to the weight of the KL term. Although it is possible to reflect the status of the optimization dynamically, ControlVAE needs extra hyperparameters to be tuned in advance. 
In \cite{dai2019diagnosing} a further step is taken to treat $\sigma_{{x}}^2$ as an usual trainable parameter. 
In the aspect of weight control, our proposed method can be interpreted as an automatic KL annealing that estimates $\sigma_{{x}}^2$ through MLE without the need of tuning hyperparameters, which makes it differ from these methods.

{\color{crevise}\cite{ghosh2019from} is another approach that attempts to regularize the decoder.} It treats the stochastic autoencoder with the reparameterization trick as a noise injection process and proposed replacing such a mechanism with an explicit regularized autoencoder (RAE). Moreover, its decoder is regularized in multiple ways: $L_2$ regularization, a gradient penalty~\cite{gulrajani2017improved} and spectral normalization~\cite{miyato2018spectral}. {\color{crevise}As what we will discuss in Section~\ref{sec:local_optim_ELBO}}, if $\sigma_{{z}}^2$ is sufficiently small, the ELBO can also be approximately represented as a sum of three losses~\beqref{eq:ELBO_linear_approx}, which correspond to the terms included in the basic RAE objective function. The approximated objective function~\beqref{eq:ELBO_linear_approx} is equal to RAE with a properly tuned gradient penalty (RAE-GP). 
In the same work, a state-of-the-art-model called WAE-MMD~\cite{tolstikhin2018wasserstein} is included as a comparison target. 
While RAE regularizes the model explicitly, WAE-MMD relies on its network architecture and generalization techniques to regularize the model implicitly.

Compared to these two methods, the proposed method do not explicitly add the regularizing terms, but it adapts the variance parameters to affect the strength of regularization. Again, both WAE and RAE include hyperparameters in their objective functions. Another advantage of the proposed method is the capability of imposing different regularization weights for different dimensions of the latent space according to the property of the input data.

\vspace{0.0pt}
\section{Background}\label{sec:preliminaries}
We begin with the standard formulation of VAE, also known as the \textit{Gaussian VAE}, which is the foundation this work. We believe considering Gaussian VAE is sufficient for typical cases due to its expressive power for general cases has been verified in \cite{dai2019diagnosing}. In addition, it also found that the Gaussian setup does not pose negative effect to the optimization process. 
In the second subsection, we propose a customized definition for the posterior collapse, called \textit{MI-induced posterior collapse}. This is induced by the loss of mutual information (MI) between the latent space and the data space, which can be caused by an over-smoothed decoder.

\vspace{0.0pt}
\subsection{Gaussian VAE}\label{sec:standard_VAE}
\vspace{0.0pt}

Consider a data space $\mathcal{X}\subset\mathbb{R}^{d_x}$ and a sample set $\{\mathbf{x}_i\}_{i=1}^{N}\subset\mathcal{X}$, where $\mathbf{x}_i\sim p_{\text{data}}(\mathbf{x})$.
The empirical distribution $\tilde{p}_{\text{data}}(\mathbf{x})$ on $\mathcal{X}$ can be evaluated by $\tilde{p}_{\text{data}}(\mathbf{x})=\frac{1}{N}\sum_{n=1}^{N}\delta(\mathbf{x}-\mathbf{x}_n)$, where $\delta(\cdot)$ denotes the Dirac delta function.
In the standard VAE framework, a latent space $\mathcal{Z}\subset\mathbb{R}^{d_z}$ is learned via the joint training of a pair of stochastic encoder and decoder, denoted by  $q_{\bm{\phi}}(\mathbf{z}|\mathbf{x})$ and $p_{\bm{\theta}}(\mathbf{x}|\mathbf{z})$, respectively. The generation of new data samples $\mathbf{x}^\prime\in\mathcal{X}$ can be done through decoding the sampled latent variables $\mathbf{z}\in\mathcal{Z}$. Trainable parameters of the two neural networks are denoted as $\bm{\phi}$ and $\bm{\theta}$. The decoder generates data samples by $p_{\bm{\theta}}(\mathbf{x}):=\E_{p(\mathbf{z})}[p_{\bm{\theta}}(\mathbf{x}|\mathbf{z})]$, where $p(\mathbf{z})$ is the prior distribution on $\mathcal{Z}$. The encoder and decoder are trained by minimizing the following objective function:
\begin{align}
  \Ls =& -\E_{p_{\text{data}}(\mathbf{x})}\left[\ln{p_{\bm{\theta}}(x)}\right] + \E_{p_{\text{data}}(\mathbf{x})}\left[\ln{p_{\text{data}}(\mathbf{x})}\right]
  + \E_{p_{\text{data}}(\mathbf{x})}\KL\left(q_{\bm{\phi}}(\mathbf{z}|\mathbf{x})\parallel  p_{\bm{\theta}}(\mathbf{z}|\mathbf{x})\right) \notag\\ 
  =& \KL\left(p_{\text{data}}(\mathbf{x})\parallel p_{\bm{\theta}}(\mathbf{x})\right) 
  + \E_{p_{\text{data}}(\mathbf{x})}\KL\left(q_{\bm{\phi}}(\mathbf{z}|\mathbf{x})\parallel p_{\bm{\theta}}(\mathbf{z}|\mathbf{x})\right).
  \label{eq:objective_vae}
\end{align}
This objective function was derived in \cite{zhao2019infovae}, which represents all the terms in Kullback--Leibler divergences, and is equivalent to ELBO maximization up to an additive constant.

In the context of the Gaussian VAE, the encoder and decoder are assumed to satisfy
\begin{align}
  q_{\bm{\phi}}(\mathbf{z}|\mathbf{x}) &= \mathcal{N}(\mathbf{z}|\bm{\mu}_{\bm{\phi}}(\mathbf{x}),\Sigma_\phi(\mathbf{x}))\quad \mathrm{and}
  \quad p_{\bm{\theta}}(\mathbf{x}|\mathbf{z}) = \mathcal{N}(\mathbf{x}|\bm{\mu}_{\bm{\theta}}(\mathbf{z}),\mathbf{\Sigma}_x),
  \label{eq:enc_and_dec_vae}
\end{align}
where $\Sigma_\phi(\mathbf{x})=\diag(\bm{\sigma}_{\bm{\phi}}^2(\mathbf{x}))$ and $\mathbf{\Sigma}_x=\sigma_x^2\mathbf{I}$.
Since the prior $p(\mathbf{z})$ is assumed to be a Gaussian distribution as $p(\mathbf{z})=\mathcal{N}(\mathbf{z}|\mathbf{0},\mathbf{I})$, substituting Eq.~\beqref{eq:enc_and_dec_vae} into Eq.~\beqref{eq:objective_vae} while omitting terms independent of $\bm{\theta}$ and $\bm{\phi}$ leads to the following objective:
\begin{align}
  \tilde{\Js}_{\sigma_x^2}(\bm{\theta},\bm{\phi})
  =& \E_{\tilde{p}_{\text{data}}(\mathbf{x})}\left[\frac{1}{2\sigma_x^2}\E_{q_{\bm{\phi}}(\mathbf{z}|\mathbf{x})}[\left\|\mathbf{x}-\bm{\mu}_{\bm{\theta}}(\mathbf{z})\right\|_2^2] 
  +\KL(q_{\bm{\phi}}(\mathbf{z}|\mathbf{x})\parallel p(\mathbf{z}))\right],
  \label{eq:loss_gaussian_vae}
\end{align}
which is the sum of the expected values of the reconstruction loss and a regularization term. In the case of a Gaussian prior and posterior, the regularization term equals to $\frac{1}{2}\sum_{i=1}^{d_z}(\sigma_{\bm{\phi},i}^2(\mathbf{x})+\mu_{\bm{\phi},i}(\mathbf{x})^2-\log\sigma_{\bm{\phi},i}^2(\mathbf{x})-1)$.
\vspace{0.0pt}%
\subsection{Posterior collapse}
\label{sec:posterior_collapse}
\vspace{0.0pt}%

In this work, we focus on a common type of posterior collapse where the MI between input data and reconstructed data through the encoder-decoder path is reduced to such an extent that the decoder can no longer generate the data distribution using the latent information.
Therefore, we suggest the following definition of \textbf{Mutual information (MI)-induced posterior collapse}.
\begin{definition}
  \textbf{MI-induced posterior collapse} is defined as the MI $\mathcal{I}(\mathbf{x};\mathbf{x}^\prime)$ becoming nearly zero, where $\mathbf{x}^\prime:=\bm{\mu}_{\bm{\theta}}(\mathbf{z})$ with $\mathbf{z}\sim q_{\bm{\phi}}(\mathbf{z}|\mathbf{x})$.
\end{definition}
In many existing works~\cite{bowman2015generating,sonderby2016ladder,alemi2017fixing,he2019lagging,razavi2019preventing}, \textit{posterior collapse} is often represented as $\E_{{p}_{\text{data}}(\mathbf{x})}\KL(q_{\bm{\phi}}(\mathbf{z}|\mathbf{x})\parallel p(\mathbf{z}))\to 0$, which is also referred to as \textit{KL collapse}~\cite{xu2018spherical}.
{\color{crevise}Here, we propose the following theorem to depict the relation between MI-induced posterior collapse and KL collapse.} It shows that MI-induced posterior collapse is a superset of KL collapse, the proof can be found in \ref{sec:collapses}.
\begin{theorem}
  \label{th:collapses}
  $\mathcal{I}(\mathbf{x};\mathbf{x}^\prime)\to0$ as $\E_{{p}_{\text{data}}(\mathbf{x})}\KL(q_{\bm{\phi}}(\mathbf{z}|\mathbf{x})\parallel p(\mathbf{z}))\to0$ holds for any $p_{\bm{\theta}}(\mathbf{x}|\mathbf{z})$.
\end{theorem}
Furthermore, in \ref{sec:fix_varz}, we demonstrate that MI-induced posterior collapse can happen even if the KL divergence is nonzero when the posterior variance is fixed in  $\mathcal{Z}$.

\vspace{0.0pt}%
\section{Variance parameters and the local smoothness}
\label{sec:major_problem}
\vspace{0.0pt}%

In this section, we provide mathematical and empirical supports for the main hypothesis.
Throughout this section, we use the following parameterization for simplicity: $q_{\bm{\phi},\sigma_z^2}(\mathbf{z}|\mathbf{x})=\mathcal{N}(\mathbf{z}|\bm{\mu}_{\bm{\phi}}(\mathbf{x}),\sigma_z^2\mathbf{I})$ and $p_{\bm{\theta}}(\mathbf{x}|\mathbf{z}) = \mathcal{N}(\mathbf{x}|\bm{\mu}_{\bm{\theta}}(\mathbf{z}),\sigma_x^2\mathbf{I})$,
where both variances are parameterized as isotropic matrices unlike the conventional VAE. A similar analysis on the conventional VAE can be found in \ref{sec:linear_approx_objective}.
It begins with showing that the choice of $\sigma_x^2$ affects the convergence point of $\sigma_z^2$, which is the variance parameter of the latent space. Then, we show that $\sigma_z^2$ acts as the weight of the gradient penalty, which is implicitly included in Eq.~\beqref{eq:loss_gaussian_vae}.
This supports the main hypothesis that the over-regulation imposed by a large $\sigma_x^2$ via $\sigma_z^2$ causes the oversmoothness of the decoder and leads to \textbf{MI-induced posterior collapse}. It is also empirically supported by observing the tendencies of the convergence point of $\sigma_z^2$, the smoothness and the MI $\mathcal{I}(\mathbf{x},\mathbf{x}^\prime)$.
Ultimately, these items of evidence motivated us to develop a method that adapts the variance parameter to prevent oversmoothing the decoder.

\vspace{0.0pt}%
\subsection{Regularization effect of variance parameters in linear approximated ELBO}
\vspace{0.0pt}%
\label{sec:local_optim_ELBO}

{\color{crevise}The effect of $\sigma_x^2$ on the convergence point of the variance parameter $\sigma_z^2$ can be explained by observing two extreme cases, $\sigma_x^2\to0+$ and $\sigma_x^2\to\infty$.
First, we propose the following theorem whose proof can be found in} \ref{sec:convergence_var}.
\begin{theorem}
  \label{th:convergence_var}
  Assuming that $\pdata(\mathbf{x})$ has finite covariance and $\bm{\mu}_{\bm{\phi}}$ is Lipschitz continuous.
  Consider the global optimum of $\Js_{\sigma_x^2}(\bm{\theta},\bm{\phi}, \sigma_z^2)$ w.r.t. a given $\sigma_x^2$. If $\sigma_x^2\to0$, then $\sigma_z^2\to0$.
\end{theorem}

According to Theorem~\ref{th:convergence_var}, when $\sigma_x^2$ approaches $0$ as the training progresses, $\sigma_z^2$ will also approaches $0$, which illustrates the first case.
In the second case, $\tilde{\mathcal{J}}_{\sigma_x^2}$ reduces to $\E_{\tilde{p}_{\text{data}}(\mathbf{x})}\KL(q_{\bm{\phi},\sigma_z^2}(\mathbf{z}|\mathbf{x})\parallel p(\mathbf{z}))$, and $\sigma_z^2$ becomes $1$ at the minimum point, from $\KL(q_{\bm{\phi},\sigma_z^2}(\mathbf{z}|\mathbf{x})\parallel p(\mathbf{z}))=\frac{d_z}{2}(\sigma_z^2-\log\sigma_z^2-1)+\|\bm{\mu}_{\bm{\phi}}(\mathbf{x})\|_2^2$.
This shows that a small $\sigma_x^2$ makes $\sigma_z^2$ converge near $0$, while a large $\sigma_x^2$ makes $\sigma_z^2$ converge near $1$.

If $\sigma_z^2$ is sufficiently small, as the training progresses to a certain extent, the perturbed decoding process $\bm{\mu}_{\bm{\theta}}(\mathbf{z}+\bm{\epsilon}_z)$ around $\mathbf{z}=\bm{\mu}_{\bm{\phi}}(\mathbf{x})$ with $\bm{\epsilon}_z\sim\mathcal{N}(\bm{\epsilon}_z|\mathbf{0},\sigma_z^2\mathbf{I})$ can be approximated as a linear function.
Therefore, the ELBO can be approximated as follows by using the linear approximation of $\bm{\mu}_{\bm{\theta}}(\cdot)$ and omitting terms independent of $\bm{\theta}$ and $\bm{\phi}$:
\begin{align}
  \tilde{\Js}_{\sigma_x^2}(\bm{\theta},\bm{\phi},\sigma_z^2)
  &\approx \frac{1}{2\sigma_x^2}\E_{\tilde{p}_{\text{data}}(\mathbf{x})}\left[
  \left\|\mathbf{x}-\bm{\mu}_{\bm{\theta}}(\bm{\mu}_{\bm{\phi}}(\mathbf{x}))\right\|_2^2\right.\notag\\
  &\qquad+\left.
  \sigma_z^2\left\|\nabla\bm{\mu}_{\bm{\theta}}(\bm{\mu}_{\bm{\phi}}(\mathbf{x}))\right\|_F^2
  +2\sigma_x^2\left\|\bm{\mu}_{\bm{\phi}}(\mathbf{x})\right\|_2^2\right].
  \label{eq:ELBO_linear_approx}
\end{align}
In the approximation above, $\|\cdot\|_F$ is the Frobenius norm and $\sigma_z^2$ is treated as a function parameter. Its derivation can be found in \ref{sec:linear_approx_objective}.
The objective function approximated by Eq.~\beqref{eq:ELBO_linear_approx} consists of three terms: a reconstruction error term, a gradient penalty term and a $L_2$ regularization term.
As one can see from Eq.~\beqref{eq:ELBO_linear_approx}, $\sigma_z^2$ regularizes the smoothness of the decoder by penalizing its gradient norm during the training.
Although the linear approximation is derived for the simplified VAE parameterization, the linear approximation of the ELBO for the standard VAE parameterization~\beqref{eq:enc_and_dec_vae} is provided in \ref{sec:linear_approx_objective}, where the second term in Eq.~\beqref{eq:ELBO_linear_approx} becomes a weighted gradient penalty.

Summarizing the observations above shows that $\sigma_x^2$ affects the decoder smoothness via $\sigma_z^2$, while $\sigma_z^2$ directly regularizes the smoothness. If $\sigma_x^2$ is excessively large, it will cause over-regularization of the decoder and suppress $\mathcal{I}(\mathbf{z},\mathbf{x}^\prime)(\geq\mathcal{I}(\mathbf{x},\mathbf{x}^\prime))$, which finally leads to MI-induced posterior collapse.
This suggests that $\sigma_x^2$ and $\sigma_z^2$ should be adapted appropriately. In  \ref{sec:fix_varz}, we further show that MI-induced posterior collapse can be triggered by manipulating $\sigma_z^2$ directly.

\vspace{0.0pt}%
\subsection{Empirical study on smoothness of decoder in the general case}
\vspace{0.0pt}%
\label{sec:exp_analysis_VAE}

Section~\ref{sec:local_optim_ELBO} shows the impact of $\sigma_x^2$ on the regularization of the decoder smoothness through the linear approximated objective function. To support the main hypothesis in the general case, an experiment on the MNIST dataset~\cite{lecun1998gradient} is conducted. Several criteria are accessed to provide evidence for the regularization effect of $\sigma_x^2$ on the decoder smoothness and its consequential effect on MI $\mathcal{I}(\mathbf{x},\mathbf{x}^\prime)$.
To confirm that $\sigma_x^2$ affects the smoothness via $\sigma_z^2$, we conduct the experiment for two cases:  \textbf{stochastic encoding} and \textbf{deterministic encoding}.
While the stochastic encoder $q_{\bm{\phi},\sigma_z^2}(\mathbf{z}|\mathbf{x})$ is used in the former case, a VAE equipped with a deterministic encoder, i.e., $\sigma_z^2$ is fixed to zero during the training, is investigated in the latter case.
Observing the difference between the two cases empirically supports Section~\ref{sec:local_optim_ELBO}.
To investigate the relation between $\sigma_x^2$ and the smoothness of the decoder clearly, common generalization techniques such as batch normalization~\cite{ioffe2015batch,santurkar2018does} and weight decay are excluded in the training.

\paragraph{Criteria}
In order to observe the smoothness of the decoder, first consider a decoding process with perturbation involved $\bm{\mu}_{\bm{\theta}}(\bm{\mu}_{\bm{\phi}}(\mathbf{x})+\bm{\epsilon}_z)$, where $\bm{\epsilon}_z\sim\mathcal{N}(\bm{\epsilon}_z|\mathbf{0},s_z^2\mathbf{I})$ is a zero-mean Gaussian distribution with variance $s_z^2$.
Assuming ${\bm{\epsilon}_z}$ and ${\bm{\epsilon}_z^\prime}$ are i.i.d. random variables. We may define the \textbf{expected gap} $\Delta^2(s_z^2)$ between the decoded samples as 
\begin{align}
  \Delta^2(s_z^2):=\E_{p_{\text{data}}(\mathbf{x})\mathcal{N}(\bm{\epsilon}_z|\mathbf{0},s_z^2\mathbf{I})\mathcal{N}(\bm{\epsilon}_z^\prime|\mathbf{0},s_z^2\mathbf{I})}[\Delta^2(\mathbf{x},\bm{\epsilon}_z,\bm{\epsilon}_z^\prime)]
\end{align}
  with $\Delta^2(\mathbf{x},\bm{\epsilon}_z,\bm{\epsilon}_z^\prime):=\|\bm{\mu}_{\bm{\theta}}(\bm{\mu}_{\bm{\phi}}(\mathbf{x})+{\bm{\epsilon}_z})-\bm{\mu}_{\bm{\theta}}(\bm{\mu}_{\bm{\phi}}(\mathbf{x})+{\bm{\epsilon}_z^\prime})\|_2^2$.
As $s_z^2$ decreases, the ratio $\Delta^2(s_z^2)/(2s_z^2)$ converges and becomes an indicator of $\E_{\tilde{p}_{\text{data}}}(\mathbf{x})[\|\nabla\bm{\mu}_{\bm{\theta}}(\bm{\mu}_{\bm{\phi}}(\mathbf{x}))\|_F^2]$, which is regularized by $\sigma_z^2$ as shown in Eq.~\beqref{eq:ELBO_linear_approx}. We define this term as the the \textbf{expected local smoothness} (ELS):
\begin{align}
 \E_{\tilde{p}_{\text{data}}(\mathbf{x})}[\|\nabla\bm{\mu}_{\bm{\theta}}(\bm{\mu}_{\bm{\phi}}(\mathbf{x}))\|_F^2].
 \label{eq:def_els}
\end{align}
ELS is a lower bound of the Lipschitz constant of the decoder, and is an indicator of the smoothness of decoder, further detail can be found in \ref{sec:smoothness_decoder}.

Finally, we include the following criteria along with ELS to observe the impact of $\sigma_x^2$: the reconstruction error (MSE), the KL divergence $\E_{\tilde{p}_{\text{data}}(\mathbf{x})}\KL(q_{\bm{\phi}}(\mathbf{z}|\mathbf{x})\parallel p(\mathbf{z}))$, the convergence value of $\sigma_z^2$ and the MI between the latent variable and the decoder output $\mathcal{I}(\mathbf{x}^\prime;\mathbf{z})$. 
Since the direct evaluation of MI is intractable, we estimate $\mathcal{I}(\mathbf{x}^\prime;\mathbf{z})$ by Monte Carlo estimation. As a reference, this is also an upper bound of $\mathcal{I}(\mathbf{x}^\prime;\mathbf{x})$.

\paragraph{Results}
Table~\ref{tb:result1} summarizes the results for different $\sigma_x^2$.
In the \textit{stochastic encoding} case, a large $\sigma_x^2$ consistently leads to a larger $\sigma_z^2$. This results in a smaller expected gap, a smaller ELS and a lower upper bound of MI. This supports the main hypothesis that a larger $\sigma_x^2$ makes the decoder smoother.
In the case of $\sigma_x^2=1.0$, which is a exceedingly large value for the MNIST dataset, all the criteria except MSE become nearly zero. This means MI-induced posterior collapse and KL collapse both occur due to the over-regularization of the smoothness of the latent space.
On the other hand, in the \textit{deterministic encoding} case, where $\sigma_z^2$ is fixed to zero, the ELS keep increasing with $\sigma_x^2$. 
This is because $\sigma_x^2$ does not directly regularize the decoder, as shown in Eq.~\beqref{eq:ELBO_linear_approx}.
As a result, the MI upper bound does not shrink to zero even if $\sigma_x^2$ becomes exceedingly large unlike the stochastic encoding case, in which MI-induced posterior collapse occurred.

The difference between the results of the two cases clearly suggests that a large $\sigma_x^2$ triggers the oversmoothness via $\sigma_z^2$, which is consistent with the discussion in Section~\ref{sec:local_optim_ELBO}. These results provide empirical support of the main hypothesis as well as the discussion in Section~\ref{sec:local_optim_ELBO}. Further details and examples of images are shown in \ref{sec:detail_exp_analysis}.
\begin{table}[t]
  \caption{Evaluation of various criteria for different $\sigma_x^2$: the expected value of $\|\mathbf{x}^\prime-\mathbf{x}\|_2^2$ (MSE), KL divergence, the converged value of $\sigma_z^2$, the upper bound of the MI $\mathcal{I}(\mathbf{x}^\prime;\mathbf{z})$, the expected gap (perturbation variance $s_z^2$ are set to $10^{-2}$ and $10^{-3}$) and expected local smoothness (ELS).}
  \small
  \centering
  \resizebox{\columnwidth}{!}{
  \begin{tabular}{r|rrrrrrr|rrrrr}
    \bhline{0.8pt}
    \multirow{3}{*}{$\log\sigma_x^2$} & \multicolumn{7}{c|}{Stochastic encoding} & \multicolumn{5}{c}{Deterministic encoding} \\ \cline{2-8}\cline{9-13}
    & \multirow{2}{*}{MSE} & \multirow{2}{*}{KL} & \multirow{2}{*}{$\sigma_z^2$} & \multirow{2}{*}{MI} & \multicolumn{2}{c}{Expected gap} & \multirow{2}{*}{ELS} & \multirow{2}{*}{MSE} & \multirow{2}{*}{MI} & \multicolumn{2}{c}{Expected gap} & \multirow{2}{*}{ELS}\\ \cline{6-7}\cline{11-12}
    &                      &                     &                               &                     & $10^{-2}$ & $10^{-3}$            &                      &                      &                     & $10^{-2}$ & $10^{-3}$            & \\
    \bhline{0.8pt}
    $0.0$  & \textbf{52.74} &  \textbf{0.00} &  \textbf{1.00} & \textbf{0.03}    & 6.31e-5 & 6.35e-6  & \textbf{3.97e-4} & 5.95 & 12.5 & 74.6 & 25.3 & 7.43e+2\\
    $-0.1$ & 18.03          &   9.39         &  9.56e-2       & 9.7              & 1.05    & 0.108    & 6.76          & 5.69 & 14.7    & 69.1 & 22.5 & 6.82e+2\\
    $-0.2$ & 15.15          &  10.93         &  6.48e-2       & 12.5             & 1.30    & 0.135    & 8.34          & 5.38 & 17.9    & 63.4 & 20.7 & 6.37e+2\\
    $-0.3$ & 13.08          &  12.54         &  4.36e-2       & 16.0             & 1.51    & 0.157    & 9.72          & 5.37 & 21.4    & 58.1 & 17.9 & 5.78e+2\\
    $-0.4$ & 11.38          &  14.13         &  3.01e-2       & 20.6             & 1.77    & 0.184    & 1.14e+1       & 5.31 & 25.8    & 58.2 & 15.5 & 5.40e+2\\
    $-0.5$ & 10.18          &  15.30         &  2.14e-2       & 26.3             & 1.99    & 0.208    & 1.28e+1       & 5.26 & 30.6    & 53.1 & 12.9 & 4.74e+2\\
    $-0.6$ &  9.16          &  16.72         &  1.55e-2       & 33.2             & 2.16    & 0.227    & 1.40e+1       & 5.14 & 38.9    & 48.9 & 11.8 & 4.42e+2\\
    $-0.7$ &  8.25          &  18.05         &  1.11e-2       & 42.3             & 2.31    & 0.244    & 1.50e+1       & 5.17 & 46.1    & 45.5 & 10.1 & 3.98e+2\\
    $-0.8$ &  7.72          &  19.27         &  8.21e-3       & 52.9             & 2.40    & 0.254    & 1.56e+1       & 5.06 & 58.0    & 43.2 & 9.15    & 3.71e+2\\
    $-0.9$ &  7.13          &  20.55         &  5.97e-3       & 64.9             & 2.43    & 0.257    & 1.58e+1       & 4.98 & 71.9    & 39.2 & 7.83    & 3.29e+2\\
    $-1.0$ &  6.70          &  21.75         &  4.45e-3       & 82.3             & 2.57    & 0.272    & 1.67e+1       & 5.01 & 89.1    & 35.3 & 6.61    & 2.89e+2\\
    \bhline{0.8pt}
  \end{tabular}
  }
  \label{tb:result1}
\end{table}
\vspace{0.0pt}%
\subsection{Difficulty of determining variance parameter}
\vspace{0.0pt}%
\label{sec:prevent_collapse}

According to the previous discussion, an excessively large $\sigma_x^2$ causes oversmoothness.
Therefore, it is intuitive to ask if fixing $\sigma_x^2$ to a sufficiently small value will solve the problem.
Here, we may invoke Theorem~\ref{th:convergence_var} to answer this question.
In Theorem~\ref{th:convergence_var}, $\Js_{\sigma_x^2}(\bm{\theta},\bm{\phi}, \sigma_z^2)$ is optimized on the basis of the true data distribution $p_{\text{data}}(\mathbf{x})$ instead of the empirical data distribution $\tilde{p}_{\text{data}}(\mathbf{x})$. According to the theorem, $\sigma_z^2$ converges to zero as $\sigma_x^2$ approaches zero, which leads to zero gradient penalty for the decoder during the VAE training process.
In practice, we have no access to $p_{\text{data}}(\mathbf{x})$, but only have access to  $\tilde{p}_{\text{data}}(\mathbf{x})$.
However, Theorem~\ref{th:convergence_var} remains true even when $p_{\text{data}}(\mathbf{x})$ is replaced with $\tilde{p}_{\text{data}}(\mathbf{x})$.
In this case, if $\sigma_x^2$ is chosen to be small, the optimization process of $\tilde{\Js}_{\sigma_x^2}$ will fit $p_{\bm{\theta},\sigma_x^2}(\mathbf{x})$ to the empirical distribution $\tilde{p}_{\text{data}}(\mathbf{x})$, which may results in overfitting.
As shown above, it is nontrivial to choose an appropriate variance parameter that avoids both oversmoothness and overfitting. 

{\color{crevise}In this section, we've shown that suboptimal decoder variance can cause MI-induced posterior collapse. This is due to the variance parameter cannot resemble the true data variance correctly, and this often makes the decoder oversmoothed. On the other hand, even if we can determine an optimal value, the oversmoothness may still happen for some data because the simple parameterization cannot capture the data variance in case it is non-uniform or conditional. To completely solve this problem, we propose to extend VAE with generalized parameterizations of the decoder variance. Moreover, it is likely that the variance parameter should be adapted depending on the status of training, which will also be described in the next section.}


\vspace{0.0pt}%
\section{Proposed extensions of VAE}
\label{sec:variance_models}
\vspace{0.0pt}%

{\color{crevise}In the standard VAE given by Eq.~\beqref{eq:enc_and_dec_vae}, the variance of the decoded distribution on $\mathcal{X}$ is modeled as an identity matrix $\sigma_x^2\mathbf{I}$. In this case, $\sigma_x^2$ is a scalar value which balances the weight between the reconstruction term and rest of regularization terms as in Eq.~\beqref{eq:loss_gaussian_vae}. However, as described in the previous section, it is desirable to extend the parameterization to better represent the data variance in general. However, the training of the extended model is often unstable due to the extra variance parameters. Therefore, we propose to incorporate MLE to estimate these variance parameters, which can be regarded as an adaptive weighting scheme for the ELBO terms. The correctness of estimated data variance is examined by using a modified MNIST dataset with injected artificial Gaussian noise.}

\vspace{0.0pt}%
\subsection{Generalized parameterization of the decoder variance}
\vspace{0.0pt}%
\label{sec:proposed_objectives}

In case that the data variance is spatially dependent or conditional, the most generalized parameterization is to extend the variance parameter from $\sigma_x^2\mathbf{I}$ to a positive semi-definite matrix $\bm{\Sigma}_x$. The ELBO for such generalized parameterization is 
\begin{subequations}
\begin{align}
  \tilde{\Js}(\bm{\theta},\bm{\phi},\bm{\Sigma}_x)
  &= \tilde{\Js}_{\text{rec}}(\bm{\theta},\bm{\phi},\bm{\Sigma}_x)
  + \E_{\tilde{p}_{\text{data}}(\mathbf{x})}\KL\left(q(\mathbf{z}|\mathbf{x})\parallel p(\mathbf{z})\right)\label{eq:objective_general_model}\\
  \tilde{\Js}_{\text{rec}}(\bm{\theta},\bm{\phi},\bm{\Sigma}_x)
  &=\frac{1}{2}\E_{\tilde{p}_{\text{data}}(\mathbf{x})q_{\bm{\phi}}(\mathbf{z}|\mathbf{x})}
  \left[\trace\left(\bm{\Sigma}_x^{-1}(\mathbf{x}-\bm{\mu}_{\bm{\theta}}(\mathbf{z}))(\mathbf{x}-\bm{\mu}_{\bm{\theta}}(\mathbf{z}))^\top\right)
  +\ln|\bm{\Sigma}_x|\right],
  \label{eq:objective_reconstruction_general_model}
\end{align}
\end{subequations}
where $\trace(\cdot)$ denotes the trace of a matrix.

\begin{figure}[tb]
  \centering
  \subfloat[Iso-I]{\includegraphics[height=0.25\textwidth]{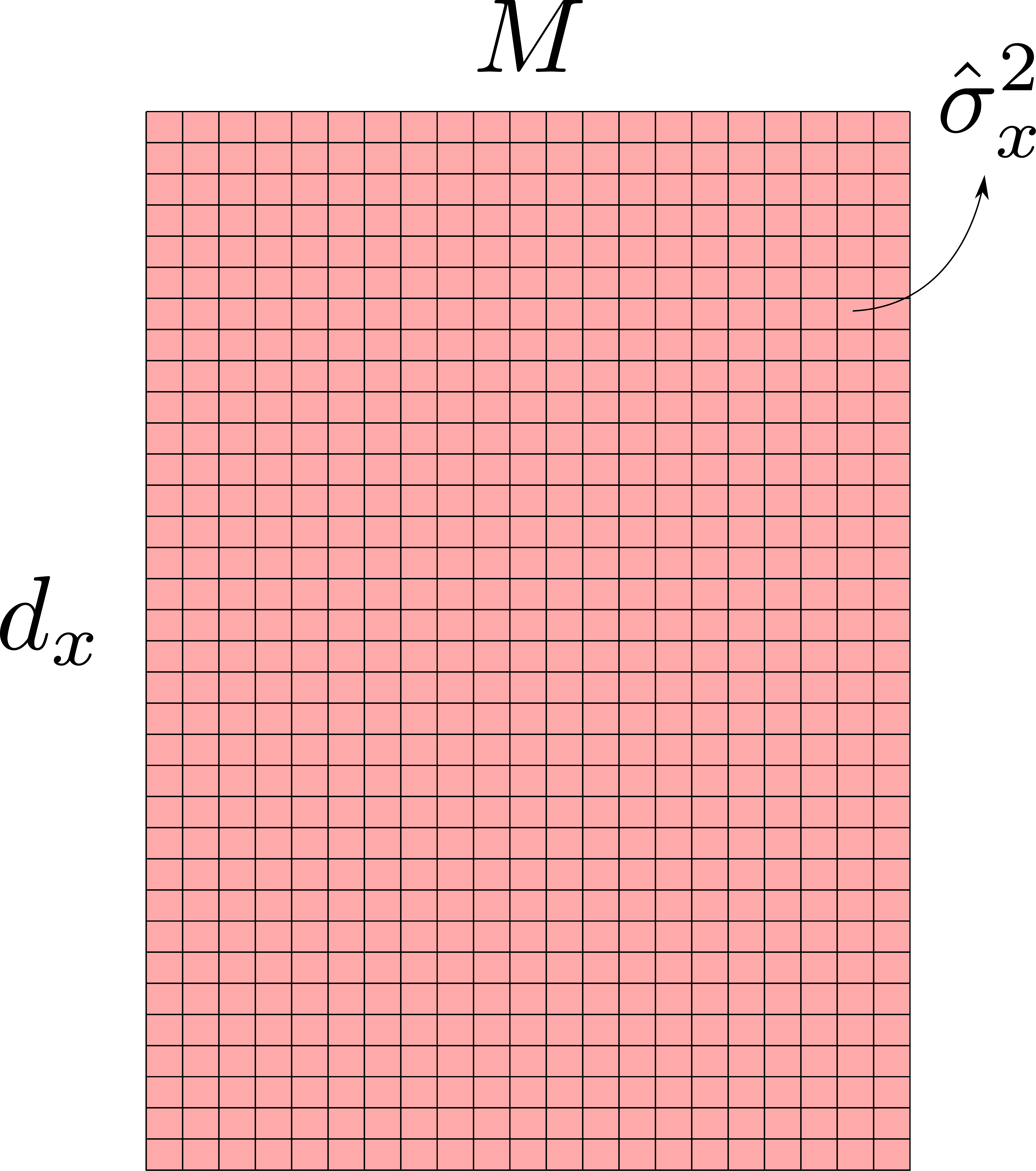}}
  \hspace{5pt}%
  \subfloat[Diag-I]{\includegraphics[height=0.25\textwidth]{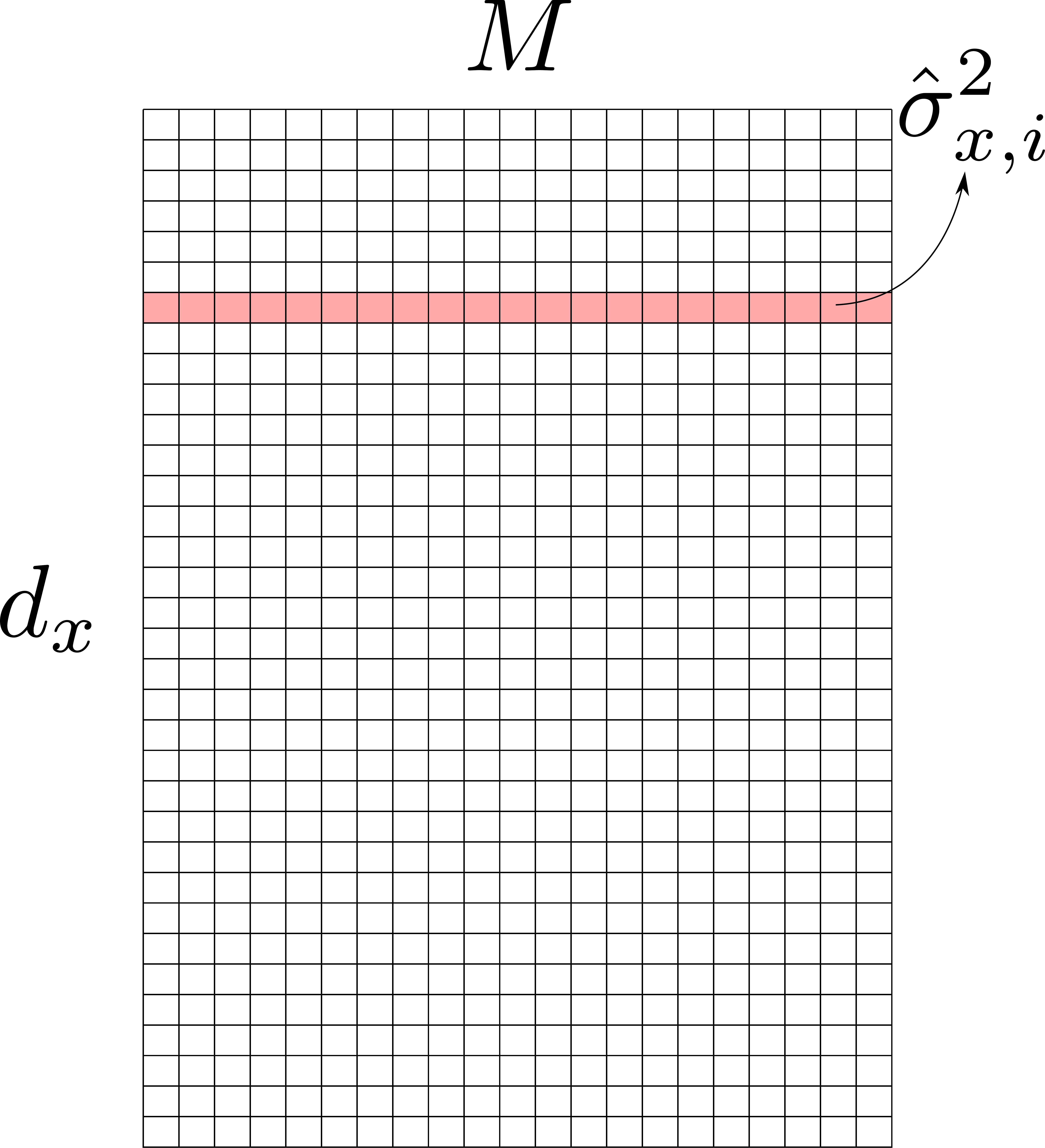}}
  \hspace{5pt}%
  \subfloat[Iso-D]{\includegraphics[height=0.25\textwidth]{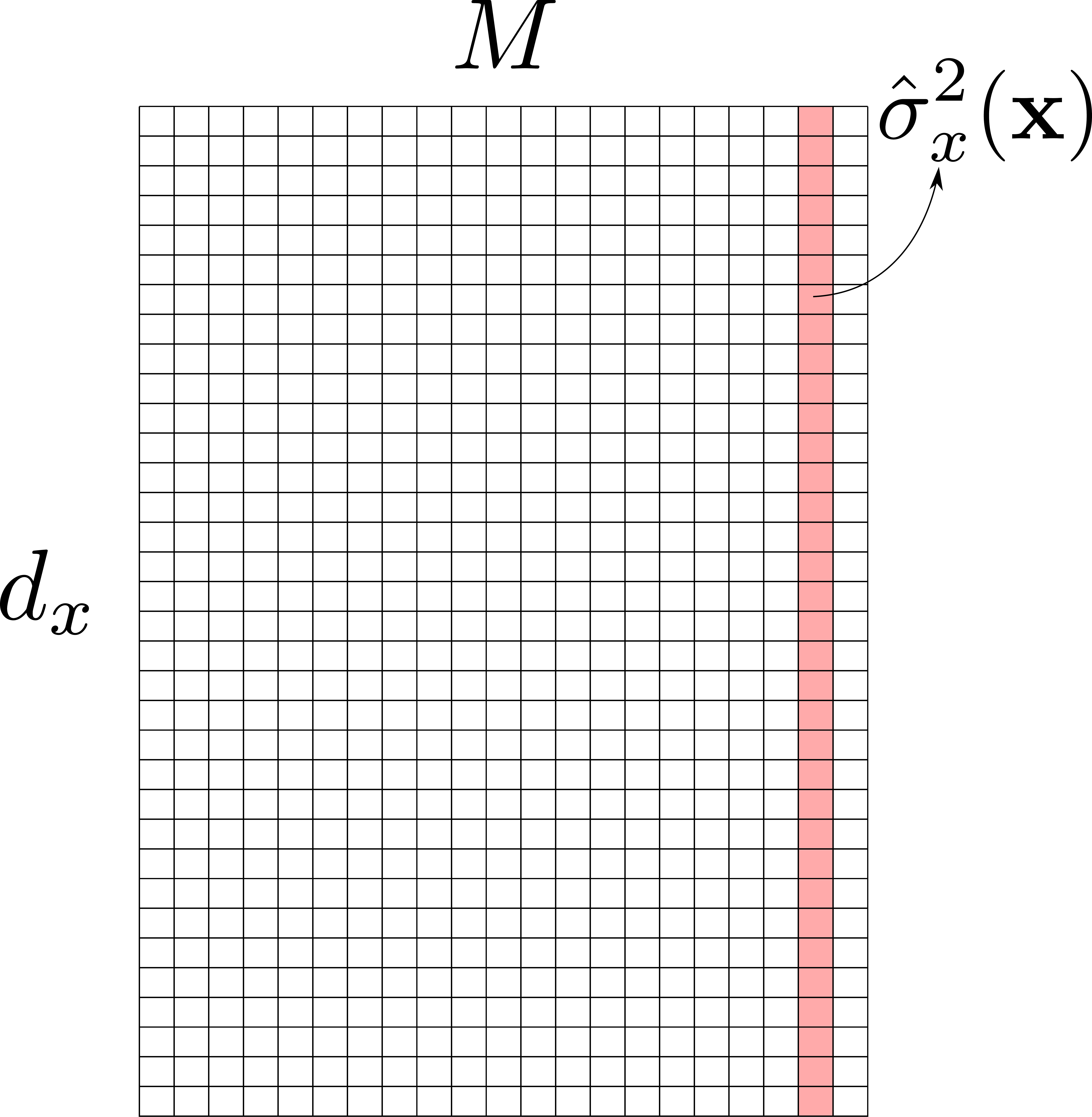}}
  \hspace{5pt}%
  \subfloat[Diag-D]{\includegraphics[height=0.25\textwidth]{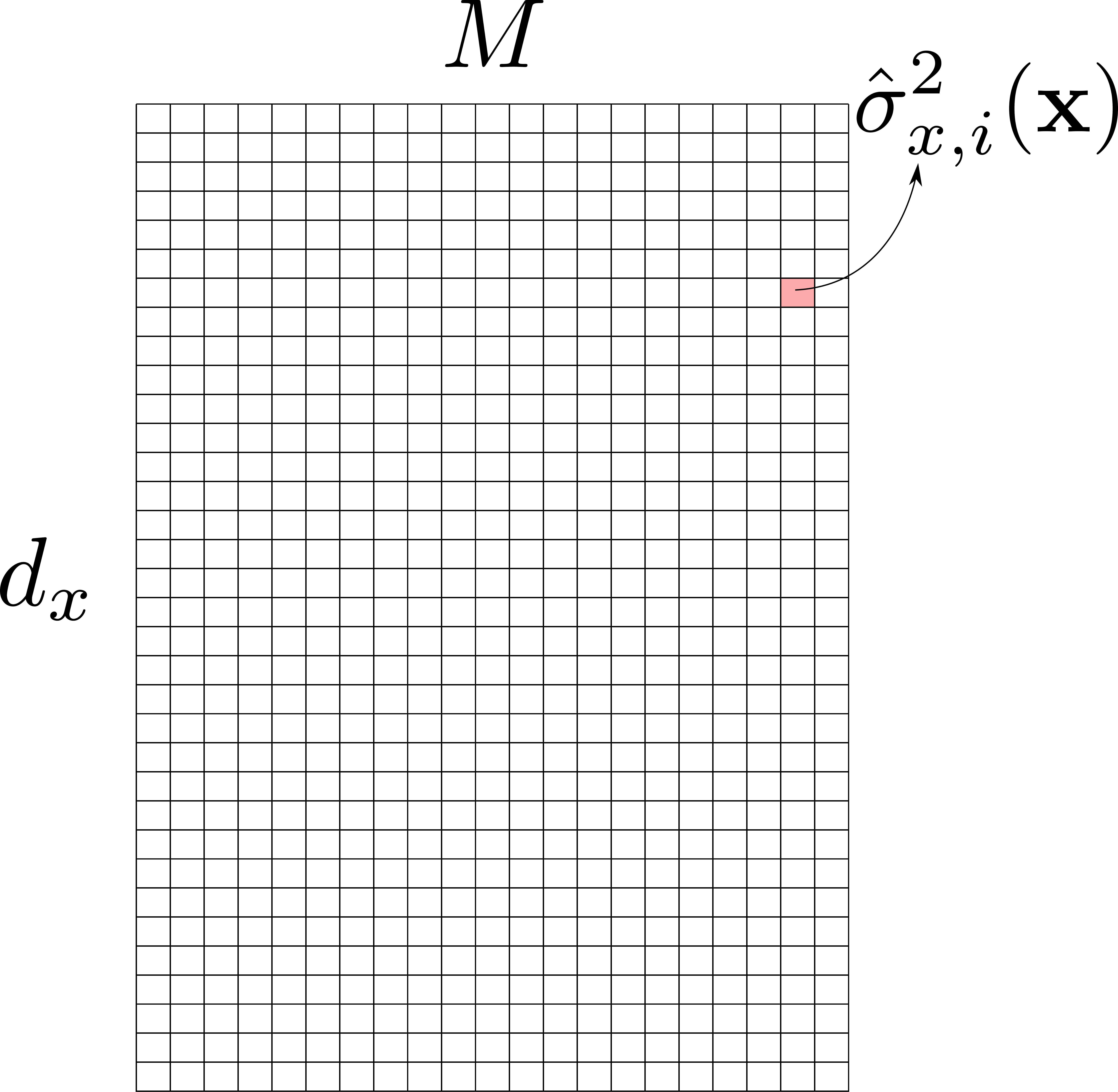}}
  \caption{\color{crevise}An illustration of the four parameterizations, where $M$ indicates batch size in a training phase. The matrix represents the data input and the pink area indicates the dependency of an element in $\Sigma_x$.}
  \label{fig:four_parameterizations}
\end{figure}

{\color{crevise}However, modeling $\bm{\Sigma}_x$ as a positive semi-definite matrix and attempting to estimate it via minibatch-based approaches often leads to unstable training in practice.} Therefore, we further concretize it into four parameterizations.
In these parameterizations, the variance parameter can not only be either an \textbf{isotropic} or \textbf{diagonal} matrix but also can be \textbf{independent} or \textbf{dependent} on $\mathbf{x}$. 
We denote these cases in Table~\ref{tb:models_and_objectives} as \textbf{Iso-I} (Isotropic-Independent), \textbf{Iso-D} (Isotropic-Dependent), \textbf{Diag-I} (Diagonal-Independent) and \textbf{Diag-D} (Diagonal-Dependent). {\color{crevise}Illustrative examples of these four parameterizations are shown in Figure~\ref{fig:four_parameterizations}}.
The first case, Iso-I, corresponds to the standard parameterization, i.e.  $\bm{\Sigma}_x=\sigma_x^2\mathbf{I}$.
Different parameterizations can be regarded as different weighting schemes for the balance between the reconstruction loss and the Jacobian of the decoder.
For example, the ELBO for Diag-D can be approximated as in Eq.~\beqref{eq:ELBO_linear_approx} with terms independent of $(\bm{\theta}, \bm{\phi})$ omitted: $\tilde{\Js}_{\text{Diag-D}}(\bm{\theta},\bm{\phi},\bm{\Sigma}_{x})\approx$
\begin{align}
    \frac{1}{2}\E_{\tilde{p}_{\text{data}}(\mathbf{x})}\left[\sum_{i=1}^{d_{x}}\frac{\left\|\mathbf{x}-\bm{\mu}_{\bm{\theta}}(\bm{\mu}_{\bm{\phi}}(\mathbf{x}))\right\|_2^2}{\sigma_{{x},i}^2(\mathbf{x})}
  +\sum_{i=1}^{d_{x}}\sum_{j=1}^{d_{z}}\frac{\sigma_{\bm{\phi},j}^2(\mathbf{x})}{\sigma_{{x},i}^2(\mathbf{x})}\left(\left.\frac{\partial \mu_{\bm{\theta},i}(\mathbf{z})}{\partial z_j}\right|_{\mathbf{z}=\bm{\mu}_{\bm{\phi}}(\mathbf{x})}\right)^2+\left\|\bm{\mu}_{\bm{\phi}}(\mathbf{x})\right\|_2^2\right].
\end{align}
Comparing this approximation with Eq.~\beqref{eq:ELBO_linear_approx}, since the variance parameters are both $\mathbf{x}$/$i$-dependent, it is clear that Diag-D offers much more flexibility than Iso-I.

\subsection{Optimization with MLE}
\label{sec:proposed_mle}
In this subsection, we consider to optimize the VAE objective function~\beqref{eq:objective_general_model} w.r.t. all the parameters including $\bm{\Sigma}_x$, which is usually modeled as $\sigma_x^2\mathbf{I}$ and fixed in existing implementations. 
Although the generalized parameterizations proposed in the previous section allow better fitting to the data variance, their optimization becomes more difficult.
Although we can implement the variance parameter $\bm{\Sigma}_x$ as trainable parameters, this na\"ive approach often leads to unstable training as demonstrated in Section~\ref{sec:experiment_comparison}.

\paragraph{Proposed objective function}
To overcome this difficulty, we propose to incorporate MLE into the ELBO. It is an update scheme that implicitly updates $\bm{\Sigma}_x$ with the rest of the parameters. 
According to the partial derivative of $\tilde{\Js}$ w.r.t. $\bm{\Sigma}_{x}$, the MLE of $\bm{\Sigma}_{x}$, denoted as $\hat{\bm{\Sigma}}_{x}$, can be evaluated with other parameters fixed. On the other hand, the ordinary network parameters $\bm{\theta}$ and $\bm{\phi}$ can also be updated by optimizing Eq.~\beqref{eq:objective_general_model} with the variance $\bm{\Sigma}_{x}$ fixed. This combination of MLE and the alternative update between $(\bm{\theta},\bm{\phi})$ and $\bm{\Sigma}_x$ guarantees that (i) if $\bm{\theta}$ and $\bm{\phi}$ are fixed, then there exists $\hat{\bm{\Sigma}}_{x}$ such that $\tilde{\Js}(\bm{\theta},\bm{\phi},\hat{\bm{\Sigma}}_{x})\leq\tilde{\Js}(\bm{\theta},\bm{\phi},\bm{\Sigma}_{x})$
and (ii) for the $\hat{\bm{\Sigma}}_{x}$ obtained in the previous step, there exist $\hat{\bm{\theta}}$ and $\hat{\bm{\phi}}$, such that  $\tilde{\Js}(\hat{\bm{\theta}},\hat{\bm{\phi}},\hat{\bm{\Sigma}}_{x})\leq\tilde{\Js}(\bm{\theta},\bm{\phi},\hat{\bm{\Sigma}}_{x})$. In this respect, the convergence of the optimization is assured and the parameter $\bm{\Sigma}_{x}$ is always kept as the result of MLE during the training. This equals to a weight scheduling scheme for $\hat{\bm{\Sigma}}_{x}$ and leads to a modified ELBO-based objective function. Consider the trainable network parameters $(\bm{\theta},\bm{\phi})$ and the variance parameter $\bm{\Sigma}_{x}$. The update scheme of the objective $\tilde{\Js}(\bm{\theta},\bm{\phi},\bm{\Sigma}_{x})$ is
\begin{subequations}
  \begin{align}
    &\bm{\Sigma}_{x}^{(t+1)}
    = \E_{\tilde{p}_{\text{data}}(\mathbf{x})q_{\bm{\phi}^{(t)}}(\mathbf{z}|\mathbf{x})}
    \left[(\mathbf{x}-\bm{\mu}_{\bm{\theta}^{(t)}}(\mathbf{z}))(\mathbf{x}-\bm{\mu}_{\bm{\theta}^{(t)}}(\mathbf{z}))^\top\right]
    \label{eq:kl_anneal_1}\\
    &\bm{\theta}^{(t+1)},\bm{\phi}^{(t+1)}
    = \argmin_{\bm{\theta},\bm{\phi}}~\tilde{\Js}(\bm{\theta},\bm{\phi},\bm{\Sigma}_{x}^{(t+1)}),
    \label{eq:kl_anneal_2}
  \end{align}
\end{subequations}
where $t$ is the iteration index. 
The update scheme above can be further simplified by substituting Eq.~\beqref{eq:kl_anneal_1} into Eq.~\beqref{eq:objective_general_model}, which converts $\tilde{\Js}(\bm{\theta},\bm{\phi},\hat{\bm{\Sigma}}_x)$ into
\begin{align}
  \tilde{\Js}_{\text{MLE}}(\bm{\theta},\bm{\phi})
  &= \tilde{\Js}_{\text{rec}}(\bm{\theta},\bm{\phi},\hat{\bm{\Sigma}}_x)+\E_{\tilde{p}_{\text{data}}(\mathbf{x})}\KL\left(q_{\bm{\phi}}(\mathbf{z}|\mathbf{x})\parallel p(\mathbf{z})\right)
  \label{eq:cost_model_general}
\end{align}
where all constant terms w.r.t. the parameters are omitted.
Moreover, optimizing Eq.~\beqref{eq:cost_model_general} guarantees that $\bm{\Sigma}_x$ remain as the result of MLE during the VAE training.
To be mentioned, due to the incorporation of MLE, the reconstruction objectives are no longer MSE. These reconstruction objectives of $\bm{\Sigma}_x$ are summarized in Table~\ref{tb:models_and_objectives} and their derivations can be found in \ref{sec:derivation_objectives}. {\color{crevise}This proposed update scheme is summarized as Algorithm~\ref{alg:proposed}.}

It is interesting to note that, for Diag-D, its reconstruction error becomes the sum of logarithms of the MSE of each dimension in the data space; meanwhile, only ordinary MSE is required for Iso-I.
Considering the optimization stability in practical situations, we suggest adding a small constant, e.g., $10^{-6}$, before taking the logarithms except for Iso-I.

It should noted that all the derived objective functions are biased estimations due to the logarithm of expectation.
Although the objectives are no longer equals to the ELBO, we found that they still work in practical situations, see Section~\ref{sec:experiment_comparison}.
On the other hand, it is difficult to stabilize the optimization of the most general setup, i.e. when $\bm{\Sigma}_x$ is a full positive semi-definite matrix. Although its update scheme with MLE can still be derived from Eq.~\beqref{eq:kl_anneal_1}, the rank of $\hat{\bm{\Sigma}}_x$ is capped by the batch size, which is usually much smaller than the dimension of the variance parameter matrix $\bm{\Sigma}_x$. Regularizing $\hat{\bm{\Sigma}}_x$ by $\lambda\mathbf{I}$ with small $\lambda$ avoids the rank deficiency issue but still not enough to stabilize the training in practice.


\paragraph{Adaptive regularization}
Eq.~\beqref{eq:cost_model_general} can also be interpreted as an adaptive regularization, which seeks to balance between the KL divergence term and the reconstruction loss. This can be explained by considering the Iso-I parameterization.
In the alternative update scheme, Eq.~\beqref{eq:kl_anneal_2} is the same as the parameter update in the ordinary VAE. On the other hand, Eq.~\beqref{eq:kl_anneal_1} can be interpreted as an extra step that determines the balance between the reconstruction error and the KL term in $\tilde{\Js}_{\sigma_x^2}(\bm{\theta},\bm{\phi})$. As the learning progresses, the parameter $\sigma_x^{2}$ will decrease along with the MSE $\E_{\tilde{p}_{\text{data}}(\mathbf{x})q_{\bm{\phi}}(\mathbf{z}|\mathbf{x})}[\|\mathbf{x}-\bm{\mu}_{\bm{\theta}}(\mathbf{z})\|_2^2]$, which is consistent with the discussion in \cite{dai2019diagnosing}.
Also, as stated by Theorem~\ref{th:convergence_var}, decreasing $\sigma_x^2$ also decreases $\sigma_z^2$. This gradually relieves the regularization of the ELS~\beqref{eq:def_els}, which can be observed from Eq.~\beqref{eq:ELBO_linear_approx}. However, this eventually diminishes the gradient penalty; therefore, we suggest using early-stopping and learning rate scheduling to achieve both appropriate smoothness and generalization capability.

\paragraph{Remark}
Although the proposed update scheme adapts $\bm{\Sigma}_x$ appropriately, there are other issues that should be addressed. For example, there usually exists a gap between the prior $p(\mathbf{z})$ and the aggregated posterior $q_{\bm{\phi}}(\mathbf{z})$. This can be observed by reformulating Eq.~\beqref{eq:objective_vae} (see \ref{sec:match_pri_pos}):
\begin{align}
  \Ls
  =&\E_{q_{\bm{\phi}}(\mathbf{z})}\KL\left(q_{\bm{\phi}}(\mathbf{x}|\mathbf{z})\parallel p_{\bm{\theta}}(\mathbf{x}|\mathbf{z})\right)
  +\KL\left(q_{\bm{\phi}}(\mathbf{z})\parallel p(\mathbf{z})\right).
  \label{eq:vae_cost_two_kl}
\end{align}
If the first term in Eq.~\beqref{eq:vae_cost_two_kl} becomes dominant during the VAE training, the gap between $q_{\bm{\phi}}(z)$ and $p(z)$ cannot be mitigated effectively. In this situation, generation through sampling the prior may results in off-distribution samples. To overcome this prior--posterior mismatch, two types of approaches are often adopted: (i) conduct another posterior estimation after the ordinary VAE training~\cite{van2017neural,razavi2019generating,dai2019diagnosing,ghosh2019from,morrow2020variational} or (ii) add another regularizing term to the objective function~\cite{makhzani2015adversarial,tolstikhin2018wasserstein,zhao2019infovae}.
We adapt the former approach due to its efficiency~\cite{ghosh2019from} and the ease of application.

To summarize, we proposed generalized parameterizations for VAE with a MLE-based update scheme that adaptively weights the gradient penalty without the need of tuning extra hyperparameters. 
The remaining mismatch between the prior and posterior is mitigated by an extra pass of posterior estimation. 

\begin{center}
\color{crevise}
\begin{minipage}{.90\textwidth}
\begin{algorithm}[H]
\color{crevise}
   \caption{\color{crevise}Proposed MLE-based update scheme}
   \label{alg:proposed}
\begin{algorithmic}
   \STATE {\bfseries Input:} Dataset $\mathbf{x}_\mathrm{data}$
   \STATE Initialize the parameters of the encoder and decoder: $\bm{\theta}^{[0]}$ and $\bm{\phi}^{[0]}$ 
   \FOR{$t=1,2,\ldots,T$}
   \STATE $\mathbf{x} \leftarrow$ Random minibatch from $\mathbf{x}_\mathrm{data}$
   \STATE $\mathbf{z} \leftarrow \mathcal{N}(\mathbf{z}|\bm{\mu}_{\bm{\phi}}(\mathbf{x}), \bm{\Sigma}_{\bm{\phi}}(\mathbf{x}))$
   \STATE $\textbf{g}\leftarrow  \nabla_{\bm{\theta},\bm{\phi}}\tilde{\Js}_\mathrm{MLE}(\bm{\theta}^{[t-1]}, \bm{\phi}^{[t-1]})$ (\eqref{eq:cost_model_general}) with sampled $\mathbf{x}$ and $\mathbf{z}$,
   \STATE where $\tilde{\Js}_\mathrm{MLE}$ is configured with the corresponding $\tilde{\Js}_\mathrm{rec}$ in Table~\ref{tb:models_and_objectives}
   \STATE $\bm{\theta}^{[t]},\bm{\phi}^{[t]} \leftarrow$ Update parameters using $\textbf{g}$
   \ENDFOR
\end{algorithmic}
\end{algorithm}
\end{minipage}
\end{center}

\begin{table}[t]
  \centering
  \caption{Parameterizations of posterior variance in $\mathcal{X}$ and corresponding reconstruction objectives.}
  \renewcommand{\arraystretch}{1.5}
  \small
  \begin{tabular}{c|c|c}
    \bhline{0.8pt}
     & Variance model ($\bm{\Sigma}_x$) & Reconstruction objective ($\tilde{\Js}_{\text{rec}}(\bm{\theta},\bm{\phi},\hat{\bm{\Sigma}}_x)$)\\
    \hline
    (Iso-I) & $\sigma_x^2\mathbf{I}$
    & $\frac{d_x}{2}\ln\E_{\tilde{p}_{\text{data}}(\mathbf{x})q_{\bm{\phi}}(\mathbf{z}|\mathbf{x})}[\|\mathbf{x}-\bm{\mu}_{\bm{\theta}}(\mathbf{z})\|_2^2]$\\
    (Iso-D)  & $\sigma_x^2(\mathbf{x})\mathbf{I}$
    & $\frac{d_x}{2}\E_{\tilde{p}_{\text{data}}(\mathbf{x})}\left[\ln\E_{q_{\bm{\phi}}(\mathbf{z}|\mathbf{x})}\|\mathbf{x}-\bm{\mu}_{\bm{\theta}}(\mathbf{z})\|_2^2\right]$\\
    (Diag-I) & $\diag(\bm{\sigma}_x^2)$
    & $\frac{1}{2}\sum_{i=1}^{d_x}\ln\E_{\tilde{p}_{\text{data}}(\mathbf{x})q_{\bm{\phi}}(\mathbf{z}|\mathbf{x})}[(x_i-\mu_{\bm{\theta},i}(\mathbf{z}))^2]$\\
    (Diag-D) & $\diag(\bm{\sigma}_x^2(\mathbf{x}))$
    & $\frac{1}{2}\sum_{i=1}^{d_x}\E_{\tilde{p}_{\text{data}}(\mathbf{x})}\left[\ln\E_{q_{\bm{\phi}}(\mathbf{z}|\mathbf{x})}(x_i-\mu_{\bm{\theta},i}(\mathbf{z}))^2\right]$\\
    \bhline{0.8pt}
  \end{tabular}
  \label{tb:models_and_objectives}
\end{table}

\vspace{0.0pt}%
\subsection{Correctness of variance estimation}
\vspace{0.0pt}
\label{sec:investigation_variance}
In this subsection, we investigate the correctness of variances estimated from our MLE-based objective function with the proposed parameterizations on a modified MNIST dataset. 

In order to simulate the conditional and spatial dependent data variance, we first divide the MNIST images into two groups; digits 0$\sim$4 and 5$\sim$9. Simultaneously, we divide the upper-half and the lower-half of images into two other groups. 
The non-empty intersections of these groups form four partitions: \textbf{SU} (Small-Upper), \textbf{SL} (Small-Lower), \textbf{BU} (Big-Upper) and \textbf{BL} (Big-Lower). Then, four datasets are created by injecting artificial noise with the four following patterns: \textit{uniform}, \textit{spatial}, \textit{conditional}, and \textit{spatial-conditional}. For the uniform pattern, a Gaussian noise with variance $s_x^2$ is injected to all the partitions. For the spatial pattern, the noise is only injected to the upper-half of images, i.e., ``SU'' and ``BU'' partitions. For the conditional pattern, the noise is only injected into those images that are labeled as 0$\sim$4, i.e., ``SU'' and ``SL'' partitions. For the spatial-conditional pattern, the noise is injected only to the upper-half of images labeled as 0$\sim$4, i.e. the ``SU'' partition. Two cases of $s_x^2$, which are $s_x^2=0.01$ and $0.1$, are used in the investigation. Partitions with noise injected are noisy partitions, where the rest are clean partitions.

Next, we train the VAE extensions proposed in Section~\ref{sec:proposed_objectives} with the MLE update scheme proposed in Section~\ref{sec:proposed_mle} on those modified datasets. These models use the same network architecture as Section~\ref{sec:experiment_comparison} and no information about the noise pattern is given to these models. Since the true data variance is unknown, we use the variance difference between the original and the modified dataset as the indicator of estimation accuracy. 

Table~\ref{tb:estimated_noise_variance} summarizes the expected values of estimated variances for each partition and each noise pattern. All the differences of estimated variances between the noisy partitions and clean partitions are close to $s^2_x$ within a reasonable margin. The result shows that the MLE update scheme is reliable as long as the complexity of the noise pattern fits within the assumption of its parameterization.

\begin{table}[t]
  \centering
  \caption{Estimated noise variances by the MLE-based objective function when the training is finished. $\hat{\sigma}^2$ is the average value of the estimated variances of the partition in the subscript. The cases underlined indicates that those parameterizations are sufficient to model the corresponding noise pattern in theory. Among these cases, bold texts are used to represent noisy partitions. All the estimations land within $5\%$ of $s^2_x+0.007$, where $0.007$ is the estimated variance of the original MNIST dataset.} 
  \small
\resizebox{\columnwidth}{!}{
  \begin{tabular}{lc|c|cc|cc|cccc}
    \bhline{0.8pt}
    \multirow{2}{*}{Noise pattern} & \multirow{2}{*}{$s_x^2$}  & Iso-I & \multicolumn{2}{c|}{Diag-I} & \multicolumn{2}{c|}{Iso-D} & \multicolumn{4}{c}{Diag-D}\\ \cline{3-11}
     &  & $\hat{\sigma}_{\text{SU+SL+BU+BL}}^2$ & $\hat{\sigma}_{\text{SU+BU}}^2$ & $\hat{\sigma}_{\text{SL+BL}}^2$ & $\hat{\sigma}_{\text{SU+SL}}^2$ & $\hat{\sigma}_{\text{BU+BL}}^2$ & $\hat{\sigma}_{\text{SU}}^2$ & $\hat{\sigma}_{\text{SL}}^2$ & $\hat{\sigma}_{\text{BU}}^2$ & $\hat{\sigma}_{\text{BL}}^2$ \\
    \hline
    \textit{Uniform}             & 0.01 & \underline{\textbf{0.013}} & \underline{\textbf{0.014}} & \underline{\textbf{0.013}} & \underline{\textbf{0.014}} & \underline{\textbf{0.014}} & \underline{\textbf{0.017}} & \underline{\textbf{0.016}} & \underline{\textbf{0.017}} & \underline{\textbf{0.017}} \\
    $s^2_x: \text{SU+SL+BU+BL}$ & 0.1  & \underline{\textbf{0.10}} & \underline{\textbf{0.11}} & \underline{\textbf{0.099}} & \underline{\textbf{0.10}} & \underline{\textbf{0.10}} & \underline{\textbf{0.11}} & \underline{\textbf{0.10}} & \underline{\textbf{0.11}} & \underline{\textbf{0.10}} \\
    \hline
    \textit{Spatial}             & 0.01 & 0.0089 & \underline{\textbf{0.016}} & \underline{0.0042} & 0.0089 & 0.0091 & \underline{\textbf{0.017}} & \underline{0.0069} & \underline{\textbf{0.017}} & \underline{0.0072} \\
    $s^2_x: \text{SU+BU}$        & 0.1  & 0.055 & \underline{\textbf{0.11}} & \underline{0.0038} & 0.055 & 0.055 & \underline{\textbf{0.11}} & \underline{0.0065} & \underline{\textbf{0.11}} & \underline{0.0070} \\
    \hline
    \textit{Conditional}         & 0.01 & 0.0085 & 0.0087 & 0.0087 & \underline{\textbf{0.015}} & \underline{0.0037} & \underline{\textbf{0.017}} & \underline{\textbf{0.017}} & \underline{0.0073} & \underline{0.0076} \\ %
    $s^2_x: \text{SU+SL}$        & 0.1  & 0.053 & 0.055 & 0.051 & \underline{\textbf{0.11}} & \underline{0.0042} & \underline{\textbf{0.11}} & \underline{\textbf{0.10}} & \underline{0.0064} & \underline{0.0067} \\ %
    \hline
    \textit{Spatial-conditional} & 0.01 & 0.0061 & 0.0099 & 0.0045 & 0.0095 & 0.0034 & \underline{\textbf{0.017}} & \underline{0.0070} & \underline{0.0070} & \underline{0.0073} \\ %
    $s^2_x: \text{SU}$           & 0.1  & 0.029 & 0.059 & 0.0038 & 0.059 & 0.0040 & \underline{\textbf{0.11}} & \underline{0.0069} & \underline{0.0068} & \underline{0.0066} \\ %
    \bhline{0.8pt}
  \end{tabular}
  }
  \label{tb:estimated_noise_variance}
\end{table}

\begin{table}[th!]
  \centering
  \caption{The result of numerical evaluation on MNIST and CelebA. The MSE of sample reconstruction is evaluated on the test set. The FIDs of generated samples are measured in three cases: (i) sampling the latent variables from the prior; (ii) sampling from the posterior estimated by \nth{2} VAE; and (iii) sampling from the posterior estimated by GMM.}
\resizebox{\columnwidth}{!}{
  \begin{threeparttable}
  \begin{small}
  \begin{tabular}{l|cccc|ccccc}
    \bhline{0.8pt}
    \renewcommand{\arraystretch}{1.2}
    \rule{0pt}{8pt} & \multicolumn{4}{c|}{MNIST}                            & \multicolumn{5}{c}{CelebA} \\ \cline{2-10}
    \rule{0pt}{8pt} & \multirow{2}{*}{MSE}     & \multicolumn{3}{c|}{FID}   & \multirow{2}{*}{MSE} & \multicolumn{4}{c}{FID} \\ \cline{3-5}\cline{7-10}
    \rule{0pt}{8pt} &  & Prior                 & \nth{2}VAE & GMM10               & &  Prior & \nth{2}VAE & GMM10 & GMM100 \\
    \bhline{0.8pt}
    \renewcommand{\arraystretch}{1.0}
    \hspace{-2pt}VAE w/ Gaussian ($\sigma_x^2=1.0$)        & 20.25 & 55.85 & 182.64 & 58.96 & 121.91 & 55.46 & 139.32 & 54.66 & 53.94 \\
    \color{crevise}VAE w/ Bernoulli                             & \color{crevise}7.26 & \color{crevise}19.16 & \color{crevise}19.65 & \color{crevise}17.49 & -- & -- & -- & -- & -- \\
    \color{crevise}VAE w/ MoL                               & -- & -- & -- & --  & \color{crevise}163.68 & \color{crevise}269.18 & \color{crevise}148.18 & \color{crevise}153.08 & \color{crevise}149.02 \\
    AE                                         & 4.31 & -- & 20.66 & 13.20 & 61.44 & -- & 62.34 & 46.47 & 43.41 \\
    WAE-MMD                                    & 4.34 & 22.76 & 15.00 & 13.70 & 62.66 & \textbf{52.89} & 51.32 & 43.57 & 41.88 \\
    RAE                                        & \textbf{4.28} & -- & 18.54 & 13.68 & 61.49 & -- & 57.26 & 46.50 & 43.89 \\
    RAE-GP                                     & 4.30 & -- & 18.89 & 13.71 & 61.48 & -- & 54.54 & 43.63 & 41.10 \\
    \hline
    Trainable $\bm{\Sigma}_x^\dagger$ & \multicolumn{4}{l|}{} & \multicolumn{5}{l}{}\\
    Iso-I               & 4.28 & 20.93 & 14.83$^\ddagger$ & 13.39 & 61.42 & 63.21 & 63.12$^\ddagger$ & 51.40 & 49.61 \\
    Diag-I  & 5.70 & 26.08 & 18.44 & 15.72 & 62.65 & 59.66 & 54.75 & 45.86 & 43.50 \\
    Iso-D & 4.45 & 27.40  & 16.38 & 13.67 & 65.29 & 195.58 & 53.13 & 50.09 & 47.31\\
    Diag-D &  5.33 & 146.43 & 26.02 &  27.90 & 85.33$^*$ & 354.20$^*$ & 217.06$^*$ & 188.30$^*$ & 188.05$^*$\\
    \hline
    Proposed MLE (Ours) & \multicolumn{4}{l|}{} & \multicolumn{5}{l}{}\\
    Iso-I                     & 4.40 & 22.78 & 15.77 & 12.21 & 62.02 & 82.20 & 52.48 & \textbf{42.82} & 41.03 \\
    Diag-I                    & 5.35 & 24.15 & 17.18 & 13.38 & 63.51 & 87.44 & 53.60 & 45.85 & 42.83 \\
    Iso-D                     & 4.31 & 22.94 & 17.57 & 12.89 & \textbf{61.38} & 78.24 & \textbf{49.97} & 43.27 & \textbf{40.39} \\
    Diag-D                    & 6.80 & \textbf{16.64} & \textbf{10.49} & \textbf{10.05} & 70.75 & 64.40 & 55.30 & 46.63 & 45.27 \\
    \hline
    \bhline{0.8pt}
  \end{tabular}

  \begin{tablenotes}
    \item[${}^\dagger$] $\bm{\Sigma}_x$ is learned as other trainable parameters as in~\cite{dai2019diagnosing}. However, the work does not include parameterizations such as Diag-I, Iso-D and Diag-D.
    \item[${}^\ddagger$] \nth{2}VAE is a second-stage VAE after the main VAE , which is proposed in~\cite{dai2019diagnosing}.
    \item[${}^*$] The training of Diag-D with $\bm{\Sigma}_x$ learned as trainable parameters does not converge to a local optima. Moreover, the trend of MSE diverges with that of the loss function. Therefore, the result which achieves the best MSE on the test set is reported here. On the other hand, training Diag-D with the proposed MLE does not suffer from this issue.
  \end{tablenotes}
  \end{small}
  \end{threeparttable}
  }
  \label{tb:main_results}
\end{table}

\begin{figure}[tb]
  \color{crevise}
  \centering
  \subfloat[MNIST]{\includegraphics[height=0.35\textwidth]{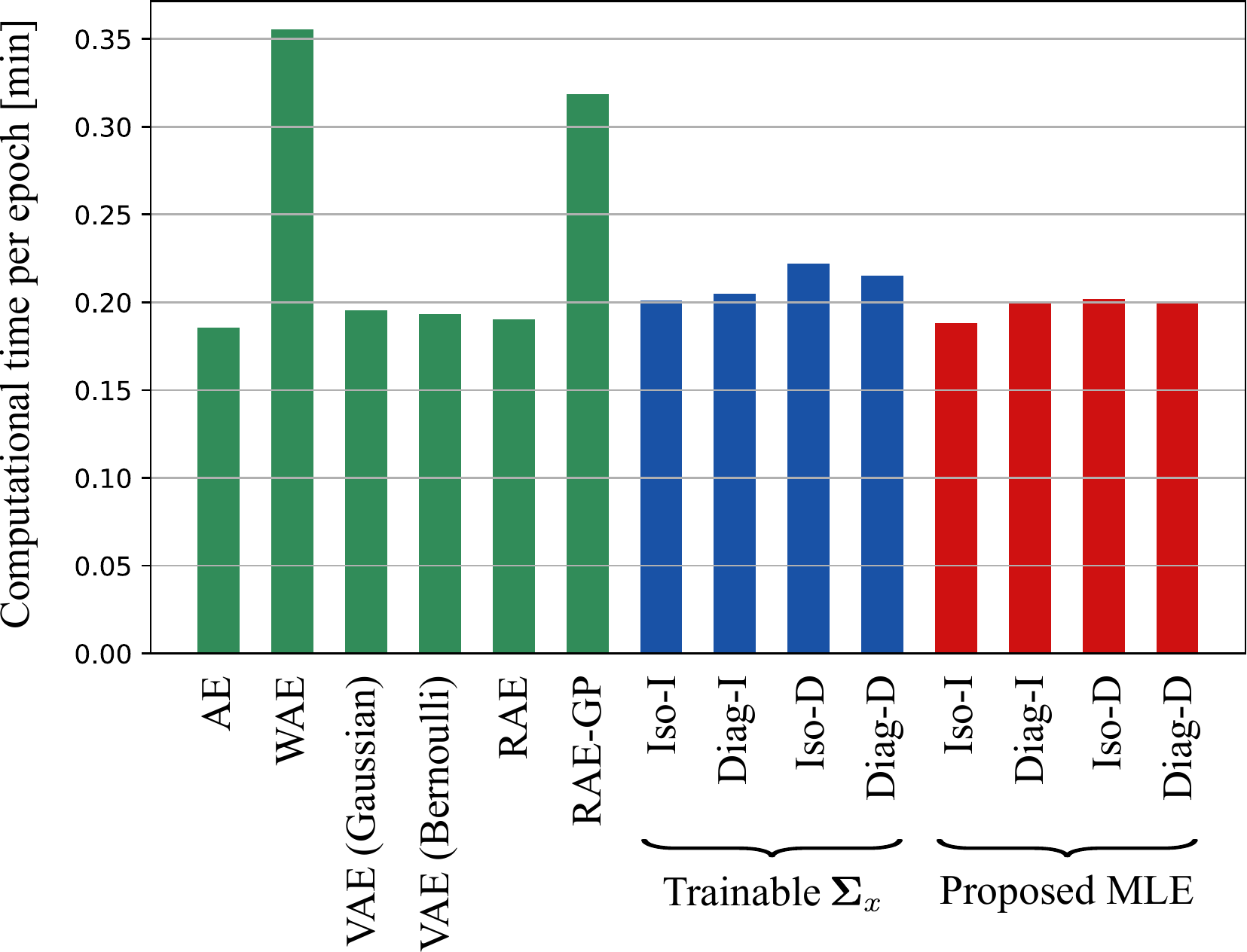}}
  \hspace{20pt}%
  \subfloat[CelebA]{\includegraphics[height=0.35\textwidth]{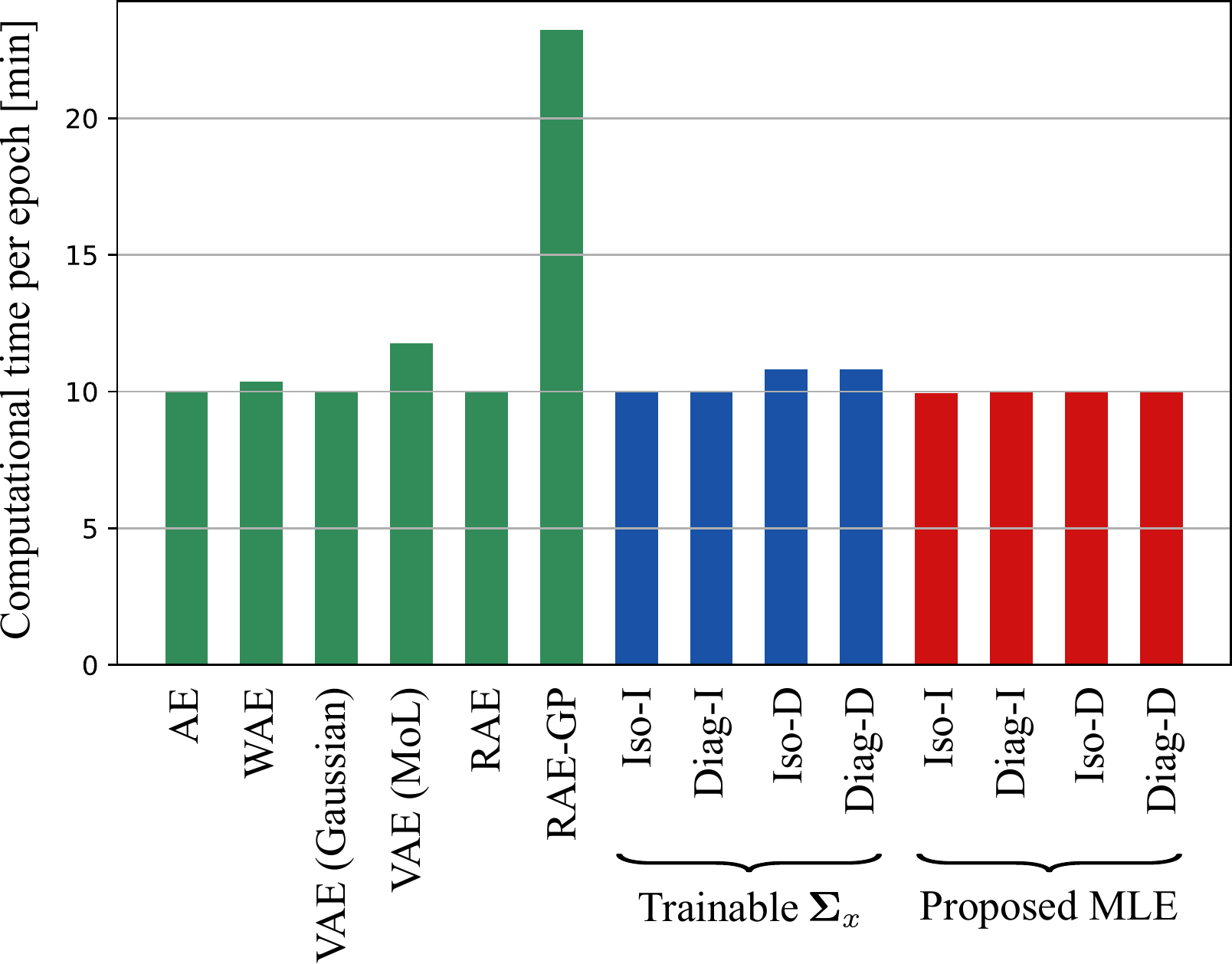}}
  \caption{\color{crevise}Computational time per epoch on MNIST and CelebA.}
  \label{fig:training_time_per_epoch}
\end{figure}

\begin{table}[th!]
  \centering
  \caption{Evaluation of FID scores of interpolated images on MNIST and CelebA. The interpolation ratio for each image pair is designated as (i) the mid-point; and (ii) a random-point between the two.}
  \small
  \begin{threeparttable}
  \begin{tabular}{l|cc|cc}
    \bhline{0.8pt}
    \renewcommand{\arraystretch}{1.2}
    \rule{0pt}{8pt}                            & \multicolumn{2}{c|}{MNIST}                            & \multicolumn{2}{c}{CelebA} \\ \cline{2-5}
    \rule{0pt}{8pt}                            & Mid-point & Random-point                & Mid-point & Random-point   \\
    \bhline{0.8pt}
    \renewcommand{\arraystretch}{1.0}
    \hspace{-2pt}VAE w/ Gaussian ($\sigma_{{x}}^2=1.0$)    & 62.18             & 63.86          & 57.87              & 55.54 \\
    \color{crevise}VAE w/ Bernoulli    & \color{crevise}18.20             & \color{crevise}18.08          & --              & -- \\
    \color{crevise}VAE w/ MoL    & --             & --          & \color{crevise}137.45              & \color{crevise}138.26 \\
    WAE-MMD                                    & 17.27             & 12.41          & \textbf{41.93}     & \textbf{38.80} \\
    AE                                         & 18.49             & 12.81          & 50.35              & 45.01 \\
    RAE                                        & 17.71             & 12.99          & 48.78              & 43.97 \\
    RAE-GP                                     & 17.98             & 12.96          & 45.22              & 40.58 \\
    \hline
    trainable $\bm{\Sigma}_x$ & \multicolumn{2}{l|}{} & \multicolumn{2}{l}{}\\
    Iso-I          & 15.66             & 12.73          & 52.01              & 48.83 \\
    Diag-I          & 17.34             & 14.77          & 44.22              & 41.23 \\
    Iso-D          & 17.03             & 13.33          & 48.51              & 43.97 \\
    Diag-D         & 51.65             & 31.66          & 238.74              & 212.39 \\
    \hline
    Proposed MLE (Ours) & \multicolumn{2}{l|}{} & \multicolumn{2}{l}{}\\
    Iso-I                    & 14.77             & 11.27          & 42.59              & 39.19 \\
    Diag-I                   & 17.24             & 13.08          & 45.55              & 41.92 \\
    Iso-D                    & 15.30             & 12.01          & 42.49              & 38.93 \\
    Diag-D                   & \textbf{11.96}    & \textbf{8.68}   & 46.86              & 44.15 \\
    \hline
    \bhline{0.8pt}
  \end{tabular}
  \end{threeparttable}
  \label{tb:main_interpolation}
\end{table}

\begin{table}[th!]
  \centering
  \color{crevise}
  \caption{{\color{crevise}The result of numerical evaluation on CelebAHQ. The MSE of sample reconstruction is evaluated on the test set. The FIDs of generated samples are measured in three cases: (i) sampling the latent variables from the prior; (ii) sampling from the posterior estimated by \nth{2} VAE; and (iii) sampling from the posterior estimated by GMM.}}
\resizebox{\columnwidth}{!}{
  \begin{small}
  \begin{tabular}{l|ccccc}
    \bhline{0.8pt}
    \renewcommand{\arraystretch}{1.2}
     & \multirow{2}{*}{MSE} & \multicolumn{4}{c}{FID} \\ \cline{3-6}
    \rule{0pt}{8pt} &  & Prior & \nth{2}VAE & GMM10 & GMM100 \\
    \bhline{0.8pt}
    \renewcommand{\arraystretch}{1.0}
    \hspace{-2pt}{\color{crevise}AE}                                         & 199.51 & -- & 157.46 & 94.53 & 85.70 \\
    {\color{crevise}WAE-MMD}                                    & 207.82 & 246.79 & 98.77 & 96.26 & 88.15 \\
    {\color{crevise}VAE w/ Gaussian ($\sigma_x^2=1.0$)}        & 404.66 & \textbf{99.71} & 139.12 & 89.54 & 88.62 \\
    {\color{crevise}VAE w/ MoL}                               & 552.17 & 396.41 & 300.80 & 287.81 & 270.12 \\
    {\color{crevise}RAE}                                        & \textbf{197.84} & -- & 95.47 & 94.54 & 89.14 \\
    {\color{crevise}RAE-GP}                                     & 202.09 & -- & 100.33 & 103.63 & 94.91 \\
    \hline
    \hline
    Proposed MLE (Ours) & \multicolumn{5}{l}{}\\
    {\color{crevise}Iso-I}                     & 203.90 & 227.72 & 87.95 & 85.24 & 78.15 \\
    {\color{crevise}Diag-I}                    & 206.56 & 255.48 & 88.63 & 87.11 & 78.98 \\
    {\color{crevise}Iso-D}                     & 203.43& 222.87 & 88.15 & 84.77 & 76.28 \\
    {\color{crevise}Diag-D}                    & 211.68 & 257.42 & \textbf{82.50} & \textbf{79.00} & \textbf{71.46} \\
    \hline
    \bhline{0.8pt}
  \end{tabular}

  \end{small}
  }
  \label{tb:celebahq_results}
\end{table}

\begin{table}[th!]
  \centering
  \color{crevise}
  \caption{{\color{crevise}The result of numerical evaluation on MNIST and CelebA. The upper bound of the negative log likelihood (nats) is obtained with $k$-sample importance weighting with $k=500$ on the test set.}}
\resizebox{\columnwidth}{!}{
  \begin{small}
  \begin{tabular}{l|ccccc}
    \bhline{0.8pt}
     & VAE ($\sigma_x^2=1.0$) & Iso-I & Diag-I & Iso-D & Diag-D \\
    \bhline{0.8pt}
    MNIST & $<736.61$ & $<-887.17$ & $<-1390.88$ & $<-995.87$ & $<\textbf{-2113.80}$\\
    \hline
    CelebA & $<11381.26$ & $<-14899.38$ & $<-15161.17$ & $<-15739.12$ & $<\textbf{-24288.90}$\\
    \hline
    CelebAHQ & $<45455.85$ & $<-64060.25$ & $<-65139.82$ & $<-66509.64$ & $<\textbf{-103572.99}$\\
    \bhline{0.8pt}
  \end{tabular}
  \end{small}
  }
  \label{tb:proposed_nll}
\end{table}

\vspace{0.0pt}%
\section{Experiments}
\vspace{0.0pt}%
\label{sec:experiment_comparison}

We compare the proposed methods with the following models: VAE, RAE~\cite{ghosh2019from}, WAE-MMD~\cite{tolstikhin2018wasserstein} and plain autoencoder (AE).
The generation quality is evaluated using Fr\'{e}chet Inception Distance (FID) ~\cite{heusel2017gans} on MNIST, CelebA~\cite{liu2015deep} and CelebAHQ~\cite{karras2018progressive} datasets with the default train/test split.
{\color{crevise}Here, the FID is defined as}
\begin{align}
    \color{crevise}
    \mathrm{FID}=\|\mu_1-\mu_2\|_2^2+\trace\left(\Sigma_1+\Sigma_2-2(\Sigma_1\Sigma_2)^\frac{1}{2}\right),
\end{align}
{\color{crevise}where $(\mu_1,\Sigma_1)$ and $(\mu_2,\Sigma_2)$ denote the mean and covariance of the Inception vectors from two target distributions, respectively.}
{\color{crevise}Examples of reconstructed and generated images are shown in \ref{sec:samples}.}

Regarding the prior--posterior mismatch, three approaches are tested on all the models. The first approach follows the conventional case, which simply samples the latent variables from the prior. 
The other two approaches are applied after the ordinary training. The second approach forms an aggregated posterior $q_{\bm{\phi}}(\mathbf{z})$ by a second-stage VAE~\cite{dai2019diagnosing}. The third approach uses a Gaussian mixture model (GMM) with 10 to 100 components~\cite{ghosh2019from} to fit the posterior.
The baseline is the standard VAE. Two methods of determining $\sigma_x^2$ are tested: (i) $\sigma_x^2$ is fixed to 1.0 as in common implementations; and (ii) $\sigma_x^2$ is learned as an usual trainable parameter~\cite{dai2019diagnosing}.
{\color{crevise}Besides the conventional Gaussian setup, we also included Bernoulli distribution and MoL~\cite{salimans2017pixelcnn++} as the decoder distribution of VAE.}

RAE is included in the comparison due to its similarity to the proposed method as mentioned in Section~\ref{sec:related_works}.
The objective function of RAE-GP is equivalent to Eq.~\beqref{eq:ELBO_linear_approx} except for that the weighting parameters are determined manually. 
WAE-MMD is included as a representative example of the implicit regularization. The kernel of WAE-MMD used here is a 7-scale inverse multi-quadratic kernel, which is the same as~\cite{tolstikhin2018wasserstein}.

\vspace{0.0pt}%
\subsection{Image generation on MNIST and CelebA}
\vspace{0.0pt}%
\label{sec:experiment_mnist_celeba}

In this experiment, the latent space dimensions for MNIST and CelebA were set to $d_z=16$ and $64$, respectively. This is to be consistent with \cite{ghosh2019from}. A common network architecture, which is adopted from~\cite{chen2016infogan} and described in \ref{sec:detail_exp_setup}, is used for all models.
In Table~\ref{tb:main_results}, we report the evaluation result of each method as: (i) the MSE of the reconstructed test data and (ii) the FID of the generated images. Since the MSE is no longer the reconstruction loss for the proposed methods (Diag-I, Iso-D and Diag-D) except for Iso-I, it is expected to see some MSE degradation. In the case of sampling $\mathbf{z}\in\mathcal{Z}$ from the prior, WAE achieved a low FID due to its relatively strong regularization of the aggregated posterior with MMD.
{\color{crevise}The training time per epoch of each method is shown in Figure~\ref{fig:training_time_per_epoch}. It shows that training proposed parameterizations do not take significant more time than other conventional methods.}

The proposed method with Diag-D parameterization achieved the best FID score on the MNIST dataset. All the proposed parameterizations show at least competitive performance on both datasets. It should be noted that the learned $\sigma_x^2$ values on MNIST and CelebA from Iso-I are $0.0037$ and $0.0040$, respectively, which are both much smaller than $1.0$. However, we do aware if the extra parameters introduced by proposed parameterizations induced extra local optima and thus degraded the stability of optimization in some cases. This would be investigated in future works.
{\color{crevise}Examples of reconstructed and generated images are shown in Figures~\ref{fig:samples_mnist} and \ref{fig:samples_celeba}.
The t-SNE visualization of MNIST latent spaces learned by all the methods is shown in Figure~\ref{fig:tsne_comp}.
}


It can be observed from Table~\ref{tb:main_results} that different parameterizations of variances can affect FID scores greatly. In order to clearly observe the advantage of estimating $\bm{\Sigma}_x$ by MLE rather than estimating it like usual trainable parameters, we examined the two approaches with the four parameterizations (Iso-I, Iso-D, Diag-I and Diag-D): (i) solve Eq.~\beqref{eq:objective_general_model} with MLE as in  Eq.~\beqref{eq:cost_model_general}; (ii) simply treat $\bm{\Sigma}_x$ as a trainable parameter like \cite{dai2019diagnosing}.
The comparative result can be obtained from the bottom eight rows of Table~\ref{tb:main_results}. It shows that applying MLE improves FID scores in most of the cases.

\vspace{0.0pt}%
\subsection{Interpolation on MNIST and CelebA}
\vspace{0.0pt}%

In addition, we empirically verified that the latent spaces learned by the proposed methods are also feasible for downstream tasks such as latent interpolation.
If high quality images can be generated by interpolating the latent variables in a latent space, the corresponding latent space is more likely to be applicable to other downstream tasks. 
Therefore, we evaluate the FID scores for the images generated by latent variable interpolation. We choose 10,000 random pairs of images from both MNIST and CelebA datasets for this experiment.

The interpolation is done by first applying spherical interpolation~\cite{ghosh2019from} in latent spaces and then generating the interpolated images with the decoders. In the end, we evaluate the FID of these interpolated images.

Furthermore, we include different mixing ratio setups in the experiment: (i) a fixed ratio of 0.5, i.e., the mid-point of two latent variables; and (ii) a uniformly distributed random ratio between $[0,1]$ for each image pair. The results are shown in Table~\ref{tb:main_interpolation}, where the proposed method achieved the best score on MNIST and is competitive on CelebA. This suggests that a generative model trained with proposed methods can not only have proper smoothness but also being feasible for downstream tasks such as the interpolation.
{\color{crevise}Examples of images generated by the interpolation are shown in Figure~\ref{fig:interpolations}}

\vspace{0.0pt}%
\subsection{\color{crevise}Image generation on CelebAHQ}
\vspace{0.0pt}%

{\color{crevise}In order to test the performance of proposed parameterization on larger scale images, we performed an experiment similar to the one in section \ref{sec:experiment_mnist_celeba} on CelebAHQ 128x128. We adapt the same architectures of encoder and decoder as in the CelebA experiment. The main difference of this experiment is the size of images and the dimension of the latent space; we set the dimension $d_z$ as $256$.
The result is shown in Table~\ref{tb:celebahq_results}. The proposed method with Diag-D parameterization achieved the best FID if the \nth{2}VAE or GMM is employed to mitigate the prior--posterior mismatch. Moreover, applying \nth{2}VAE or GMM also improves the FID of other models except for the ordinary VAE. We suspect that this is due to the prior--posterior mismatch becomes more prominent in large scale datasets. In this case, it is crucial to have such extra treatments as suggested in \cite{ghosh2019from}.
Examples of reconstructed and generated images are shown in Figure~\ref{fig:samples_celebahq}}.

\vspace{0.0pt}%
\subsection{\color{crevise}Estimation of the negative log-likelihood}
\vspace{0.0pt}%

{\color{crevise}We further estimate the negative log-likelihood (NLL) for proposed models on MNIST, CelebA and CelebAHQ via the $k$-sample importance weighting with $k=500$~\cite{burda2015importance}. As a baseline we estimate the NLL of VAE with $\sigma_x^2=1.0$. The results is shown in Table~\ref{tb:proposed_nll}. For all the datasets, the proposed models outperform the baseline. Among the proposed models, the one with Diag-D parameterization achieved the lowest NLL, which justifies the effectiveness of using a more flexible parameterization.}

\vspace{0.0pt}%
\section{Conclusion}
\vspace{0.0pt}%
\label{sec:conclusion}
In this work, the importance of the variance parameter in VAE training is investigated. The variance parameter determines the weighting between the terms of the objective function and regularizes the smoothness of the decoder. We proposed several parameterizations with a MLE-based self-adaptation scheme without introducing extra hyperparameters. This stabilizes the training of VAE on datasets that have non-isotropic data variances and thus avoids oversmoothing the decoder. Empirically, we have shown that the estimated variance parameter is sufficiently close to the data variance and therefore prevents the model from posterior collapse due to oversmoothness. An extra pass of posterior estimation is applied to deal with the prior--posterior mismatch which is a result of the regularization.

In the end, the evaluation result shows that the models trained by the proposed method have competitive generation quality compared to state-of-the-art results while maintaining the feasibility of the latent space for downstream tasks such as interpolation. In the future, we will investigate more powerful parameterizations such as extending the parameterization into the full matrix case and find proper ways to stabilize its training.




\bibliographystyle{elsarticle-num}
\bibliography{str_def_abrv,refs_dgm,refs_ml}

\newpage
\appendix

\vspace{0.0pt}
\section{Proof of Theorem~\ref{th:collapses}}
\vspace{0.0pt}
\label{sec:collapses}

Let $\mathbf{x}$ be the input sample. We denote its corresponding latent space vector as $\mathbf{z}$ and the reconstructed sample as $\mathbf{x}^\prime$. We have the following relation:
\begin{align}
  \mathcal{I}(\mathbf{x};\mathbf{z})\geq\mathcal{I}(\mathbf{x};\mathbf{x}^\prime),
  \label{eq:ineq_mi_x_z_1}
\end{align}
which can be proved similarly to the proof of Lemma~\ref{th:mi_inequation} in \ref{sec:convergence_var}.
On the other hand, $\mathcal{I}(\mathbf{x};\mathbf{z})$ can be evaluated by using the definition of the MI as
\begin{align}
  \mathcal{I}(\mathbf{x};\mathbf{z})
  =& \KL(\tilde{p}_{\text{data}}(x)q_{\bm{\phi}}(\mathbf{z}|\mathbf{x})\parallel \tilde{p}_{\text{data}}(x)q_{\bm{\phi}}(\mathbf{z}))\notag\\
  =& \E_{\tilde{p}_{\text{data}}(x)q_{\bm{\phi}}(\mathbf{z}|\mathbf{x})}[\ln q_{\bm{\phi}}(\mathbf{z}|\mathbf{x})-\ln q_{\bm{\phi}}(\mathbf{z})]\notag\\
  =& \E_{\tilde{p}_{\text{data}}(x)}\KL(q_{\bm{\phi}}(\mathbf{z}|\mathbf{x})\parallel p(\mathbf{z}))-\KL(q_{\bm{\phi}}(\mathbf{z})\parallel p(\mathbf{z}))\label{eq:decom_mi_x_z}\\
  \leq& \E_{\tilde{p}_{\text{data}}(x)}\KL(q_{\bm{\phi}}(\mathbf{z}|\mathbf{x})\parallel p(\mathbf{z})),
  \label{eq:ineq_mi_x_z_2}
\end{align}
where $\mathcal{I}(\mathbf{x};\mathbf{z})$, $\KL(q_{\bm{\phi}}(\mathbf{z}|\mathbf{x})\parallel p(\mathbf{z}))$ and $\KL(q_{\bm{\phi}}(\mathbf{z})\parallel p(\mathbf{z}))$ are all non-negative.
Inequalities~\beqref{eq:ineq_mi_x_z_1} and \beqref{eq:ineq_mi_x_z_2} lead to the proof.

\vspace{0.0pt}
\section{Linear approximation of the ELBO-based objective $\mathcal{J}_{\sigma_{{x}}^2}$}
\vspace{0.0pt}
\label{sec:linear_approx_objective}

We start with parameterizing the encoder while following the assumption in Eq.~\beqref{eq:enc_and_dec_vae}.
Given a sufficiently small perturbation with  $p(\bm{\epsilon}_{{z}})=\mathcal{N}(\bm{\epsilon}_{{z}}|\mathbf{0},\diag(\bm{\sigma}_{\phi}^2(\mathbf{x})))$, the linear approximation of $\bm{\mu}_{\theta}(\cdot)$ at $\bm{\mu}_{\phi}(\mathbf{x})$ can be represented as
\begin{align}
  \bm{\mu}_{\theta}(\bm{\mu}_{\phi}(\mathbf{x})+\bm{\epsilon}_{{z}})
  = \bm{\mu}_{\theta}(\bm{\mu}_{\phi}(\mathbf{x})) + J_{\bm{\mu}_{\theta}}(\bm{\mu}_{\phi}(\mathbf{x}))\bm{\epsilon}_{{z}},
  \label{eq:linear_approx_decoder}
\end{align}
where $J_{\bm{\mu}_{\theta}}(\bm{\mu}_{\phi}(\mathbf{x}))$ represents the Jacobian matrix of $\bm{\mu}_{\theta}(\mathbf{z})$ at $\mathbf{z}=\bm{\mu}_{\phi}(\mathbf{x})$.
Substituting Eq.~\beqref{eq:linear_approx_decoder} into Eq.~\beqref{eq:loss_gaussian_vae} leads to
\begin{align}
  \E_{q_{\bm{\phi}}(\mathbf{z}|\mathbf{x})}[\|\mathbf{x}-\bm{\mu}_{\theta}(\mathbf{z})\|_2^2]
  &= \E_{p(\bm{\epsilon}_{{z}})}\left[\left\|\mathbf{x}-(\bm{\mu}_{\theta}(\bm{\mu}_{\phi}(\mathbf{x})) + J_{\bm{\mu}_{\theta}}(\bm{\mu}_{\phi}(\mathbf{x}))\bm{\epsilon}_{{z}})\right\|_2^2\right]\notag\\
  &= \left\|\mathbf{x}-\bm{\mu}_{\theta}(\bm{\mu}_{\phi}(\mathbf{x}))\right\|_2^2
  + \E_{p(\bm{\epsilon}_{{z}})}\left[\bm{\epsilon}_{{z}}^\top \mathbf{G}_J\bm{\epsilon}_{{z}}\right]\notag\\
  &\qquad+ \underbrace{\E_{p(\bm{\epsilon}_{{z}})}\left[(\mathbf{x}-\bm{\mu}_{\theta}(\bm{\mu}_{\phi}(\mathbf{x})))^\top J_{\bm{\mu}_{\theta}}(\bm{\mu}_{\phi}(\mathbf{x}))\bm{\epsilon}_{{z}}\right]}_{=0},
  \label{eq:linear_approx_elbo_full}
\end{align}
where $\mathbf{G}_{J}$ denotes the Gram matrix $J_{\bm{\mu}_{\theta}}(\bm{\mu}_{\phi}(\mathbf{x}))^\top J_{\bm{\mu}_{\theta}}(\bm{\mu}_{\phi}(\mathbf{x}))$.
Note that the last term in Eq.~\beqref{eq:linear_approx_elbo_full} is zero under the assumption that the perturbation is sufficiently small.
The expectation in the second right-hand-side term can be evaluated as
\begin{align}
  \E_{p(\bm{\epsilon}_{{z}})}\left[\bm{\epsilon}_{{z}}^\top \mathbf{G}_J\bm{\epsilon}_{{z}}\right]
  &= \trace\left(\E_{p(\bm{\epsilon}_{{z}})}\left[\bm{\epsilon}_{{z}}\bm{\epsilon}_{{z}}^\top\right]\mathbf{G}_J\right)\notag\\
  &= \trace\left(\diag(\bm{\sigma}_{\phi}^2(\mathbf{x}))\mathbf{G}_J\right)\notag\\
  &= \sum_{i=1}^{d_{{x}}}\sum_{j=1}^{d_{{z}}}\sigma_{\phi,j}^2(\mathbf{x})\left(\left.\frac{\partial \mu_{\theta,i}(\mathbf{z})}{\partial z_j}\right|_{z=\bm{\mu}_{\phi}(\mathbf{z})}\right)^2,
\end{align}
which can be interpreted as the gradient penalty for the decoder weighted by $\bm{\sigma}_{\phi}^2(\mathbf{x})$.
By substituting the above result into Eq.~\beqref{eq:loss_gaussian_vae}, its linear approximation can be obtained as
\begin{align}
  \tilde{\Js}_{\sigma_{{x}}^2}(\theta,\phi)
  &\approx \frac{1}{2\sigma_{{x}}^2}\E_{\tilde{p}_{\text{data}}(\mathbf{x})}\Biggl[
  \left\|\mathbf{x}-\bm{\mu}_{\theta}(\bm{\mu}_{\phi}(\mathbf{x}))\right\|_2^2\notag\\
  &\qquad+\sum_{i=1}^{d_{{x}}}\sum_{j=1}^{d_{{z}}}\sigma_{\phi,j}^2(\mathbf{x})\left(\left.\frac{\partial \mu_{\theta,i}(\mathbf{z})}{\partial z_j}\right|_{\mathbf{z}=\bm{\mu}_{\phi}(\mathbf{x})}\right)^2
  +2\sigma_{{x}}^2\|\bm{\mu}_{\phi}(\mathbf{x})\|_2^2\Biggr].
  \label{eq:linear_approximation_full}
\end{align}
In the case of the simplified parameterization described in Section~\ref{sec:major_problem}, the second right-hand-side term in Eq.~\beqref{eq:linear_approximation_full} can be further reduced to  $\sigma_{{z}}^2\|\nabla\bm{\mu}_{\theta}(\bm{\mu}_{\phi}(\mathbf{x}))\|_F^2$. In the simplified case, the perturbation follows a multivariate i.i.d. Gaussian distribution, $p(\bm{\epsilon}_{{z}})=\mathcal{N}(\bm{\epsilon}_{{z}}|\mathbf{0},\sigma_{{z}}^2 \mathbf{I})$. Under this assumption, we have
\begin{align}
  \E_{p(\bm{\epsilon}_{{z}})}\left[\bm{\epsilon}_{{z}}^\top \mathbf{G}_J \bm{\epsilon}_{{z}}\right]
  &= \E_{\mathcal{N}(\bm{\epsilon}_{{z}}|\mathbf{0},\sigma_{{z}}^2 \mathbf{I})}\left[\bm{\epsilon}_{{z}}^\top \left(\sum_{i=1}^{d_{{z}}}\lambda_i\mathbf{u}_i(\mathbf{x})\mathbf{u}_i(\mathbf{x})^\top\right) \bm{\epsilon}_{{z}}\right]\nonumber\\
  &= \sum_{i=1}^{d_{{z}}}\lambda_i\mathbf{u}_i(\mathbf{x})^\top\E_{\mathcal{N}(\bm{\epsilon}_{{z}}|\mathbf{0},\sigma_{{z}}^2 \mathbf{I})}[\bm{\epsilon}_{{z}}\bm{\epsilon}_{{z}}^\top] \mathbf{u}_i(\mathbf{x})\nonumber\\
  &=\sigma_{{z}}^2\sum_{i=1}^{d_{{z}}}\lambda_i,
  \label{eq:perturb_simple}
\end{align}
where $\lambda_i$ is the $i$th eigenvalue of $\mathbf{G}_J$, which is a symmetrical positive definite matrix, and the corresponding eigenvectors are $(\mathbf{u}_i(\mathbf{x}))_{i=1}^{d_{{z}}}$.
Following the simplified assumption,
the second right-hand-side term in Eq.~\beqref{eq:linear_approx_elbo_full} now becomes $\E_{\mathcal{N}(\bm{\epsilon}_{{z}}|\mathbf{0},\sigma_{{z}}^2\mathbf{I})}\left[\bm{\epsilon}_{{z}}^\top \mathbf{G}_J\bm{\epsilon}_{{z}}\right]$. Combining Eq.~\beqref{eq:perturb_simple} and the fact that $\sum_{i=1}^{d_z}\lambda_i = \trace(\mathbf{G}_J) = \|\nabla\bm{\mu}_{\theta}(\bm{\mu}_{\phi}(\mathbf{x}))\|_2^2$, we can finally obtain the following linear approximation for the simplified parameterization:
\begin{align}
  \E_{\mathcal{N}(\mathbf{z}|\mathbf{0},\sigma_{{z}}^2\mathbf{I})}\left[\bm{\epsilon}_{{z}}^\top \mathbf{G}_J\bm{\epsilon}_{{z}}\right]
  &= \sigma_{{z}}^2\sum_{i=1}^{d_{{x}}}\sum_{j=1}^{d_{{z}}}\left(\left.\frac{\partial \mu_{\theta,i}(\mathbf{z})}{\partial z_j}\right|_{\mathbf{z}=\bm{\mu}_{\phi}(\mathbf{z})}\right)^2\notag\\
  &= \sigma_{{z}}^2\left\|\nabla\bm{\mu}_{\theta}(\bm{\mu}_{\phi}(\mathbf{x}))\right\|_F^2.
\end{align}

\vspace{0.0pt}
\section{Expected local smoothness of decoder}
\vspace{0.0pt}
\label{sec:smoothness_decoder}

Here, we describe the relation between the expected local smoothness $\E_{\tilde{p}_{\text{data}}(\mathbf{x})}[\|\nabla\bm{\mu}_{\theta}(\bm{\mu}_{\phi}(\mathbf{x}))\|_F^2]$ and the expected gap $\Delta^2(s_{{z}}^2)$.
First, consider the relation
\begin{align}
  \Delta^2(\mathbf{x},\bm{\epsilon}_{{z}},\bm{\epsilon}_{{z}}^\prime)
  &:=\|\bm{\mu}_{\theta}(\bm{\mu}_{\phi}(\mathbf{x})+{\bm{\epsilon}_{{z}}})-\bm{\mu}_{\theta}(\bm{\mu}_{\phi}(\mathbf{x})+{\bm{\epsilon}_{{z}}^\prime})\|_2^2\notag\\
  &=K_{\theta}(\bm{\mu}_{\phi}(\mathbf{x}),{\bm{\epsilon}_{{z}}},{\bm{\epsilon}_{{z}}^\prime})^2\|{\bm{\epsilon}_{{z}}}-{\bm{\epsilon}_{{z}}^\prime}\|_2^2,
  \label{eq:lipschtz_like}
\end{align}
with the perturbation $\bm{\epsilon}_{{z}}$ following the Gaussian distribution $\mathcal{N}(\bm{\epsilon}_{{z}}|\mathbf{0},s_{{z}}^2\mathbf{I})$.
Applying the expectation operator to Eq.~\beqref{eq:lipschtz_like} leads to
\begin{align}
  \Delta^2(\mathbf{x},\bm{\epsilon}_{{z}},\bm{\epsilon}_{{z}}^\prime)
  &=\E_{p({\bm{\epsilon}_{{z}}}, {\bm{\epsilon}_{{z}}^\prime})}\left[K_{\theta}(\bm{\mu}_{\phi}(\mathbf{x}),{\bm{\epsilon}_{{z}}},{\bm{\epsilon}_{{z}}^\prime})^2\|{\bm{\epsilon}_{{z}}}-{\bm{\epsilon}_{{z}}^\prime}\|_2^2\right]\label{eq:rel_pair_x_z}\\
  &\leq\E_{p({\bm{\epsilon}_{{z}}}, {\bm{\epsilon}_{{z}}^\prime})}\left[K_{\theta}(\bm{\mu}_{\phi}(\mathbf{x}),{\bm{\epsilon}_{{z}}},{\bm{\epsilon}_{{z}}^\prime})^2\right]\E_{p({\bm{\epsilon}_{{z}}}, {\bm{\epsilon}_{{z}}^\prime})}\left[\|{\bm{\epsilon}_{{z}}}-{\bm{\epsilon}_{{z}}^\prime}\|_2^2\right]\label{eq:ineq_pair_x_z}\\
  &=: 2K_{\theta}^2(\bm{\mu}_{\phi}(\mathbf{x}),s_{{z}}^2)d_{{z}}s_{{z}}^2,
  \label{eq:rel_K_eg}
\end{align}
where $p({\bm{\epsilon}_{{z}}}, {\bm{\epsilon}_{{z}}^\prime}):=\mathcal{N}(\bm{\epsilon}_{{z}}|\mathbf{0},s_{{z}}^2\mathbf{I})\mathcal{N}(\bm{\epsilon}_{{z}}^\prime|\mathbf{0},s_{{z}}^2\mathbf{I})$, $\bm{\epsilon}_{{z}}-\bm{\epsilon}_{{z}}^\prime\sim\mathcal{N}(\bm{\epsilon}_{{z}}-\bm{\epsilon}_{{z}}^\prime|\mathbf{0},2s_{{z}}^2\mathbf{I})$ and $K_{\theta}^2(\bm{\mu}_{\phi}(\mathbf{x})):=\E_{p({\bm{\epsilon}_{{z}}}, {\bm{\epsilon}_{{z}}^\prime})}\left[K_{\theta}(\bm{\mu}_{\phi}(\mathbf{x}),{\bm{\epsilon}_{{z}}},{\bm{\epsilon}_{{z}}^\prime})^2\right]$.
Note that in Eq.~\beqref{eq:ineq_pair_x_z}, we assume that $K_{\theta}^2(\bm{\mu}_{\phi}(\mathbf{x}))$ is independent of $\bm{\epsilon}_{{z}}$ and $\bm{\epsilon}_{{z}}^\prime$.
Consider the case that the variance $s_{{z}}^2$ is sufficiently small to approximate $\bm{\mu}_{\theta}(\mathbf{z})$ linearly around $\mathbf{z}=\bm{\mu}_{\phi}(\mathbf{x})$, which is perturbed with variance $s_{{z}}^2$.
In such a case, $K_{\theta}(\bm{\mu}_{\phi}(\mathbf{x}),{\bm{\epsilon}_{{z}}},{\bm{\epsilon}_{{z}}^\prime})$ is independent of ${\bm{\epsilon}_{{z}}}$ and ${\bm{\epsilon}_{{z}}^\prime}$, which fits the assumption in \beqref{eq:ineq_pair_x_z}.
Under this local linearity assumption, $K_{\theta}^2(\bm{\mu}_{\phi}(\mathbf{x}))$ is bounded as
\begin{align}
  K_{\theta}^2(\bm{\mu}_{\phi}(\mathbf{x}),s_{{z}}^2)\leq K_{\theta}^2,
\end{align}
where $K_{\theta}$ denotes the Lipschitz constant of the decoder.

Following the assumption, $K_{\theta}^2(\bm{\mu}_{\phi}(\mathbf{x}),s_{{z}}^2)$ can be formulated by invoking Eq.~\beqref{eq:linear_approx_decoder} as
\begin{align}
  K_{\theta}^2(\bm{\mu}_{\phi}(\mathbf{x}),s_{{z}}^2)
  &=\frac{\E_{p({\bm{\epsilon}_{{z}}}, {\bm{\epsilon}_{{z}}^\prime})}[\|\bm{\mu}_{\theta}(\bm{\mu}_{\phi}(\mathbf{x})+{\bm{\epsilon}_{{z}}})-\bm{\mu}_{\theta}(\bm{\mu}_{\phi}(\mathbf{x})+{\bm{\epsilon}_{{z}}^\prime})\|_2^2]}
  {\E_{p({\bm{\epsilon}_{{z}}}, {\bm{\epsilon}_{{z}}^\prime})}[\|{\bm{\epsilon}_{{z}}}-{\bm{\epsilon}_{{z}}^\prime}\|_2^2]}\notag\\
  &=\frac{\E_{p(\bm{\epsilon}_{{z}},\bm{\epsilon}_{{z}}^\prime)}[(\bm{\epsilon}_{{z}}-\bm{\epsilon}_{{z}}^\prime)^\top \mathbf{G}_J (\bm{\epsilon}_{{z}}-\bm{\epsilon}_{{z}}^\prime)]}{2d_{{z}}s_{{z}}^2}\notag\\
  &=\frac{\trace\left(\E_{p(\bm{\epsilon}_{{z}},\bm{\epsilon}_{{z}}^\prime)}[(\bm{\epsilon}_{{z}}-\bm{\epsilon}_{{z}}^\prime)(\bm{\epsilon}_{{z}}-\bm{\epsilon}_{{z}}^\prime)^\top] \mathbf{G}_J\right)}{2d_{{z}}s_{{z}}^2}\notag\\
  &=\frac{\trace\left( \mathbf{G}_J\right)}{d_{{z}}}.
  \label{eq:ELT_linear_approx}
\end{align}
Applying the expectation operator to Eq.~\beqref{eq:ELT_linear_approx} leads to
\begin{align}
  K_{\theta}^2(s_{{z}}^2)
  &:=\E_{\tilde{p}_{\text{data}}}\left[K_{\theta}^2(\bm{\mu}_{\phi}(\mathbf{x}),s_{{z}}^2)\right]\notag\\
  &=\frac{\E_{\tilde{p}_{\text{data}}(\mathbf{x})}\left[\trace\left(\mathbf{G}_J\right)\right]}{d_{{z}}}\notag\\
  &= \frac{\E_{\tilde{p}_{\text{data}}}\left[\left\|\nabla\bm{\mu}_{\theta}(\bm{\mu}_{\phi}(\mathbf{x}))\right\|_F^2\right]}{d_{{z}}}.
  \label{eq:rel_K_els}
\end{align}

Finally, combining Eqs.~\beqref{eq:rel_K_eg} and \beqref{eq:rel_K_els} yields the following connection between the expected gap and the expected local smoothness:
\begin{align}
  \Delta^2(s_{{z}}^2)=2\E_{\tilde{p}_{\text{data}}}\left[\left\|\nabla\bm{\mu}_{\theta}(\bm{\mu}_{\phi}(\mathbf{x}))\right\|_F^2\right] s_{{z}}^2.
\end{align}

\vspace{0.0pt}
\section{Experimental details for Section~\ref{sec:exp_analysis_VAE}}
\vspace{0.0pt}
\label{sec:detail_exp_analysis}

\vspace{0.0pt}
\subsection{Experimental setup}
\vspace{0.0pt}
\label{sec:detail_exp_analysis_setup}

In the experiment, the model is trained with the Adam optimizer with a learning rate of $10^{-3}$. The dimension of the latent space is set to 8. We run 200 epochs with a minibatch size of $64$ for all $\sigma_{{x}}^2$.
We use the following DNN architectures for the encoder and decoder, respectively:
\begin{flalign*}
  x\in\mathbb{R}^{28\times28}
  &\to\mathrm{Conv}_{64}\to\mathrm{ReLU}&\text{size: }(64,14,14)\\
  &\to\mathrm{Conv}_{128}\to\mathrm{ReLU}\to\mathrm{Reshape}\\
  &\to\mathrm{Flatten}\to\mathrm{FC}_{1024}\to\mathrm{ReLU}\\
  &\to\mathrm{FC}_{16},\\
  z\in\mathbb{R}^{16}
  &\to\mathrm{FC}_{1024}\to\mathrm{ReLU}\\
  &\to\mathrm{FC}_{128\times7\times7}\to\mathrm{ReLU}&\text{size: }(128,7,7)\\
  &\to\mathrm{ConvT}_{64}\to\mathrm{ReLU}&\text{size: }(64,14,14)\\
  &\to\mathrm{ConvT}_{1}\to\mathrm{Sigmoid}&\text{size: }(1,28,28).
\end{flalign*}
Here, $\mathrm{FC}_{k}$, $\mathrm{Conv}_{k}$, $\mathrm{ConvT}_{k}$ and $\mathrm{ReLU}$ denote the fully connected layer mapping to $\mathbb{R}^{k}$, the convolutional layer mapping to $k$ channels, the transpose convolutional layer mapping to $k$ channels and the rectified linear units (ReLU), respectively.
The 3-tuple $(\mathrm{channels},\mathrm{height},\mathrm{width})$ in the right column represents the output shape of each layer.
In all the $\mathrm{Conv}_{k}$ and $\mathrm{ConvT}_{k}$ layers, $4\times4$ convolutional filters are used with a common stride of $(2,2)$.

Regarding the evaluation of criteria, MSE and KL are evaluated on the training set because the aim of the experiment is to validate the relation between $\sigma_{{z}}^2$ and the smoothness of the decoder.
The upper bound of the MI is obtained by first calculating
\begin{align}
  -\E_{q_{\bm{\phi},\sigma_{{z}}^2}(\mathbf{z})}\left[\ln\E_{\tilde{p}_{\text{data}}(\mathbf{x}^\prime)}\exp\left(-\frac{\|\mathbf{z}-\bm{\mu}_{\phi}(\mathbf{x}^\prime)\|_2^2}{2\sigma_{{z}}^2}\right)\right]-\frac{d_{{x}}}{2}
\end{align}
for each minibatch then taking the average, where $q_{\bm{\phi},\sigma_{{z}}^2}(\mathbf{z}):=\E_{\tilde{p}_{\text{data}}(\mathbf{x})}[q_{\bm{\phi},\sigma_{{z}}^2}(\mathbf{z}|\mathbf{x})]$ . The batch size is $10,000$ for all the evaluations.

\vspace{0.0pt}
\subsection{Samples of generated images and t-SNE visualization of latent spaces}
\vspace{0.0pt}
\label{sec:detail_exp_analysis_vis}

\begin{figure}[t]
   \centering
   \includegraphics[width=.96\textwidth]{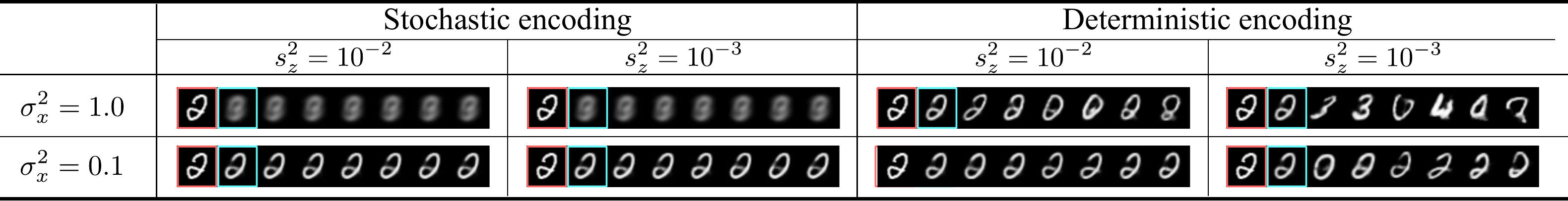}
   \caption{Images in red boxes are the original images sampled from the MNIST dataset. The images in blue boxes are reconstructed by $\bm{\mu}_{\theta}(\bm{\mu}_{\phi}(\mathbf{x}))$. The other images are decoded from neighbor points of $\bm{\mu}_{\phi}(\mathbf{x})$, which are perturbed by $\bm{\epsilon}_{{z}}\sim\mathcal{N}(\bm{\epsilon}_{{z}}|\mathbf{0},s_{{z}}^2\mathbf{I})$.}
   \label{fig:perturbation}
 \end{figure}

Figure~\ref{fig:perturbation} shows several images decoded from $\bm{\mu}_{\phi}(\mathbf{x})+\bm{\epsilon}_{{z}}$ with $\bm{\epsilon}_{{z}}\sim\mathcal{N}(\bm{\epsilon}_{{z}}|\mathbf{0},s_{{z}}^2\mathbf{I})$ for the cases with $\sigma_{{x}}^2=1.0$ and $0.1$.
Posterior collapse can be observed from these blurry images decoded from the \textit{stochastic encoding} case with $\sigma_{{x}}^2=1.0$. This is due to the removal of batch normalization, which makes $\sigma_{{x}}^2=1.0$ become an inappropriate choice. However, if $\sigma_x^2$ is determined or adapted appropriately such as by using the proposed method, posterior collapse will not happen.
In the other settings, the tendency of how the image changes with the perturbation is similar, as shown in Table~\ref{tb:result1}.

\begin{figure}[tb]
  \centering
  \subfloat[$\log\sigma_{{x}}^2=0.0$]{\includegraphics[width=0.250\textwidth]{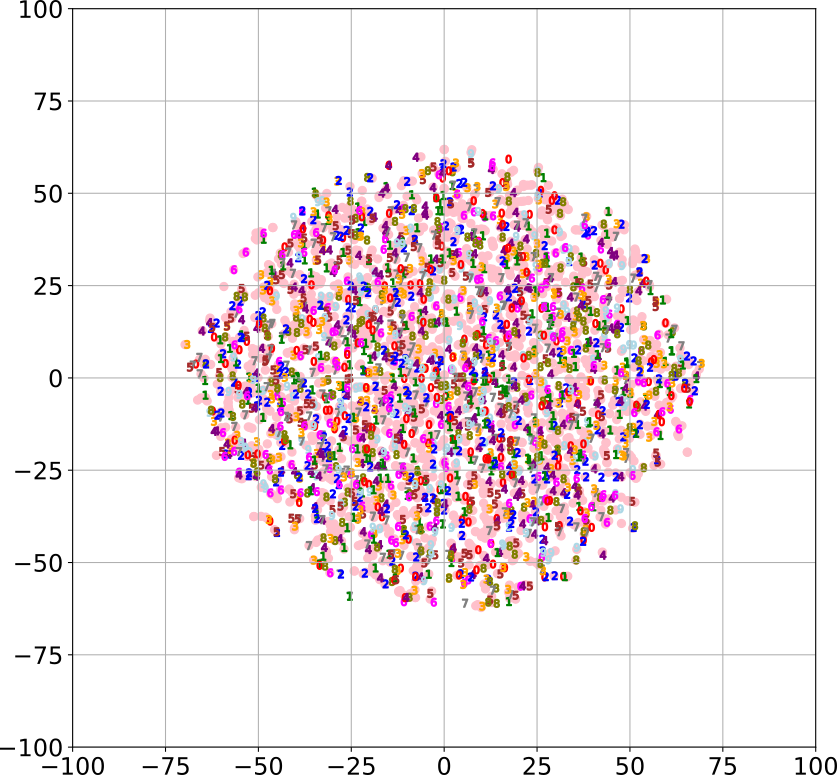}}
  \hspace{20pt}%
  \subfloat[$\log\sigma_{{x}}^2=-0.2$]{\includegraphics[width=0.250\textwidth]{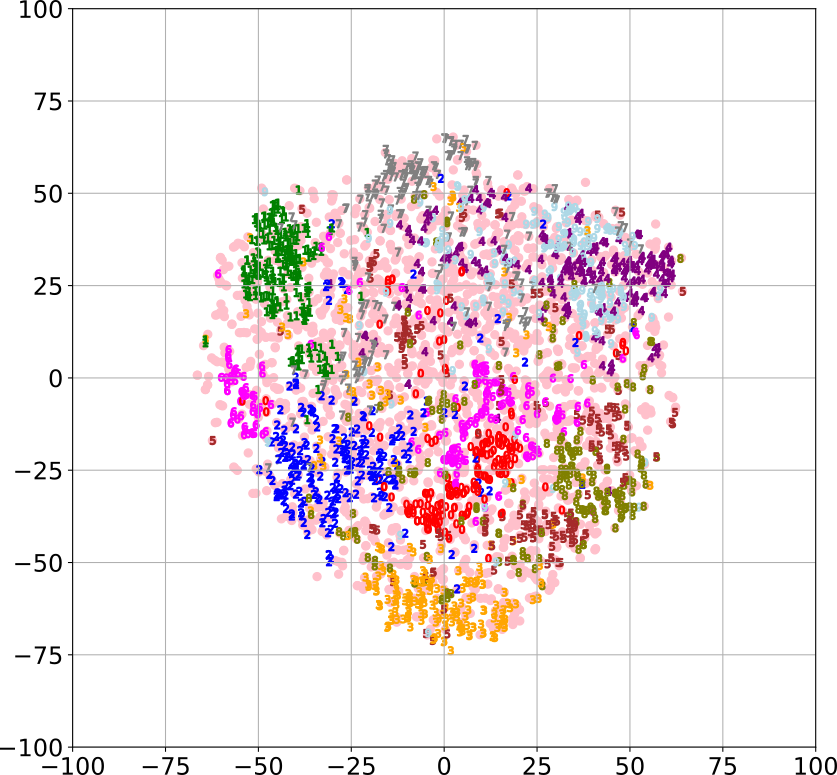}}
  \hspace{20pt}%
  \subfloat[$\log\sigma_{{x}}^2=-0.4$]{\includegraphics[width=0.250\textwidth]{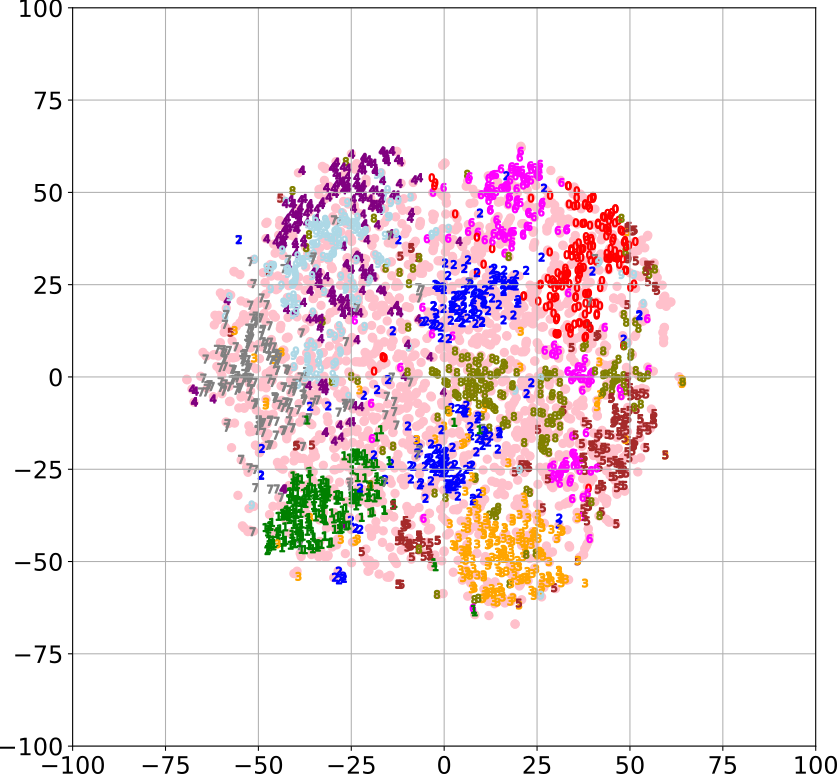}}\\
  \subfloat[$\log\sigma_{{x}}^2=-0.6$]{\includegraphics[width=0.250\textwidth]{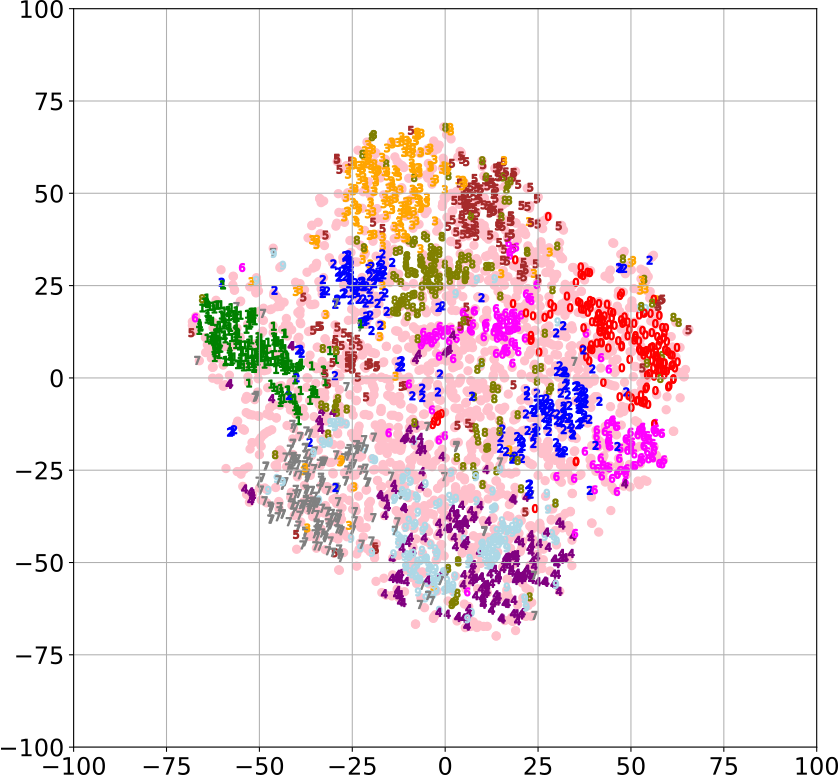}}
  \hspace{20pt}%
  \subfloat[$\log\sigma_{{x}}^2=-0.8$]{\includegraphics[width=0.250\textwidth]{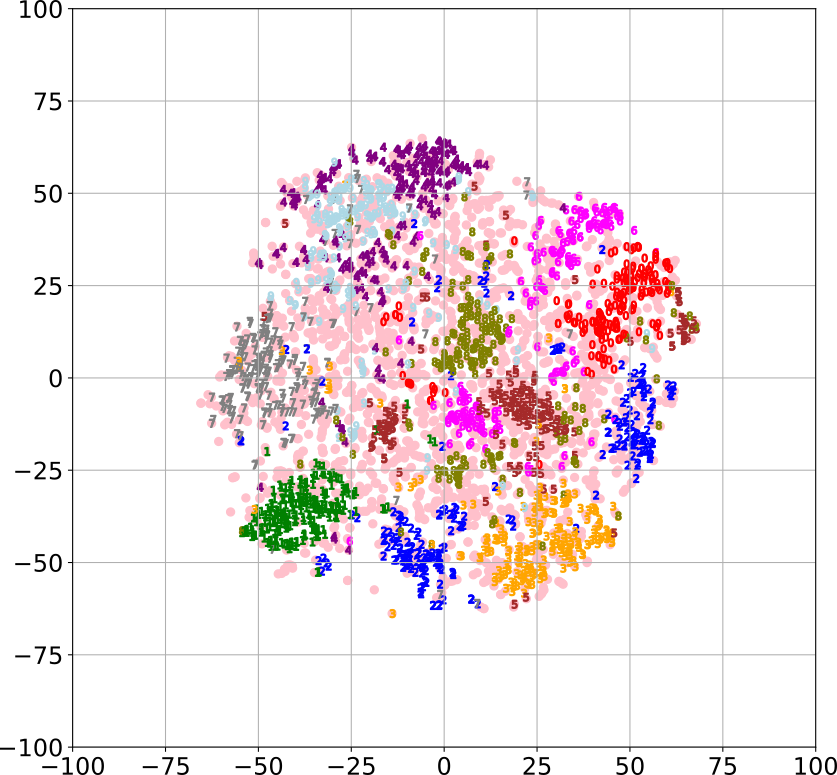}}
  \hspace{20pt}%
  \subfloat[$\log\sigma_{{x}}^2=-1.0$]{\includegraphics[width=0.250\textwidth]{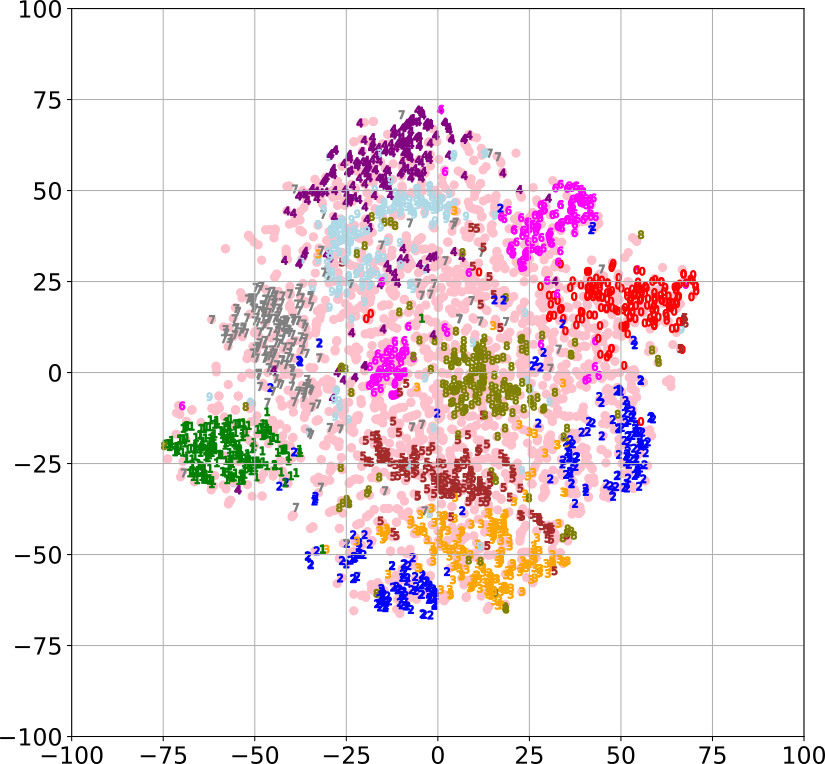}}
  \caption{Visualization of latent space via t-SNE. Pink dots are sampling points generated from the prior $p(\mathbf{z})$.}
  \label{fig:sect3_CX-VAE_tsne}
\end{figure}

The latent spaces are also visualized via t-SNE~\cite{maaten2008visualizing} in Figure \ref{fig:sect3_CX-VAE_tsne}.
The dots with different colors represent the latent vectors encoded from images of different labels (numbers), and the pink dots are the sampling points generated from the prior $p(\mathbf{z})$.
As mentioned earlier, to observe the effect of $\sigma_x^2$ clearly, we remove batch normalization, which usually helps prevent posterior collapse to a certain extent. As a result, the latent space with $\sigma_{{x}}^2=1.0$ completely collapses and $q_{\bm{\phi}}(z)$ approaches $p(z)$ as shown in Figures~\ref{fig:perturbation} and \ref{fig:sect3_CX-VAE_tsne}(a). In this case, both KL collapse and  posterior collapse occur.

\vspace{0.0pt}
\section{Fixing the posterior variance of latent space}
\vspace{0.0pt}
\label{sec:fix_varz}
\begin{table}[th!]
\caption{Evaluation of various criteria for different $\sigma_z^2$. These criteria are the expected value of $\|\mathbf{x}^\prime-\mathbf{x}\|_2^2$ (MSE), KL divergence, the upper bound of MI $\mathcal{I}(\mathbf{x}^\prime,\mathbf{z})$, the expected gap (the perturbation variance $s_z^2$ is set to $10^{-2}$ and $10^{-3}$) and expected local smoothness (ELS).}
  \centering
  \footnotesize
  \begin{tabular}{r|rrrrrr}
    \bhline{0.8pt}
    \multirow{2}{*}{$\log\sigma_{{z}}^2$} & \multirow{2}{*}{MSE} & \multirow{2}{*}{KL} & \multirow{2}{*}{MI} & \multicolumn{2}{c}{Expected gap} & \multirow{2}{*}{ELS} \\ \cline{5-6}
                                      &                      &                     &                     & $10^{-2}$      & $10^{-3}$       & \\
    \bhline{0.8pt}
    $1.0$                             & \textbf{52.74}       &  \textbf{26.79}     & \textbf{8.0e-3}     & 6.20e-6        & 6.15e-7         & \textbf{3.81e-5} \\
    $0.9$                             & \textbf{52.74}       &  \textbf{19.48}     & \textbf{6.8e-3}     & 7.19e-6        & 7.11e-7         & \textbf{4.42e-5}       \\
    $0.8$                             & 22.96                &  134.83             & 8.5e+1              & 2.03e-2        & 2.03e-3         & 1.27e-1       \\
    $0.7$                             & 20.50                &  139.09             & 1.0e+2              & 2.37e-2        & 2.37e-3         & 1.48e-1       \\
    $0.6$                             & 19.30                &  132.61             & 1.2e+2              & 2.87e-2        & 2.87e-3         & 1.80e-1       \\
    $0.5$                             & 17.38                &  132.05             & 1.7e+2              & 3.44e-2        & 3.44e-3         & 2.16e-1       \\
    $0.4$                             & 15.91                &  128.96             & 1.9e+2              & 3.94e-2        & 3.94e-3         & 2.47e-1       \\
    $0.3$                             & 14.63                &  125.97             & 2.3e+2              & 4.50e-2        & 4.50e-3         & 2.82e-1       \\
    $0.2$                             & 13.50                &  122.07             & 2.8e+2              & 5.16e-2        & 5.16e-3         & 3.23e-1       \\
    $0.1$                             & 12.40                &  118.89             & 3.1e+2              & 5.83e-2        & 5.81e-3         & 3.67e-1       \\
    $0.0$                             & 11.71                &  112.50             & 3.6e+2              & 6.79e-2        & 6.80e-3         & 4.26e-1       \\
    $-0.1$                            & 10.97                &  107.32             & 4.0e+2              & 7.53e-2        & 7.55e-3         & 4.74e-1       \\
    $-0.2$                            & 10.23                &  103.87             & 4.2e+2              & 8.69e-2        & 8.70e-3         & 5.46e-1       \\
    $-0.3$                            &  9.63                &   98.48             & 4.6e+2              & 9.84e-2        & 9.88e-3         & 6.18e-1       \\
    $-0.4$                            &  9.12                &   93.86             & 5.2e+2              & 1.12e-1        & 1.12e-2         & 7.05e-1       \\
    $-0.5$                            &  8.71                &   88.35             & 5.2e+2              & 1.25e-1        & 1.26e-2         & 7.88e-1       \\
    $-0.6$                            &  8.26                &   83.68             & 5.9e+2              & 1.42e-1        & 1.43e-2         & 8.94e-1       \\
    $-0.7$                            &  7.82                &   79.70             & 6.6e+2              & 1.62e-1        & 1.62e-2         & 1.02          \\
    $-0.8$                            &  7.55                &   74.75             & 7.1e+2              & 1.80e-1        & 1.80e-2         & 1.13          \\
    $-0.9$                            &  7.26                &   70.63             & 7.3e+2              & 2.04e-1        & 2.05e-2         & 1.29          \\
    $-1.0$                            &  7.05                &   66.14             & 7.6e+2              & 2.28e-1           & 2.30e-2         & 1.45          \\
    \bhline{0.8pt}
  \end{tabular}\label{tb:fixed_z}
\end{table}
\begin{figure}[t]
  \centering
  \subfloat[$\log\sigma_{{z}}^2=1.0$]{\includegraphics[width=0.250\textwidth]{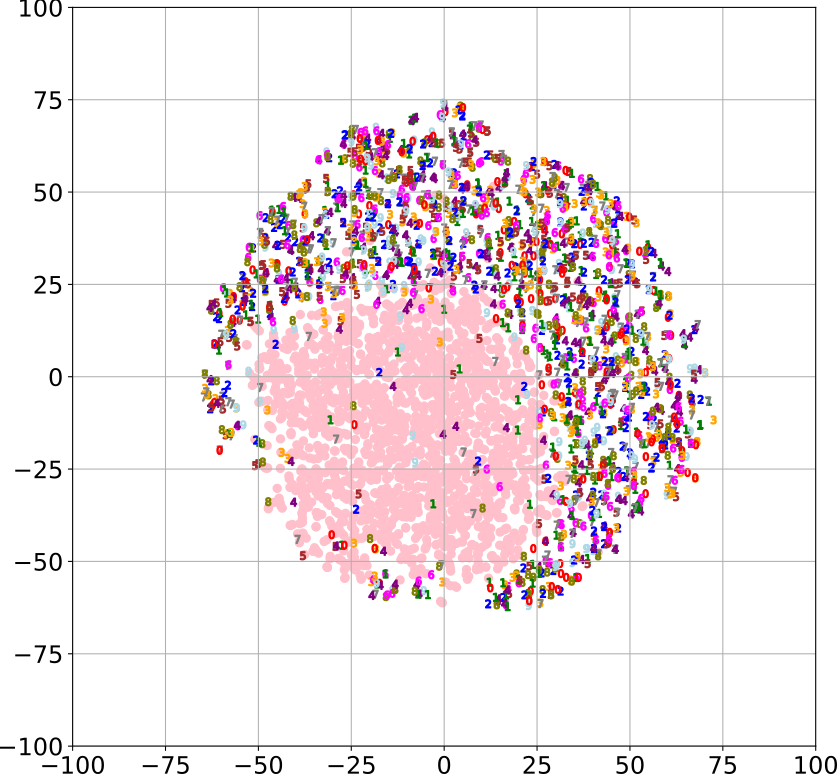}}
  \hspace{20pt}%
  \subfloat[$\log\sigma_{{z}}^2=0.6$]{\includegraphics[width=0.250\textwidth]{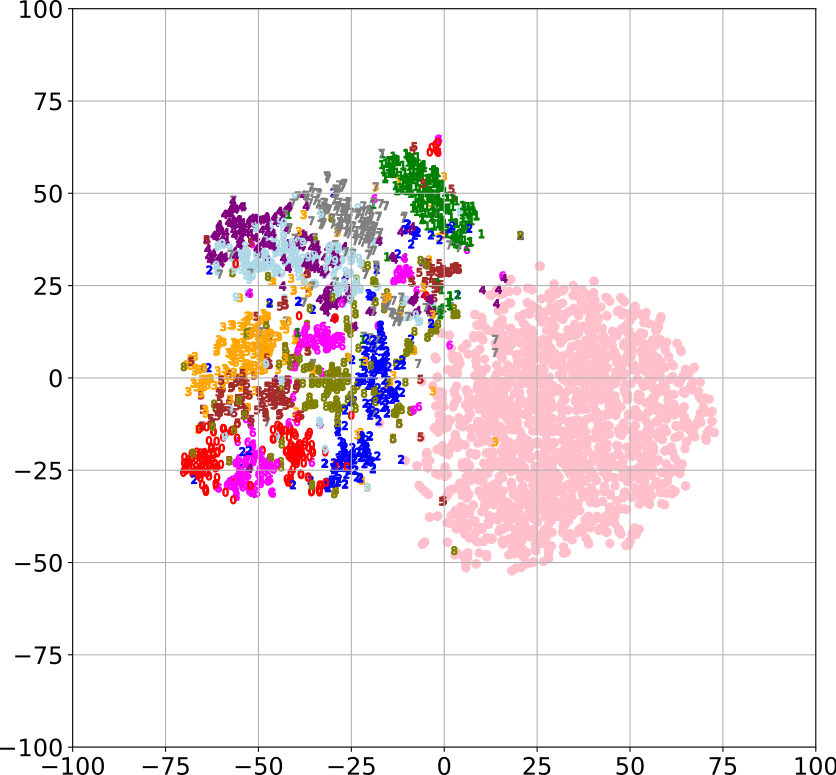}}
  \hspace{20pt}%
  \subfloat[$\log\sigma_{{z}}^2=0.2$]{\includegraphics[width=0.250\textwidth]{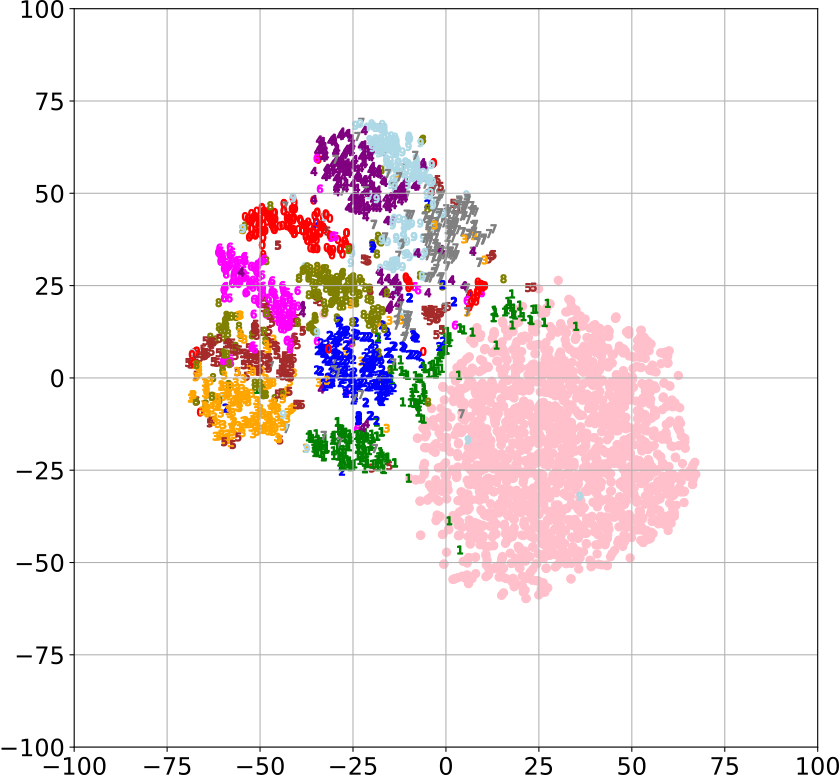}}\\
  \subfloat[$\log\sigma_{{z}}^2=-0.2$]{\includegraphics[width=0.250\textwidth]{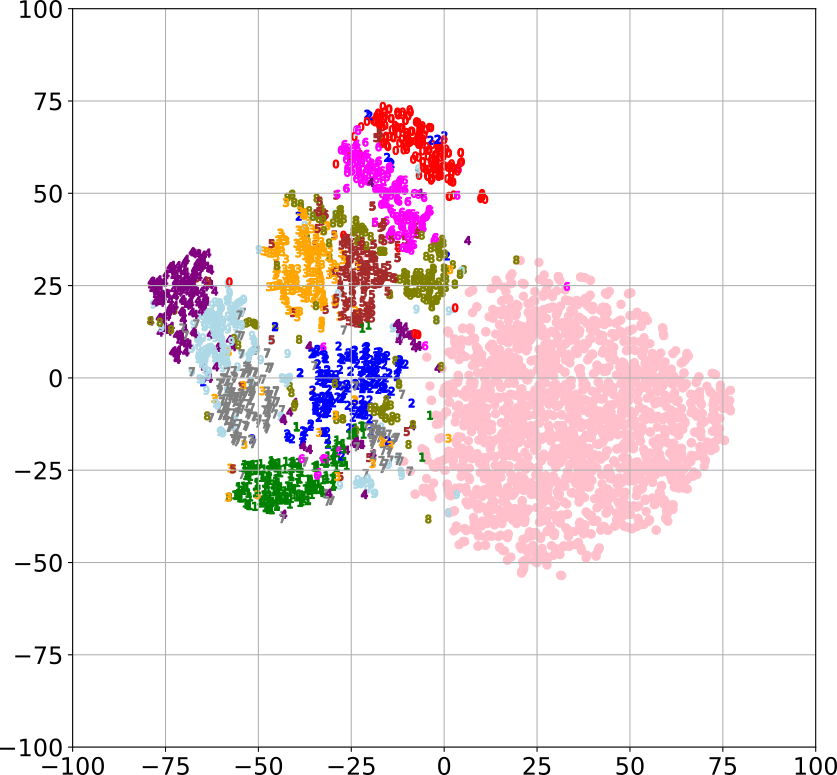}}
  \hspace{20pt}%
  \subfloat[$\log\sigma_{{z}}^2=-0.6$]{\includegraphics[width=0.250\textwidth]{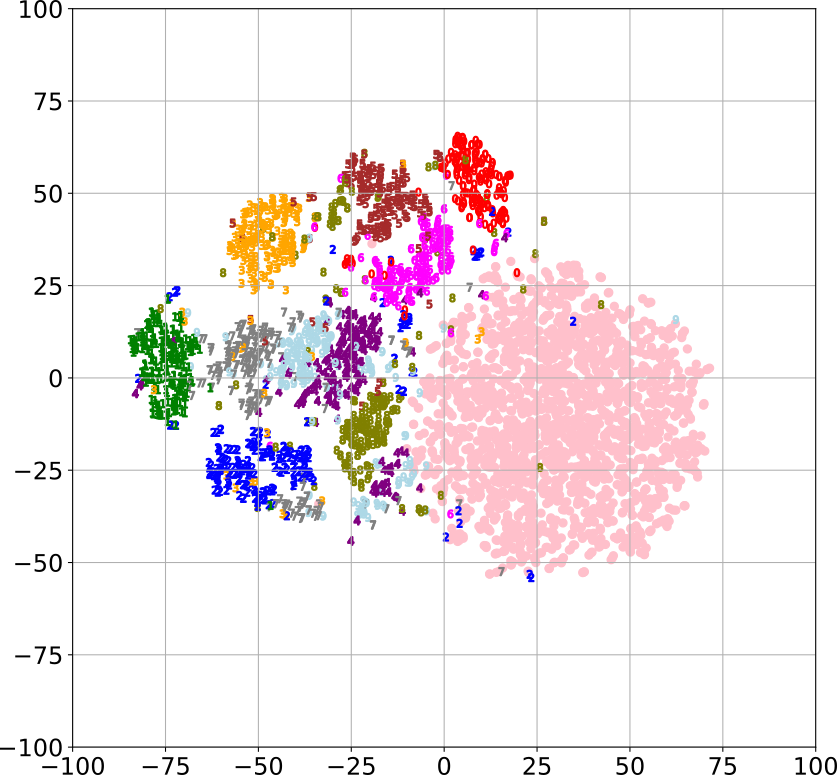}}
  \hspace{20pt}%
  \subfloat[$\log\sigma_{{z}}^2=-1.0$]{\includegraphics[width=0.250\textwidth]{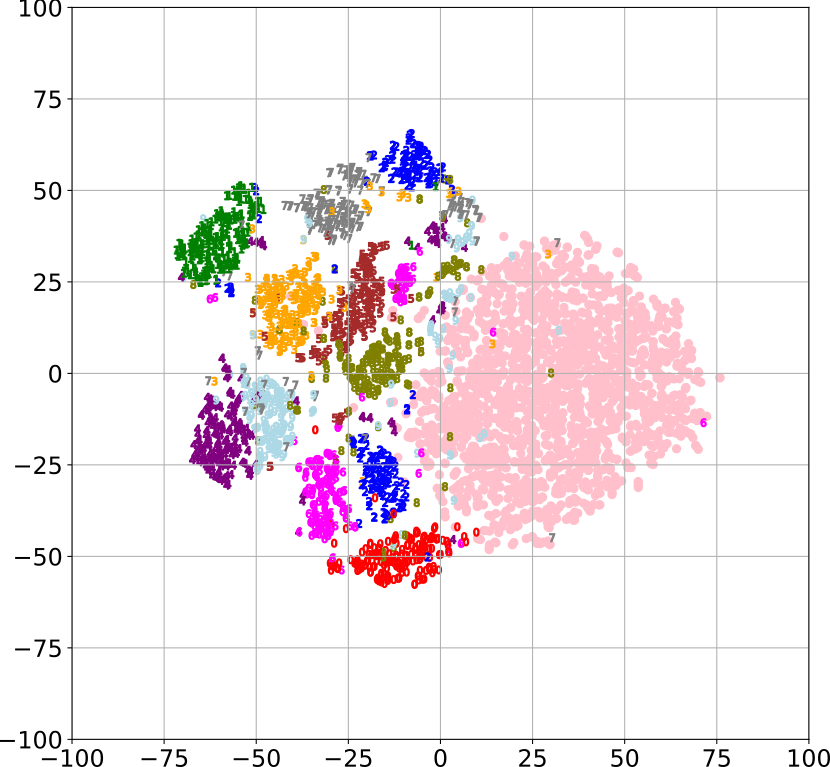}}
  \caption{Visualization of latent space via t-SNE. Pink dots are sampling points generated from the prior $p(\mathbf{z})$.}
  \label{fig:sect3_CZ-VAE_tsne}
\end{figure}
From the previous sections, we know that $\sigma_x^2$ affects the smoothness via $\sigma_x^2$. However, it would be interesting to see what will happen if $\sigma_z^2$ is fixed while $\sigma_x^2$ is optimized. In this experiment, the variance parameter $\sigma_{{z}}^2$ is fixed while $\sigma_{{x}}^2$ is optimized with the proposed MLE~\beqref{eq:cost_model_general} under the parameterization in Section~\ref{sec:major_problem}.
The other settings remain the same as those in Section~\ref{sec:exp_analysis_VAE}.
We evaluate the numerical results for different $\sigma_z^2$ with the criteria listed in Section~\ref{sec:exp_analysis_VAE}.
According to Table \ref{tb:fixed_z}, the tendencies of the expected gap and ELS show that a large $\sigma_z^2$ makes the decoder smoother, which is consistent with the discussion in Section~\ref{sec:local_optim_ELBO}.
However, the tendency of the KL divergence is different from that in Section~\ref{sec:exp_analysis_VAE}.
Although a larger $\sigma_z^2$ consistently leads to a smaller MI, and eventually the MI collapses to zero; the KL divergence still remains far from zero, which means that posterior collapse can happen without KL collapse.
This phenomenon can be visually confirmed by observing the t-SNE plot in Figure~\ref{fig:sect3_CZ-VAE_tsne}.
The cause of this phenomenon can be roughly reasoned from the linear approximated ELBO~\beqref{eq:ELBO_linear_approx}, in which $\sigma_z^2$ directly affects the gradient penalty and causes oversmoothness.
It should be pointed out that the strength of $L_2$ regularization in Eq.~\beqref{eq:ELBO_linear_approx} is gradually decreased with decreasing $\sigma_x^2$; therefore, it does not dominate the whole objective function. As a result, the mean of the approximated posterior $q_{\bm{\phi}}(z)$ is far from the mean of the prior $p(z)$ (which is $\mathbf{0}$), and therefore $\KL(q_{\bm{\phi}}(z)\parallel p(z))$ in Eq.~\beqref{eq:decom_mi_x_z} does not diminish to zero.

\vspace{0.0pt}
\section{Proof of Theorem \ref{th:convergence_var}}
\vspace{0.0pt}
\label{sec:convergence_var}

According to Theorem 4 in~\cite{dai2019diagnosing}, we know that
\begin{align}
  \lim_{\sigma_{{x}}^2\to0}\E_{p_{\text{data}}(\mathbf{x})q_{\bm{\phi},\sigma_{{z}}^2}(\mathbf{z}|\mathbf{x})}\left[\|\mathbf{x}-\bm{\mu}_{\theta}(\mathbf{z})\|_2^2\right]=0,
\end{align}
which also leads to $\hat{\sigma}_{{x}}^2\to0~(\sigma_{{x}}^2\to0)$.
Here, $\hat{\sigma}_{{x}}^2$ is estimated through MLE and is given by
\begin{align}
  \hat{\sigma}_{{x}}^2=\frac{1}{d_{{x}}}\E_{\tilde{p}_{\text{data}}(\mathbf{x})q_{\bm{\phi},\sigma_{{z}}^2}(\mathbf{z}|\mathbf{x})}\left[\|\mathbf{x}-\bm{\mu}_{\theta}(\mathbf{z})\|_2^2\right].
\end{align}
To prove Theorem \ref{th:convergence_var}, we need the following auxiliary theorem:
\begin{theorem}
  \label{th:convergence_var_2}
  In the training stage of VAE, we have $\sigma_{{z}}^2\to0~(\hat{\sigma}_{{x}}^2\to0)$.
\end{theorem}
First, we state three lemmas with proofs.
\begin{lemma}
  \label{th:mi_inequation}
  In a VAE, $\mathcal{I}(\mathbf{x},\mathbf{x}^\prime)\leq\mathcal{I}(\mathbf{z},\mathbf{z}_{\text{e}})$ always holds, where $\mathbf{z}_{\text{e}}$ is the encoded latent variable $\mathbf{z}_{\text{e}}=\bm{\mu}_{\phi}(\mathbf{x})$ with $x\sim p_{\text{data}}(\mathbf{x})$.
\end{lemma}
\begin{proof}
  The data processing flow of the VAE is $\mathbf{x}\to \mathbf{z}_{\text{e}}\to \mathbf{z}\to \mathbf{x}^\prime$; $\mathbf{z}_{\text{e}}=\bm{\mu}_{\phi}(\mathbf{x})$, $\mathbf{z}=\mathbf{z}_{\text{e}}+\bm{\epsilon}_{{z}}$, and $\mathbf{x}^\prime=\bm{\mu}_{\theta}(\mathbf{z})$, where $\bm{\epsilon}_{{z}}\sim\mathcal{N}(\bm{\epsilon}_{{z}}|\mathbf{0},\sigma_{{z}}^2\mathbf{I})$.
  The MI $\mathcal{I}(\mathbf{x};\mathbf{z},\mathbf{x}^\prime)$ can be represented as
  \begin{align}
    \mathcal{I}(\mathbf{x};\mathbf{z},\mathbf{x}^\prime)
    &=\mathcal{I}(\mathbf{x};\mathbf{x}^\prime)+\mathcal{I}(\mathbf{x};\mathbf{z}|\mathbf{x}^\prime)\\
    &=\mathcal{I}(\mathbf{x};\mathbf{z})+\mathcal{I}(\mathbf{x};\mathbf{x}^\prime|\mathbf{z}).
  \end{align}
  Since $\mathbf{x}$ and $\mathbf{x}^\prime$ are conditionally independent on the given $\mathbf{z}$, it follows that  $\mathcal{I}(\mathbf{x};\mathbf{x}^\prime|\mathbf{z})=0$.
  From the non-negativity of MI, we have $\mathcal{I}(\mathbf{x};\mathbf{z})\geq\mathcal{I}(\mathbf{x};\mathbf{x}^\prime)$.
  Repeating the same procedure for $\mathcal{I}(\mathbf{x};\mathbf{z}_{\text{e}},\mathbf{z})$ leads to the proof.
\end{proof}

\begin{lemma}
  \label{th:mi_x_reconstructed_x}
  The MI between
  $\mathbf{x}$ and $\mathbf{x}^\prime$ diverges to positive infinity as $\sigma_{{x}}^2\to0$, where $\mathbf{x}^\prime$ is obtained from $\mathbf{x}\sim p_{\text{data}}(\mathbf{x})$ as $\mathbf{x}^\prime=\bm{\mu}_{\theta}(\bm{\mu}_{\phi}(\mathbf{x})+\bm{\epsilon}_{{z}})$.
\end{lemma}
\begin{proof}
  A lower bound of $\mathcal{I}(\mathbf{x};\mathbf{x}^\prime)$ is
  \begin{align}
    \mathcal{I}(\mathbf{x};\mathbf{x}^\prime)
    &=\KL\left(p_{\text{data}}(\mathbf{x})p_{\theta,\phi}(\mathbf{x}^\prime|\mathbf{x})\parallel p_{\text{data}}(\mathbf{x})p_{\theta,\phi}(\mathbf{x}^\prime)\right)\notag\\
    &= \E_{p_{\text{data}}(\mathbf{x})p_{\theta,\phi}(\mathbf{x}^\prime|\mathbf{x})}\left[\ln p_{\theta,\phi}(\mathbf{x}^\prime|\mathbf{x})-\ln p_{\theta,\phi}(\mathbf{x}^\prime)\right]\notag\\
    &= \mathcal{H}\left[p_{\theta,\phi}(\mathbf{x}^\prime)\right] - \E_{p_{\text{data}}(\mathbf{x})}\mathcal{H}\left[p_{\theta,\phi}(\mathbf{x}^\prime|\mathbf{x})\right]\notag\\
    &\geq \mathcal{H}\left[p_{\theta,\phi}(\mathbf{x}^\prime)\right] - \E_{p_{\text{data}}(\mathbf{x})}H(\hat{\sigma}_{{x}}^2\mathbf{I}),
    \label{eq:mi_inequality}
  \end{align}
  where $p_{\theta,\phi}(\mathbf{x}^\prime|\mathbf{x}):=\E_{q_{\bm{\phi}}(\mathbf{z}|\mathbf{x})}[p_{\theta}(\mathbf{x}^\prime|\mathbf{z})]$ and $p_{\theta,\phi}(\mathbf{x}^\prime):=\E_{p_{\text{data}}(\mathbf{x})q_{\bm{\phi}}(\mathbf{z}|\mathbf{x})}[p_{\theta}(\mathbf{x}^\prime|\mathbf{z})]$.
  Here, we denote the differential entropy of the Gaussian with variance $\hat{\sigma}_{{x}}^2\mathbf{I}$ as
  \begin{align}
    H(\hat{\sigma}_{{x}}^2\mathbf{I}):=\frac{1}{2}\ln(2\pi e\hat{\sigma}_{{x}}^{2d_{{x}}}).
  \end{align}
  Since $\hat{\sigma}_{{x}}^2\to0$ as $\sigma_{{x}}^2\to0$,  $\mathcal{H}[p_{\theta,\phi}(\mathbf{x}^\prime)]\to\mathcal{H}[p_{\text{data}}(\mathbf{x})]$ and $H(\hat{\sigma}_{{x}}^2)\to-\infty$ in the inequality of \beqref{eq:mi_inequality}.
  Therefore, $\mathcal{I}(\mathbf{x};\mathbf{x}^\prime)\to\infty$ as $\sigma_{{x}}^2\to0$.
\end{proof}

\begin{lemma}
  \label{th:bounded_variance}
  Suppose $\bm{\Sigma}_{\bm{\phi},\sigma_{{z}}^2}$ be the variance of $q_{\bm{\phi},\sigma_{{z}}^2}(\mathbf{z})$.
  If $\pdata(\mathbf{x})$ has finite covariance and $\bm{\mu}_{\bm{\phi}}$ is Lipschitz continuous, then $\det(\bm{\Sigma}_{\bm{\phi},\sigma_{{z}}^2})<\infty$.
\end{lemma}
\begin{proof}
  Let $V := \sigma_{\mathrm{max}}(\mathrm{var}_{\pdata(\mathbf{x})}(\mathbf{x}))$ and $L$ be the Lipschitz constant of $\bm{\mu}_{\bm{\phi}}$ where $\sigma_{\mathrm{max}}$ denotes the maximum singular value. Since $\pdata(\mathbf{x})$ has finite covariance, we know that  $\det(\mathrm{Var}_{\pdata(\mathbf{x})}(\mathbf{x}))<\infty$.

  Consider a probability distribution $\tilde{q}_{\bm{\phi}}(\mathbf{z})$ that  $\mathbf{z}=\bm{\mu}_{\bm{\phi}}(\mathbf{x})$ follows with $\mathbf{x}\sim\pdata(\mathbf{x})$.
  The variance of $\tilde{q}_{\bm{\phi}}(\mathbf{z})$ is denoted as  $\tilde{\bm{\Sigma}}_{\bm{\phi}}$.
  The randomness imposed by the reparameterization trick of VAE can be described as the convolution of $\mathcal{N}(\mathbf{z}|\mathbf{0},\sigma_z^2\mathbf{I})$ and $\tilde{q}_{\bm{\phi}}(\mathbf{z})$.
  The mean and variance of $q_{\bm{\phi},\sigma_{{z}}^2}(\mathbf{z})$, denoted as $\mathbf{m}_{\bm{\phi},\sigma_{{z}}^2}$ and $\bm{\Sigma}_{\bm{\phi},\sigma_{{z}}^2}$, can be calculated by
  \begin{align}
      \mathbf{m}_{\bm{\phi},\sigma_{{z}}^2}
      &=\int_{\mathbf{z}\in\mathcal{Z}}\int_{\mathbf{z}^\prime\in\mathcal{Z}}\mathbf{z}\mathcal{N}(\mathbf{z}-\mathbf{z}^\prime|\mathbf{0},\sigma_z^2\mathbf{I})\tilde{q}_{\bm{\phi}}(\mathbf{z}^\prime)d\mathbf{z}d\mathbf{z}^\prime\notag\\
      &=\int_{\mathbf{z}\in\mathcal{Z}}\int_{\mathbf{z}^\prime\in\mathcal{Z}}(\mathbf{z}+\mathbf{z}^\prime)\mathcal{N}(\mathbf{z}|\mathbf{0},\sigma_z^2\mathbf{I})\tilde{q}_{\bm{\phi}}(\mathbf{z}^\prime)d\mathbf{z}d\mathbf{z}^\prime\notag\\
      &=\int_{\mathbf{z}\in\mathcal{Z}}\mathbf{z}\tilde{q}_{\bm{\phi}}(\mathbf{z})d\mathbf{z}=:\tilde{\mathbf{m}}_{\bm{\phi}},\\
      \bm{\Sigma}_{\bm{\phi},\sigma_{{z}}^2}
      &=\int_{\mathbf{z}\in\mathcal{Z}}\int_{\mathbf{z}^\prime\in\mathcal{Z}}(\mathbf{z}-\mathbf{m}_{\bm{\phi},\sigma_{{z}}^2})(\mathbf{z}-\mathbf{m}_{\bm{\phi},\sigma_{{z}}^2})^\top\mathcal{N}(\mathbf{z}-\mathbf{z}^\prime|\mathbf{0},\sigma_z^2\mathbf{I})\tilde{q}_{\bm{\phi}}(\mathbf{z}^\prime)d\mathbf{z}d\mathbf{z}^\prime\notag\\
      &=\int_{\mathbf{z}\in\mathcal{Z}}\int_{\mathbf{z}^\prime\in\mathcal{Z}}(\mathbf{z}+\mathbf{z}^\prime-\tilde{\mathbf{m}}_{\bm{\phi}})(\mathbf{z}+\mathbf{z}^\prime-\tilde{\mathbf{m}}_{\bm{\phi}})^\top\mathcal{N}(\mathbf{z}|\mathbf{0},\sigma_z^2\mathbf{I})\tilde{q}_{\bm{\phi}}(\mathbf{z}^\prime)d\mathbf{z}d\mathbf{z}^\prime\notag\\
      &=\sigma_z^2\mathbf{I}+\int_{\mathbf{z}^\prime\in\mathcal{Z}}(\mathbf{z}^\prime-\tilde{\mathbf{m}}_{\bm{\phi}})(\mathbf{z}^\prime-\tilde{\mathbf{m}}_{\bm{\phi}})^\top\tilde{q}_{\bm{\phi}}(\mathbf{z}^\prime)d\mathbf{z}^\prime\notag\\
      &=\sigma_z^2\mathbf{I}+\tilde{\bm{\Sigma}}_{\bm{\phi}}.
      \label{eq:variance_decomposition}
  \end{align}
  From Eq.~\beqref{eq:variance_decomposition} and the fact that $\sigma_{\mathrm{max}}(\tilde{\bm{\Sigma}}_{\bm{\phi}})$ is bounded above by $VL^2$, therefore we have 
  \begin{align}
      \det(\bm{\Sigma}_{\bm{\phi},\sigma_{{z}}^2})
      &=\det(\sigma_z^2\mathbf{I}+\tilde{\bm{\Sigma}}_{\bm{\phi}})\notag\\
      &\leq (\sigma_z^2+VL^2)^{d_z}<\infty.
  \end{align}
\end{proof}
Now we begin to prove Theorem \ref{th:convergence_var}. The MI $\mathcal{I}(\mathbf{z};\mathbf{z}_{\text{e}})$ satisfies
\begin{align}
  \mathcal{I}(\mathbf{z};\mathbf{z}_{\text{e}})
  &=\KL\left(q_{\bm{\phi}}(\mathbf{z}_{\text{e}})q_{\sigma_z^2}(\mathbf{z}|\mathbf{z}_{\text{e}})\parallel q_{\bm{\phi}}(\mathbf{z}_{\text{e}})q_{\bm{\phi},\sigma_z^2}(\mathbf{z})\right)\notag\\
  &= \E_{q_{\bm{\phi}}(\mathbf{z}_{\text{e}})q_{\sigma_z^2}(\mathbf{z}|\mathbf{z}_{\text{e}})}\left[\ln q_{\sigma_z^2}(\mathbf{z}|\mathbf{z}_{\text{e}})-\ln q_{\bm{\phi},\sigma_z^2}(\mathbf{z})\right]\notag\\
  &= \mathcal{H}\left[q_{\bm{\phi},\sigma_z^2}(\mathbf{z})\right] - \E_{q_{\bm{\phi}}(\mathbf{z}_{\text{e}})}\mathcal{H}\left[q_{\sigma_z^2}(\mathbf{z}|\mathbf{z}_{\text{e}})\right]\notag\\
  &\leq H(\bm{\Sigma}_{\bm{\phi},\sigma_{{z}}^2})-H(\sigma_z^2\mathbf{I})\notag\\
  &= \frac{d_{{z}}}{2}\ln\left(\frac{\det(\bm{\Sigma}_{\bm{\phi},\sigma_{{z}}^2})}{\sigma_{{z}}^2}\right),
  \label{eq:mi_ineauality_z}
\end{align}
where $\bm{\Sigma}_{\bm{\phi},\sigma_{{z}}^2}$ denotes the variance of $q_{\bm{\phi},\sigma_{{z}}^2}(\mathbf{z})$.
Invoking Lemma \ref{th:mi_inequation} and \beqref{eq:mi_ineauality_z} leads to
\begin{align}
  \mathcal{I}(\mathbf{x};\mathbf{x}^\prime) \leq \frac{d_{{z}}}{2}\ln\left(\frac{\det(\bm{\Sigma}_{\bm{\phi},\sigma_{{z}}^2})}{\sigma_{{z}}^2}\right).
  \label{eq:mi_ineauality_x}
\end{align}
Now, consider $\sigma_{{z}}^2\not\to0$ as $\sigma_x^2\to0$.
According to Lemma~\ref{th:mi_x_reconstructed_x}, it follows that $\det(\bm{\Sigma}_{\bm{\phi},\sigma_{{z}}^2})\to+\infty$, which contradicts Lemma~\ref{th:bounded_variance}.
Thus, we must have $\sigma_{{z}}^2\to0$ as $\sigma_{{x}}^2$ converges to zero.

\vspace{0.0pt}
\section{Derivation of proposed objectives}
\vspace{0.0pt}
\label{sec:derivation_objectives}
Here, we derive the objectives listed in Table~\ref{tb:models_and_objectives}.
Consider an arbitrary $\bm{\Sigma}_{{x}}$ without any condition. The MLE of $\bm{\Sigma}_{{x}}$, $\hat{\bm{\Sigma}}_{{x}}$, can be obtained by
\begin{align}
  \hat{\bm{\Sigma}}_{{x}}
  =\E_{\tilde{p}_{\text{data}}(\mathbf{x})q_{\bm{\phi}}(\mathbf{z}|\mathbf{x})}\left[(\mathbf{x}-\bm{\mu}_{\theta}(\mathbf{z}))(\mathbf{x}-\bm{\mu}_{\theta}(\mathbf{z}))^\top\right].
\end{align}
From the partial derivative of $\tilde{\mathcal{J}}_{\text{rec}}(\theta,\phi,\bm{\Sigma}_{{x}})$ w.r.t. $\bm{\Sigma}_{{x}}$, we have
\begin{align}
  \frac{\partial\tilde{{\mathcal{J}}}_{\text{rec}}(\theta,\phi,\bm{\Sigma}_{{x}})}{\partial\bm{\Sigma}_{{x}}}
  = \frac{1}{2}\left(\E_{\tilde{p}_{\text{data}}(x)q_{\bm{\phi}}(\mathbf{z}|\mathbf{x})}\left[(\mathbf{x}-\bm{\mu}_{\theta}(\mathbf{z}))(\mathbf{x}-\bm{\mu}_{\theta}(\mathbf{z}))^\top\right]\right.
  +\left.\bm{\Sigma}_{{x}}^{-1}\right).
\end{align}
The MLE of $\hat{\bm{\Sigma}}_{{x}}$ and the objectives for the different parameterizations are described in the following.

\vspace{0.0pt}
\subsection{Iso-I}
\vspace{0.0pt}
\label{sec:derivation_objectives_1}
First, substitute $\bm{\Sigma}_{{x}}=\sigma_{{x}}^2\mathbf{I}$ into Eq.~\beqref{eq:objective_general_model}:
\begin{align}
  \tilde{\Js}_{\text{rec}}(\theta,\phi,{\sigma}_{{x}}^2)
  &= \E_{\tilde{p}_{\text{data}}(\mathbf{x})}\left[\frac{1}{2{\sigma}_{{x}}^2}\E_{q_{\bm{\phi}}(\mathbf{z}|\mathbf{x})}[\left\|\mathbf{x}-\bm{\mu}_{\theta}(\mathbf{z})\right\|_2^2]\right]
  + \frac{d_{{x}}}{2}\ln{\sigma}_{{x}}^2.
  \label{eq:cost_model_ii}
\end{align}
Also, we know that the MLE of $\sigma_{x}^2$ is
\begin{align}
  \hat{\sigma}_{{x}}^2
  = \frac{1}{d_{{x}}}\E_{\tilde{p}_{\text{data}}(\mathbf{x})q_{\bm{\phi}}(\mathbf{z}|\mathbf{x})}\left[\left\|\mathbf{x}-\bm{\mu}_{\theta}(\mathbf{z})\right\|_2^2\right].
  \label{eq:mle_ii}
\end{align}
Substituting Eq.~\beqref{eq:mle_ii} into Eq.~\beqref{eq:cost_model_ii} leads to
\begin{align}
  \tilde{\Js}_{\text{rec}}(\theta,\phi,\hat{\sigma}_{{x}}^2)
  &= \E_{\tilde{p}_{\text{data}}(\mathbf{x})}\left[\frac{1}{2\hat{\sigma}_{{x}}^2}\E_{q_{\bm{\phi}}(\mathbf{z}|\mathbf{x})}[\left\|\mathbf{x}-\bm{\mu}_{\theta}(\mathbf{z})\right\|_2^2]\right]
  + \frac{d_{{x}}}{2}\ln\hat{\sigma}_{{x}}^2\notag\\
  &= \frac{d_{{x}}}{2}
  + \frac{d_{{x}}}{2}\ln\E_{\tilde{p}_{\text{data}}(\mathbf{x})q_{\bm{\phi}}(\mathbf{z}|\mathbf{x})}\left[\left\|\mathbf{x}-\bm{\mu}_{\theta}(\mathbf{z})\right\|_2^2\right]-\frac{d_{{x}}}{2}\ln{d_{{x}}}.
\end{align}

\vspace{0.0pt}
\subsection{Iso-D}
\vspace{0.0pt}
\label{sec:derivation_objectives_2}
First, substitute $\bm{\Sigma}_{{x}}=\sigma_{{x}}^2(\mathbf{x})\mathbf{I}$ into Eq.~\beqref{eq:objective_general_model}:
\begin{align}
  \tilde{\Js}_{\text{rec}}(\theta,\phi,\bm{\Sigma}_{{x}})
  &= \E_{\tilde{p}_{\text{data}}(\mathbf{x})q_{\bm{\phi}}(\mathbf{z}|\mathbf{x})}\left[\frac{1}{2\sigma_{x}^2(\mathbf{x})}\left\|\mathbf{x}-\bm{\mu}_{\theta}(\mathbf{z})\right\|_2^2\right.
  \left.\vphantom{\frac{1}{2\sigma_{x}^2(\mathbf{x})}} +\frac{d_{{x}}}{2}\ln{\sigma}_{x}^2(\mathbf{x})\right].
  \label{eq:cost_model_id}
\end{align}
Also, we know that the MLE of $\sigma_{x}^2(\mathbf{x})$ is
\begin{align}
  \hat{\sigma}_{x}^2(\mathbf{x})
  =\E_{q_{\bm{\phi}}(\mathbf{z}|\mathbf{x})}\left[\frac{1}{d_{{x}}}\left\|\mathbf{x}-\bm{\mu}_{\theta}(\mathbf{z})\right\|_2^2\right].
  \label{eq:mle_id}
\end{align}
Substituting Eq.~\beqref{eq:mle_id} into Eq.~\beqref{eq:cost_model_id} leads to the reconstruction objective of Iso-D:
\begin{align}
  \tilde{\Js}_{\text{rec}}(\theta,\phi,\hat{\bm{\Sigma}}_{{x}})
  &= \E_{\tilde{p}_{\text{data}}(\mathbf{x})q_{\bm{\phi}}(\mathbf{z}|\mathbf{x})}\left[\frac{1}{2\hat{\sigma}_{x}^2(\mathbf{x})}\left\|\mathbf{x}-\bm{\mu}_{\theta}(\mathbf{z})\right\|_2^2\right.
  +\left.\frac{d_{{x}}}{2}\ln\hat{\sigma}_{x}^2(\mathbf{x})\right]\\
  &= \frac{d_{{x}}}{2} + \frac{d_{{x}}}{2}\E_{\tilde{p}_{\text{data}}(\mathbf{x})}\left[\ln\E_{q_{\bm{\phi}}(\mathbf{z}|\mathbf{x})}[\left\|\mathbf{x}-\bm{\mu}_{\theta}(\mathbf{z})\right\|_2^2]\right]
  -\frac{d_{{x}}}{2}\ln{d_{{x}}}.
\end{align}

\vspace{0.0pt}
\subsection{Diag-I}
\vspace{0.0pt}
\label{sec:derivation_objectives_3}
First, substitute $\bm{\Sigma}_{{x}}=\diag(\bm{\sigma}_{{x}}^2)$ into Eq.~\beqref{eq:objective_general_model}:
\begin{align}
  \tilde{\Js}_{\text{rec}}(\theta,\phi,\bm{\Sigma}_{{x}})
  = \E_{\tilde{p}_{\text{data}}(\mathbf{x})}\left[\sum_{i=1}^{d_{{x}}}\frac{1}{2\sigma_{x,i}^2}\E_{q_{\bm{\phi}}(\mathbf{z}|\mathbf{x})}\left[\left(x_{i}-\mu_{\theta,i}(\mathbf{z})\right)^2\right]\right]
  + \sum_{i=1}^{d_{{x}}}\frac{1}{2}\ln{\sigma}_{x,i}^2.
  \label{eq:cost_model_di}
\end{align}
Also, we know that the MLE of $\sigma_{x,i}^2$ is
\begin{align}
  \hat{\sigma}_{x,i}^2
  = \E_{\tilde{p}_{\text{data}}(\mathbf{x})q_{\bm{\phi}}(\mathbf{z}|\mathbf{x})}\left[\left(x_{i}-\mu_{\theta,i}(\mathbf{z})\right)^2\right].
  \label{eq:mle_di}
\end{align}
Substituting Eq.~\beqref{eq:mle_di} into Eq.~\beqref{eq:cost_model_di} leads to the reconstruction objective for Diag-I:
\begin{align}
  \tilde{\Js}_{\text{rec}}(\theta,\phi,\hat{\bm{\Sigma}}_{{x}})
  &= \E_{\tilde{p}_{\text{data}}(\mathbf{x})}\left[\sum_{i=1}^{d_{{x}}}\frac{1}{2\hat{\sigma}_{x,i}^2}\E_{q_{\bm{\phi}}(\mathbf{z}|\mathbf{x})}\left[\left(x_{i}-\mu_{\theta,i}(\mathbf{z})\right)^2\right]\right]
  + \sum_{i=1}^{d_{{x}}}\frac{1}{2}\ln\hat{\sigma}_{x,i}^2\\
  &= \frac{d_{{x}}}{2} + \frac{1}{2}\sum_{i=1}^{d_{{x}}}\ln\E_{\tilde{p}_{\text{data}}(\mathbf{x})q_{\bm{\phi}}(\mathbf{z}|\mathbf{x})}\left[\left(x_{i}-\mu_{\theta,i}(\mathbf{z})\right)^2\right].
\end{align}

\vspace{0.0pt}
\subsection{Diag-D}
\vspace{0.0pt}
\label{sec:derivation_objectives_4}
First, substitute $\bm{\Sigma}_{{x}}=\diag(\bm{\sigma}_{{x}}^2(\mathbf{x}))$ into Eq.~\beqref{eq:objective_general_model}:
\begin{align}
  \tilde{\Js}_{\text{rec}}(\theta,\phi,\bm{\Sigma}_{{x}})
  &= \E_{\tilde{p}_{\text{data}}(\mathbf{x})q_{\bm{\phi}}(\mathbf{z}|\mathbf{x})}\left[\sum_{i=1}^{d_{{x}}}\left(\frac{1}{2\sigma_{x,i}^2(\mathbf{x})}\left(x_{i}-\mu_{\theta,i}(\mathbf{z})\right)^2\right)\vphantom{\sum_{i=1}^{d_{{x}}}}
  +\frac{1}{2}\ln{\sigma}_{x,i}^2(\mathbf{z})\right].
  \label{eq:cost_model_dd}
\end{align}
Also, we know that the MLE of $\sigma_{x,i}^2(\mathbf{x})$ is
\begin{align}
  \hat{\sigma}_{x,i}^2(\mathbf{z})
  = \E_{q_{\bm{\phi}}(\mathbf{z}|\mathbf{x})}\left[(x_{i}-\mu_{\theta,i}(\mathbf{z}))^2)\right].
  \label{eq:mle_dd}
\end{align}
Substituting Eq.~\beqref{eq:mle_dd} into Eq.~\beqref{eq:cost_model_dd} leads to the reconstruction objective for Diag-D:
\begin{align}
  \tilde{\Js}_{\text{rec}}(\theta,\phi,\hat{\bm{\Sigma}}_{{x}})
  &= \E_{\tilde{p}_{\text{data}}(\mathbf{x})q_{\bm{\phi}}(\mathbf{z}|\mathbf{x})}\left[\sum_{i=1}^{d_{{x}}}\left(\frac{1}{2\hat{\sigma}_{x,i}^2(\mathbf{z})}\left(x_{i}-\mu_{\theta,i}(\mathbf{z})\right)^2+\frac{1}{2}\ln{\hat{\sigma}}_{x,i}^2(\mathbf{z})\right)\right]\\
  &= \frac{d_{{x}}}{2} + \frac{1}{2}\sum_{i=1}^{d_{{x}}}\E_{\tilde{p}_{\text{data}}(\mathbf{x})}\left[\ln\E_{q_{\bm{\phi}}(\mathbf{z}|\mathbf{x})}\left(x_{i}-\mu_{\theta,i}(\mathbf{z})\right)^2\right].
\end{align}

\vspace{0.0pt}
\section{Derivation of Eq.~\beqref{eq:vae_cost_two_kl}}
\vspace{0.0pt}
\label{sec:match_pri_pos}

The KL divergence terms of Eq.~\beqref{eq:objective_vae} can be represented as
\begin{align}
    \KL(p_{\text{data}}(\mathbf{x})\parallel p_{\theta}(\mathbf{x}))
    &= \E_{p_{\text{data}}(\mathbf{x})q_{\bm{\phi}}(\mathbf{z}|\mathbf{x})}\left[\ln\frac{p_{\text{data}}(\mathbf{x})}{p_{\theta}(\mathbf{x})}\right],
    \label{eq:appendix_h_1}
\end{align}
where
\begin{align}
    &\E_{p_{\text{data}}(\mathbf{x})}\KL(q_{\bm{\phi}}(\mathbf{z}|\mathbf{x})\parallel p_{\theta}(\mathbf{z}|\mathbf{x}))\notag\\
    &=\E_{p_{\text{data}}(\mathbf{x})q_{\bm{\phi}}(\mathbf{z}|\mathbf{x})}[\ln q_{\bm{\phi}}(\mathbf{z}|\mathbf{x})-\ln p_{\theta}(\mathbf{z}|\mathbf{x})]\nonumber\\
    &=\E_{p_{\text{data}}(\mathbf{x})q_{\bm{\phi}}(\mathbf{z}|\mathbf{x})}\left[\ln\frac{p_{\theta}(\mathbf{x})q_{\bm{\phi}}(\mathbf{z}|\mathbf{x})}{p(\mathbf{z})p_{\theta}(\mathbf{x}|\mathbf{z})}\right]\\
    &=\E_{p_{\text{data}}(\mathbf{x})q_{\bm{\phi}}(\mathbf{z}|\mathbf{x})}\left[\ln\frac{p_{\theta}(\mathbf{x})q_{\bm{\phi}}(\mathbf{z}|\mathbf{x})}{q_{\bm{\phi}}(\mathbf{z})p_{\theta}(\mathbf{x}|\mathbf{z})}+\ln\frac{q_{\bm{\phi}}(\mathbf{z})}{p(\mathbf{z})}\right].
    \label{eq:appendix_h_2}
\end{align}
By substituting the two equations above into Eq.~\beqref{eq:objective_vae}, $\Ls$ can be reformulated into
\begin{align}
    \Ls=\E_{p_{\text{data}}(\mathbf{x})q_{\bm{\phi}}(\mathbf{z}|\mathbf{x})}\left[\ln\frac{p_{\text{data}}(\mathbf{x})q_{\bm{\phi}}(\mathbf{z}|\mathbf{x})}{q_{\bm{\phi}}(\mathbf{z})p_{\theta}(\mathbf{x}|\mathbf{z})}+\ln\frac{q_{\bm{\phi}}(\mathbf{z})}{p(\mathbf{z})}\right],
    \label{eq:appendix_h_3}
\end{align}
which is equivalent to Eq.~\beqref{eq:vae_cost_two_kl}.

\vspace{0.0pt}
\section{Details of experimental setup in Section~\ref{sec:experiment_comparison}}
\vspace{0.0pt}
\label{sec:detail_exp_setup}
In this experiment, the Adam optimizer~\cite{kingma2014adam} is used and the maximum number of epochs is set to $100$ for MNIST and $70$ for CelebA. The learning rates are $0.001$ for MNIST and $0.0002$ for CelebA. A minibatch size of 64 is used.
All the FID\footnote{We used the PyTorch version of the FID implementation from \url{https://github.
com/mseitzer/pytorch-fid} for all the models. However, the result may slightly differ from that obtained with the TensorFlow implementation \url{https://github.
com/bioinf-jku/TTUR}.} values are evaluated with $10,000$ generated samples.

For the posterior estimation by the second-stage VAE, we adopt the same networks for the encoder and decoder as those in \cite{dai2019diagnosing}.
For GMM fitting, we use the same settings as those in \cite{ghosh2019from}.
Experimental details including the network architectures for each dataset are described in the following.

\vspace{0.0pt}%
\subsection{MNIST}
\vspace{0.0pt}%
\label{sec:detail_exp_setup_MNIST}
We construct the encoder and decoder for the MNIST dataset using the architecture in \cite{chen2016infogan}.
The encoder is constructed as
\begin{align*}
  x\in\mathbb{R}^{28\times28}
  &\to\mathrm{Conv}_{64}\to\mathrm{ReLU}&\text{size: }(64,14,14)\\
  &\to\mathrm{Conv}_{128}\to\mathrm{ReLU}\to\mathrm{Reshape}&\text{size: }(128,7,7)\\
  &\to\mathrm{Flatten}\to\mathrm{FC}_{1024}\\
  &\to\mathrm{BN}\to\mathrm{ReLU}\\
  &\to\mathrm{FC}_{16\times 2}.
\end{align*}
The decoder is constructed as
\begin{align*}
  z\in\mathbb{R}^{16}
  &\to\mathrm{FC}_{1024}\to\mathrm{BN}\to\mathrm{ReLU}\\
  &\to\mathrm{FC}_{128\times7\times7}\to\mathrm{BN}\to\mathrm{ReLU}&\text{size: }(128,7,7)\\
  &\to\mathrm{ConvT}_{64}\to\mathrm{BN}\to\mathrm{ReLU}&\text{size: }(64,14,14)\\
  &\to\mathrm{ConvT}_{1}\to\mathrm{Sigmoid}&\text{size of }(1,28,28).
\end{align*}
In all the $\mathrm{Conv}_{k}$ layers and all the $\mathrm{ConvT}_{k}$ layers except for the last, $5\times5$ convolutional filters with stride $(2,2)$ are used.
The difference between this architecture and those used in \ref{sec:detail_exp_analysis_setup} is whether batch normalization is applied or not.
Although in the original work of \cite{chen2016infogan}, the discriminator used leaky ReLU (lReLU), we adopt ReLU for the encoder part, which improves the performance for all the models evenly.

\vspace{0.0pt}%
\subsection{CelebA}
\vspace{0.0pt}%
\label{sec:detail_exp_setup_CelebA}
The CelebA images are preprocessed with center cropping of $140\times 140$, then resized to $64\times 64$ as described in \cite{tolstikhin2018wasserstein} and \cite{ghosh2019from}.
It should be noted that the size of cropping differs among the previous works, and it markedly affects the FID score. We choose the above cropping size as is the largest among the related works and seems to be the most difficult case for image generation. Moreover, this cropping size was used also in \cite{tolstikhin2018wasserstein} and \cite{ghosh2019from}.
Similarly to in the previous section, the encoder and decoder are constructed on the basis of the discriminator and generator for CelebA used in \cite{chen2016infogan}.
The encoder is constructed as
\begin{align*}
  x\in\mathbb{R}^{64\times64}
  &\to\mathrm{Conv}_{128}\to\mathrm{ReLU}&\text{size: }(128,32,32)\\
  &\to\mathrm{Conv}_{256}\to\mathrm{BN}\to\mathrm{ReLU}&\text{size: }(256,16,16)\\
  &\to\mathrm{Conv}_{512}\to\mathrm{BN}\to\mathrm{ReLU}&\text{size: }(512,8,8)\\
  &\to\mathrm{Conv}_{1024}\to\mathrm{BN}\to\mathrm{ReLU}&\text{size: }(1024,4,4)\\
  &\to\mathrm{Flatten}\to\mathrm{FC}_{64\times 2}.
\end{align*}
The decoder is constructed as
\begin{align*}
  z\in\mathbb{R}^{64}
  &\to\mathrm{FC}_{8\times8\times1024}\\
  &\to\mathrm{ConvT}_{512}\to\mathrm{ReLU}&\text{size: }(512,16,16)\\
  &\to\mathrm{ConvT}_{256}\to\mathrm{BN}\to\mathrm{ReLU}&\text{size: }(256,32,32)\\
  &\to\mathrm{ConvT}_{128}\to\mathrm{BN}\to\mathrm{ReLU}&\text{size: }(128,64,64)\\
  &\to\mathrm{ConvT}_{3}\to\mathrm{Sigmoid}&\text{size: }(3,64,64).
\end{align*}
In all the $\mathrm{Conv}_{k}$ layers and all the $\mathrm{ConvT}_{k}$ layers except for the last, $5\times5$ convolutional filters with stride $(2,2)$ are used. We use ReLU instead of leaky ReLU due to the performance consideration described in the previous subsection.
To fit the size of the input images in our experiment, one extra convolutional layer is added for the encoder and the channel size is twice as large as that in \cite{chen2016infogan},

\vspace{0.0pt}%

\vspace{0.0pt}%
\subsection{\color{crevise}CelebAHQ}
\vspace{0.0pt}%
\label{sec:detail_exp_setup_CelebAHQ}
{\color{crevise}The CelebAHQ images are preprocessed in the similar fashion as in CelebA. The difference is that cropping is skipped and the images are resized from 1024x1024 to 128x128.
We construct the encoder and decoder by following those in the experiment on CelebA as
\begin{align*}
  x\in\mathbb{R}^{64\times64}
  &\to\mathrm{Conv}_{128}\to\mathrm{ReLU}&\text{size: }(128,64,64)\\
  &\to\mathrm{Conv}_{256}\to\mathrm{BN}\to\mathrm{ReLU}&\text{size: }(256,32,32)\\
  &\to\mathrm{Conv}_{512}\to\mathrm{BN}\to\mathrm{ReLU}&\text{size: }(512,16,16)\\
  &\to\mathrm{Conv}_{1024}\to\mathrm{BN}\to\mathrm{ReLU}&\text{size: }(1024,8,8)\\
  &\to\mathrm{Flatten}\to\mathrm{FC}_{256\times 2}.
\end{align*}
and
\begin{align*}
  z\in\mathbb{R}^{256}
  &\to\mathrm{FC}_{16\times16\times1024}\\
  &\to\mathrm{ConvT}_{512}\to\mathrm{ReLU}&\text{size: }(512,32,32)\\
  &\to\mathrm{ConvT}_{256}\to\mathrm{BN}\to\mathrm{ReLU}&\text{size: }(256,64,64)\\
  &\to\mathrm{ConvT}_{128}\to\mathrm{BN}\to\mathrm{ReLU}&\text{size: }(128,128,128)\\
  &\to\mathrm{ConvT}_{3}\to\mathrm{Sigmoid}&\text{size: }(3,128,128),
\end{align*}
where $\mathrm{Conv}_{k}$ and $\mathrm{ConvT}_{k}$ are the same ones as in our experiment on CelebA.}

\vspace{0.0pt}
\section{\color{crevise}Examples of reconstructed and generated images in Section~\ref{sec:experiment_comparison}}
\vspace{0.0pt}
\label{sec:samples}

\begin{figure}[tb]
  \color{crevise}
  \centering
  \subfloat[VAE ($\sigma_{x}^2=1.0$)]{\includegraphics[width=0.250\textwidth]{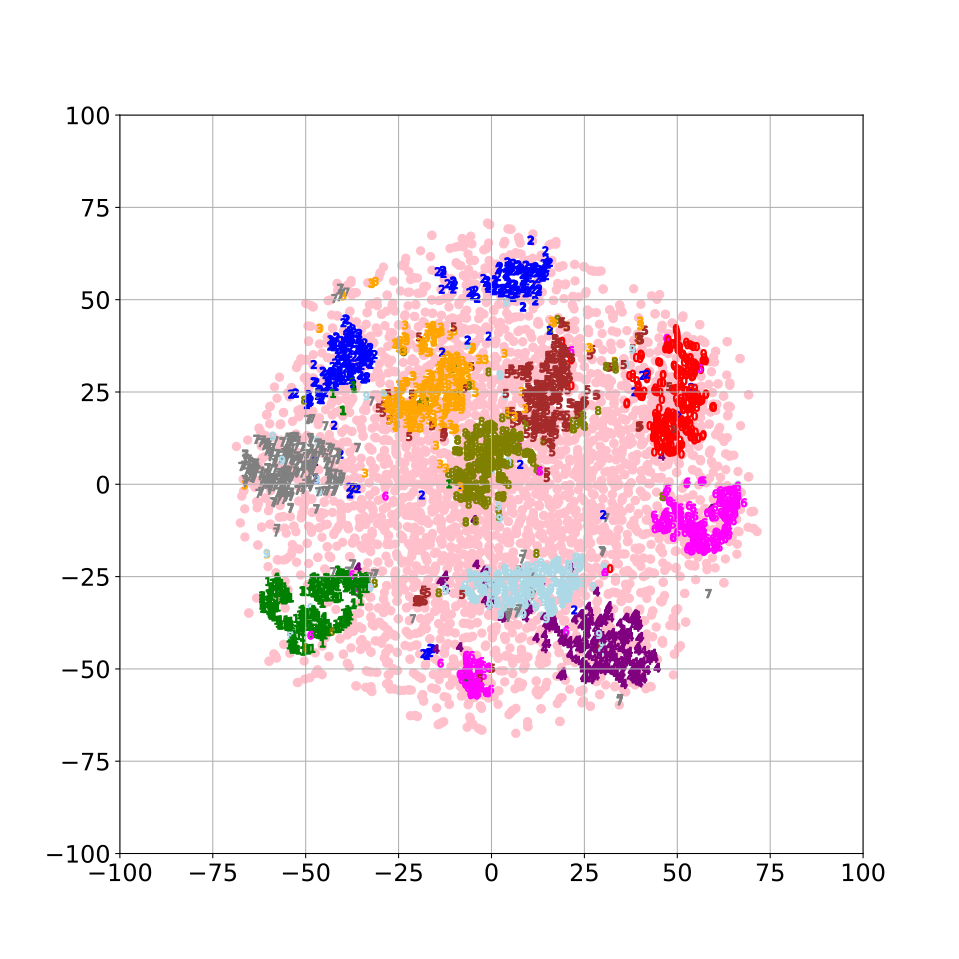}}
  \hspace{20pt}%
  \subfloat[WAE-MMD]{\includegraphics[width=0.250\textwidth]{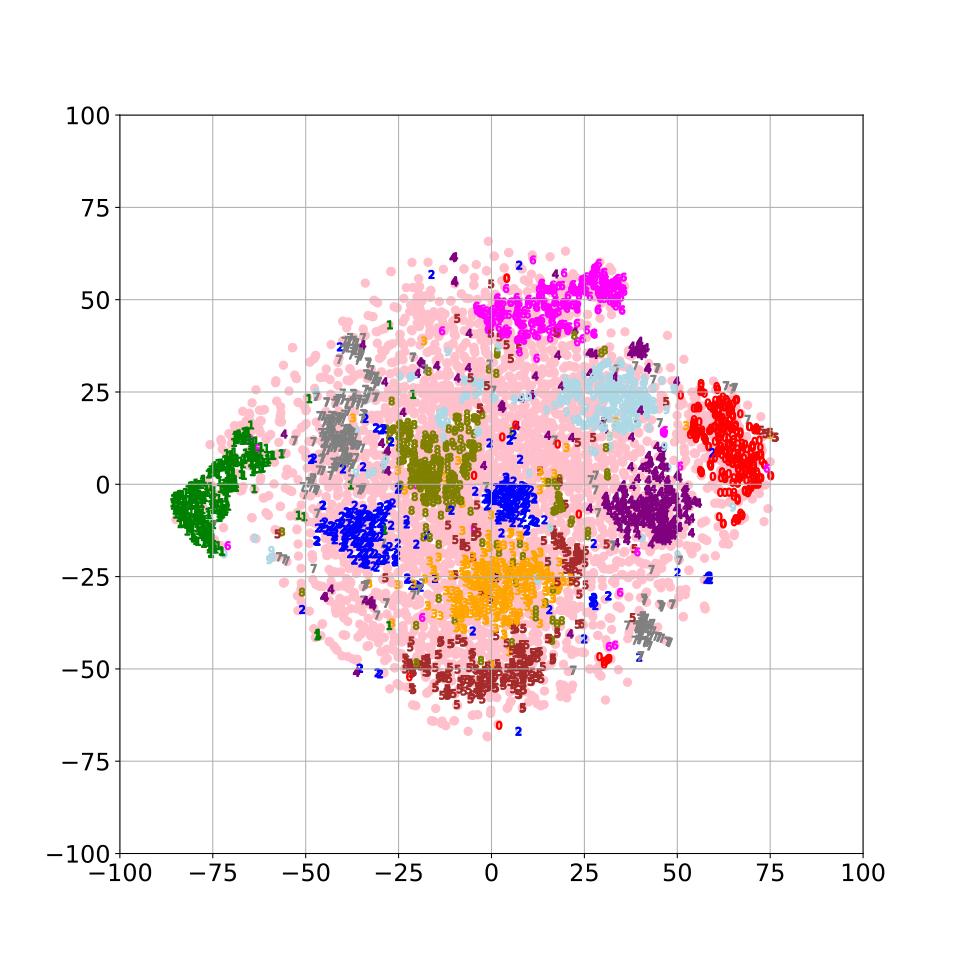}}
  \hspace{20pt}%
  \subfloat[AE]{\includegraphics[width=0.250\textwidth]{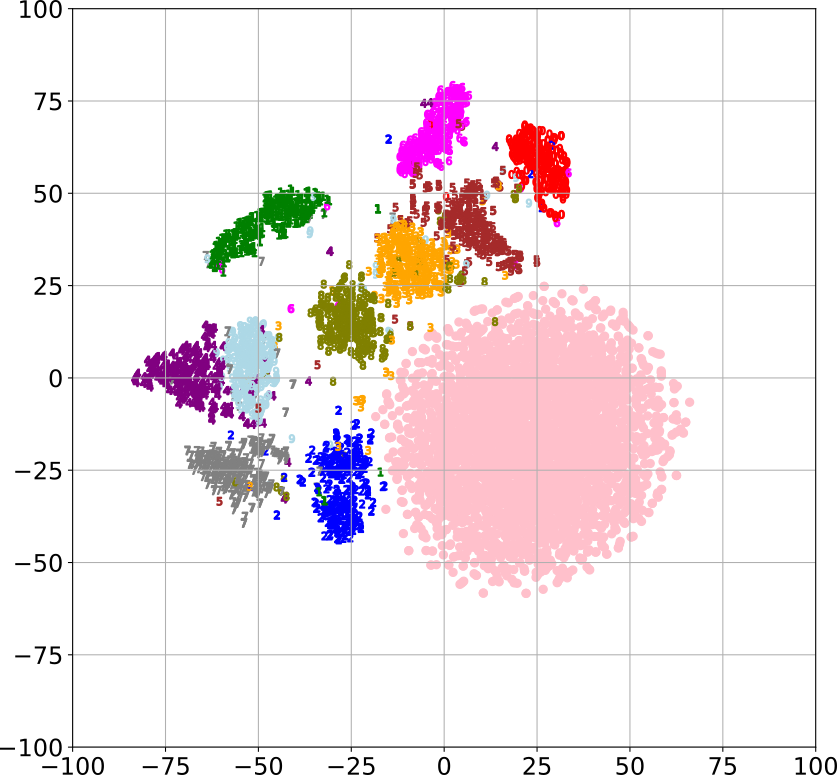}}\\
  \subfloat[RAE]{\includegraphics[width=0.250\textwidth]{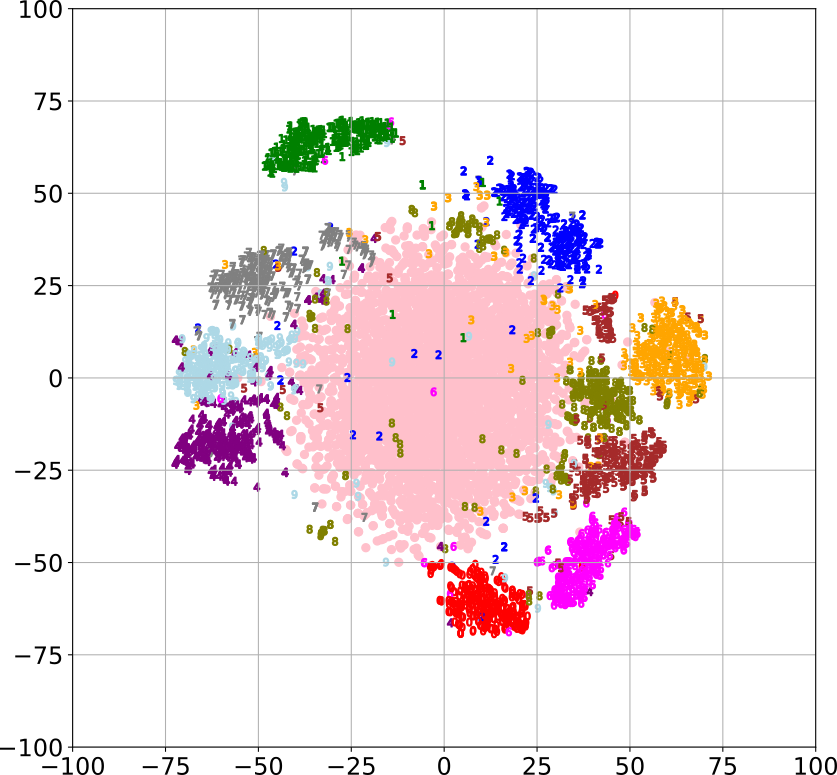}}
  \hspace{20pt}%
  \subfloat[RAE-GP]{\includegraphics[width=0.250\textwidth]{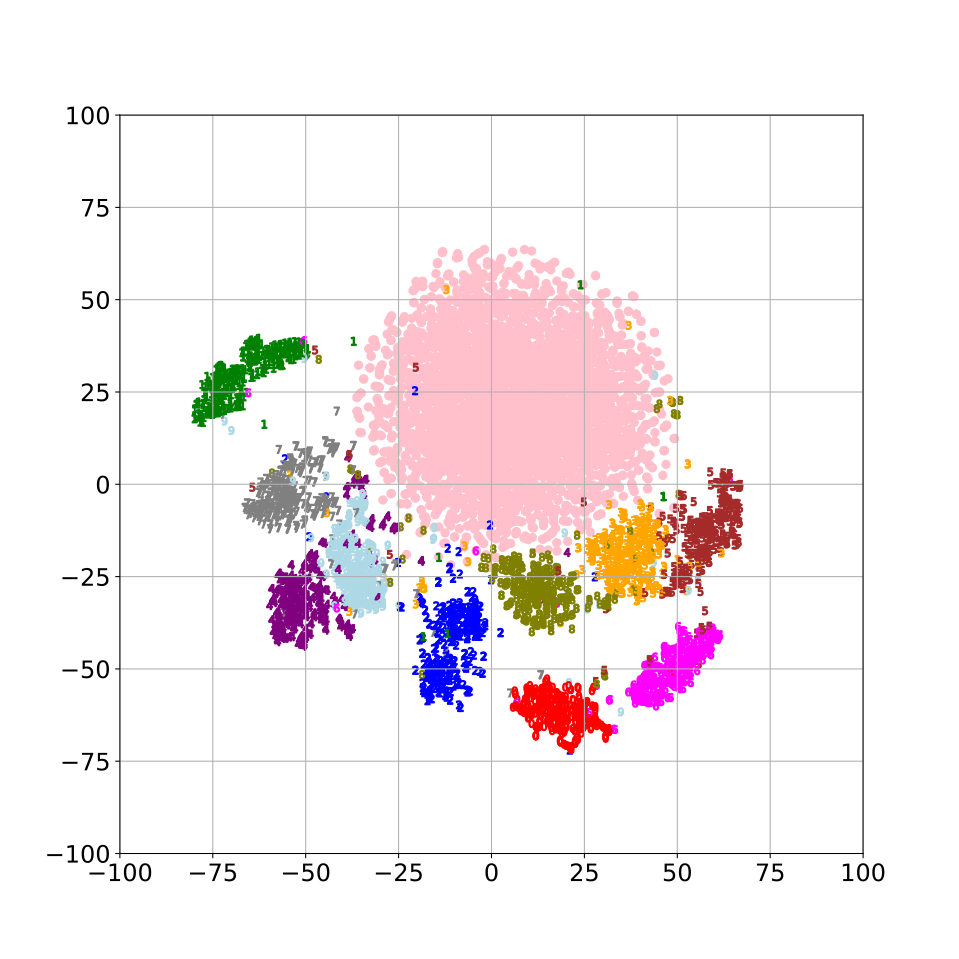}}
  \hspace{20pt}%
\subfloat[Iso-I (Ours)]{\includegraphics[width=0.250\textwidth]{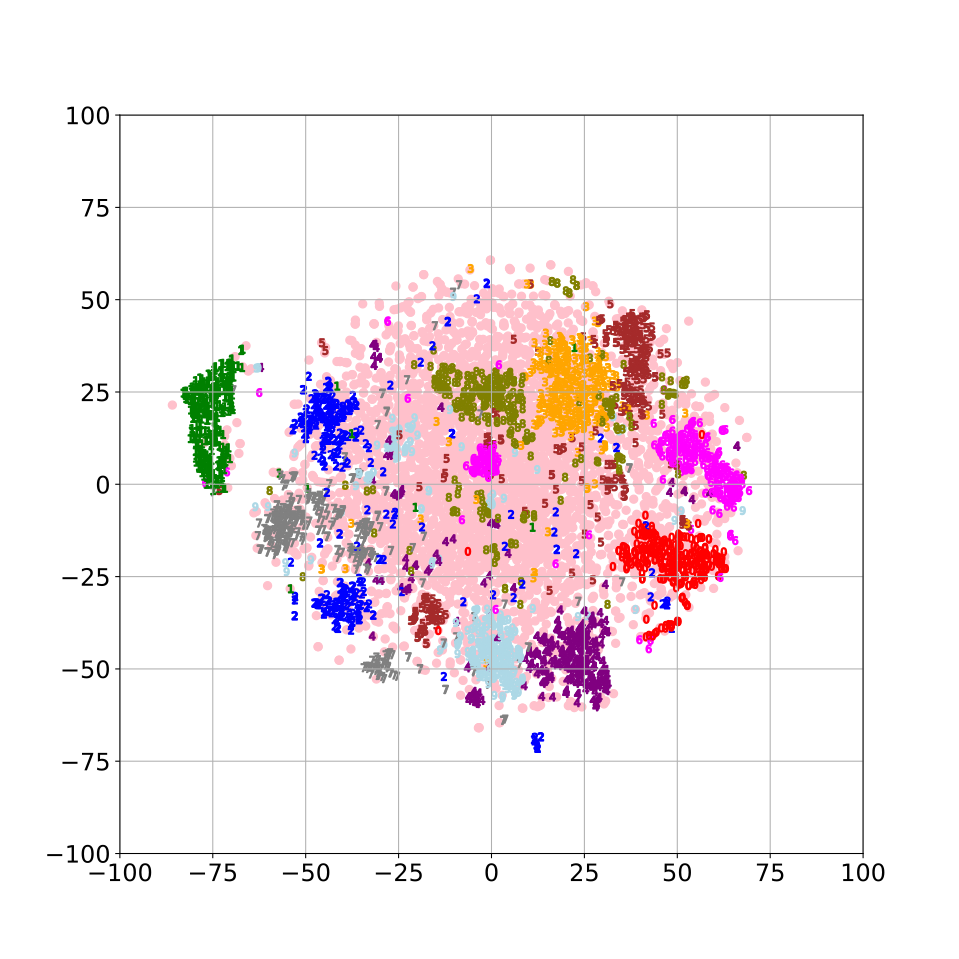}}\\
  \subfloat[Diag-I (Ours)]{\includegraphics[width=0.250\textwidth]{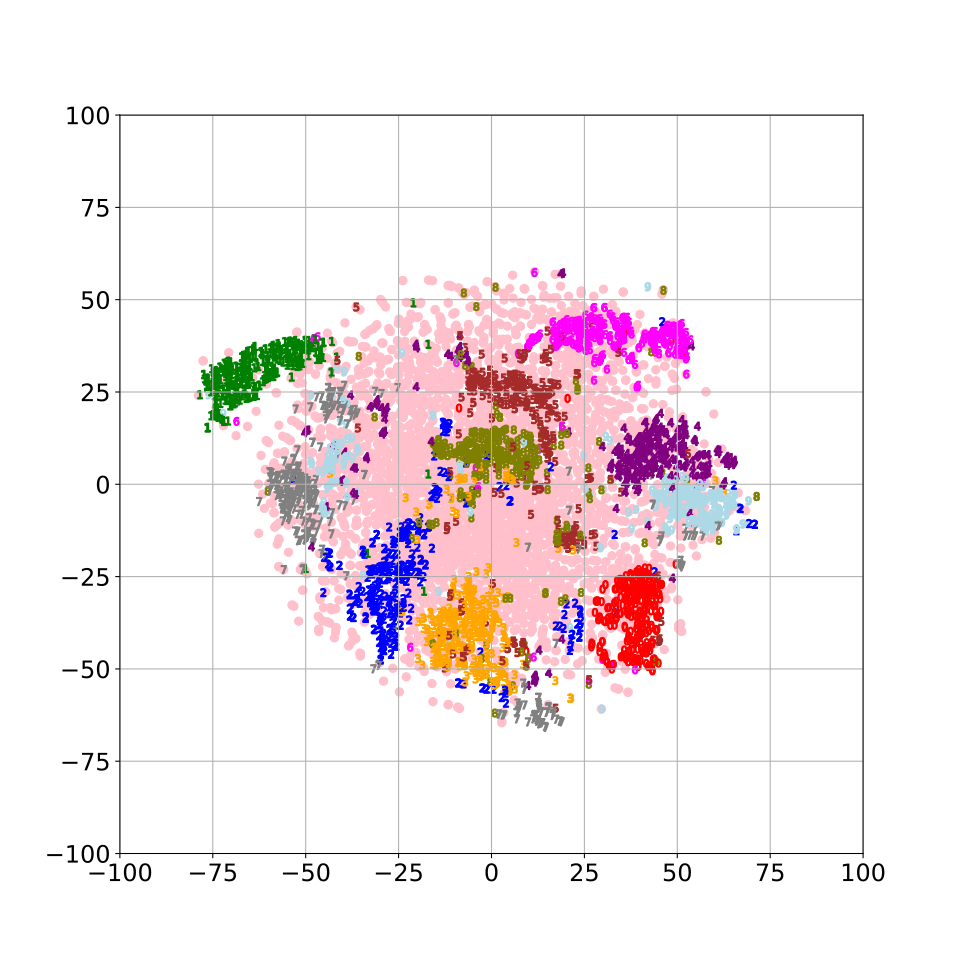}}
  \hspace{20pt}%
  \subfloat[Iso-D (Ours)]{\includegraphics[width=0.250\textwidth]{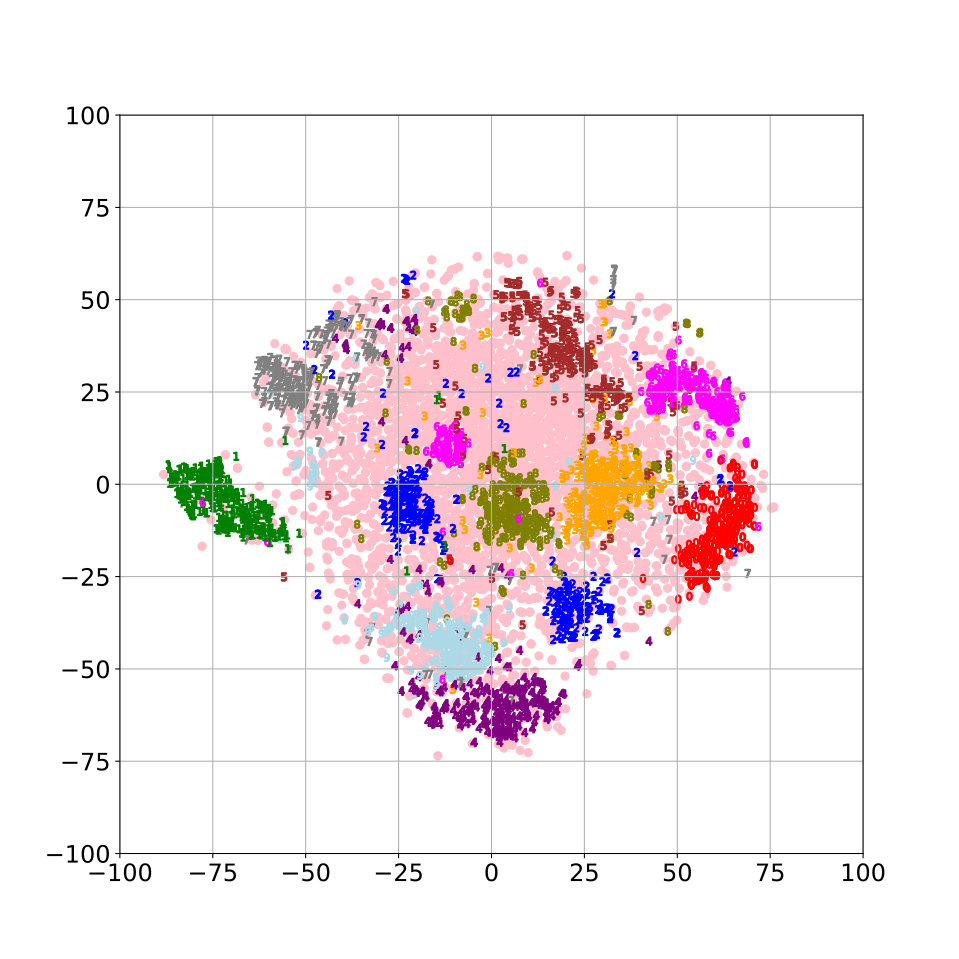}}
  \hspace{20pt}%
  \subfloat[Diag-D (Ours)]{\includegraphics[width=0.250\textwidth]{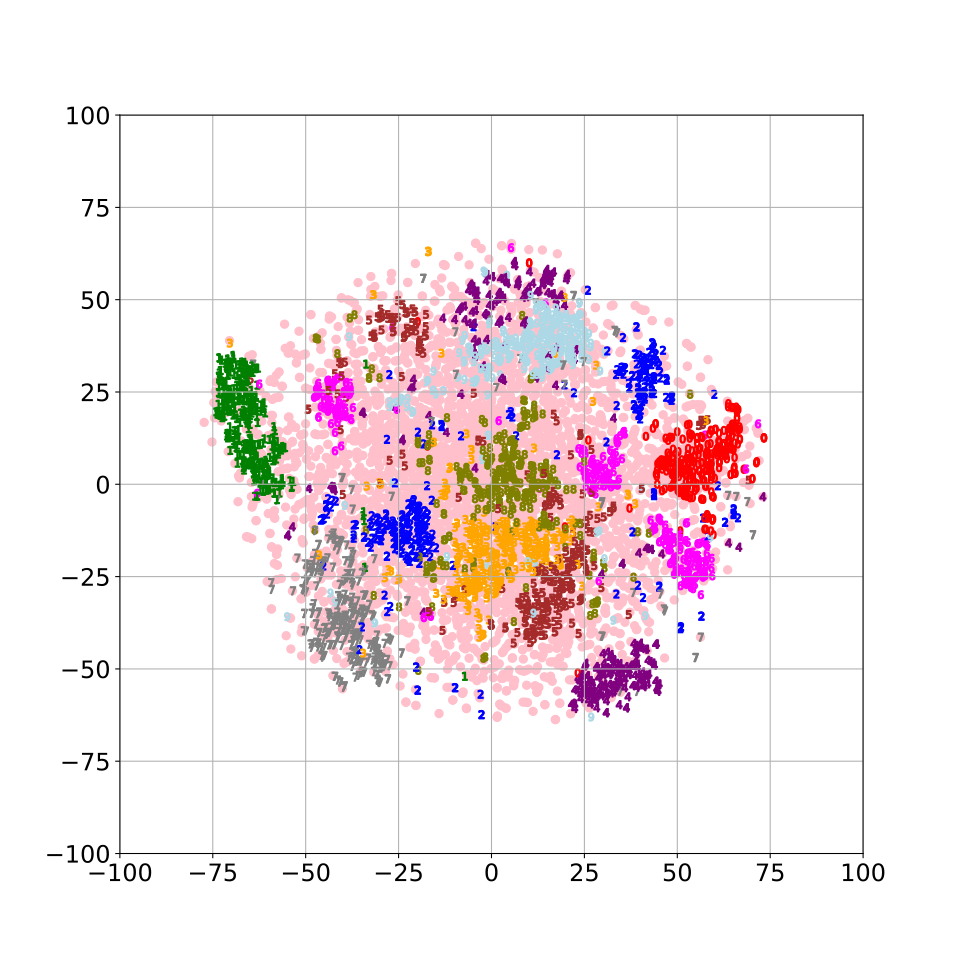}}\\
  \caption{\color{crevise}Visualization of latent space via t-SNE. Pink dots are sampling points generated from the prior $p(\mathbf{z})$.}
  \label{fig:tsne_comp}
\end{figure}

{\color{crevise}We visualize the latent space learned by each method via t-SNE in Figure~\ref{fig:tsne_comp}. The dots with different colors represent the latent vectors encoded from images of different labels (numbers) as in Figures~\ref{fig:sect3_CX-VAE_tsne} and \ref{fig:sect3_CZ-VAE_tsne}.}

\begin{figure}[h!]
  \centering
  \includegraphics[width=0.95\textwidth]{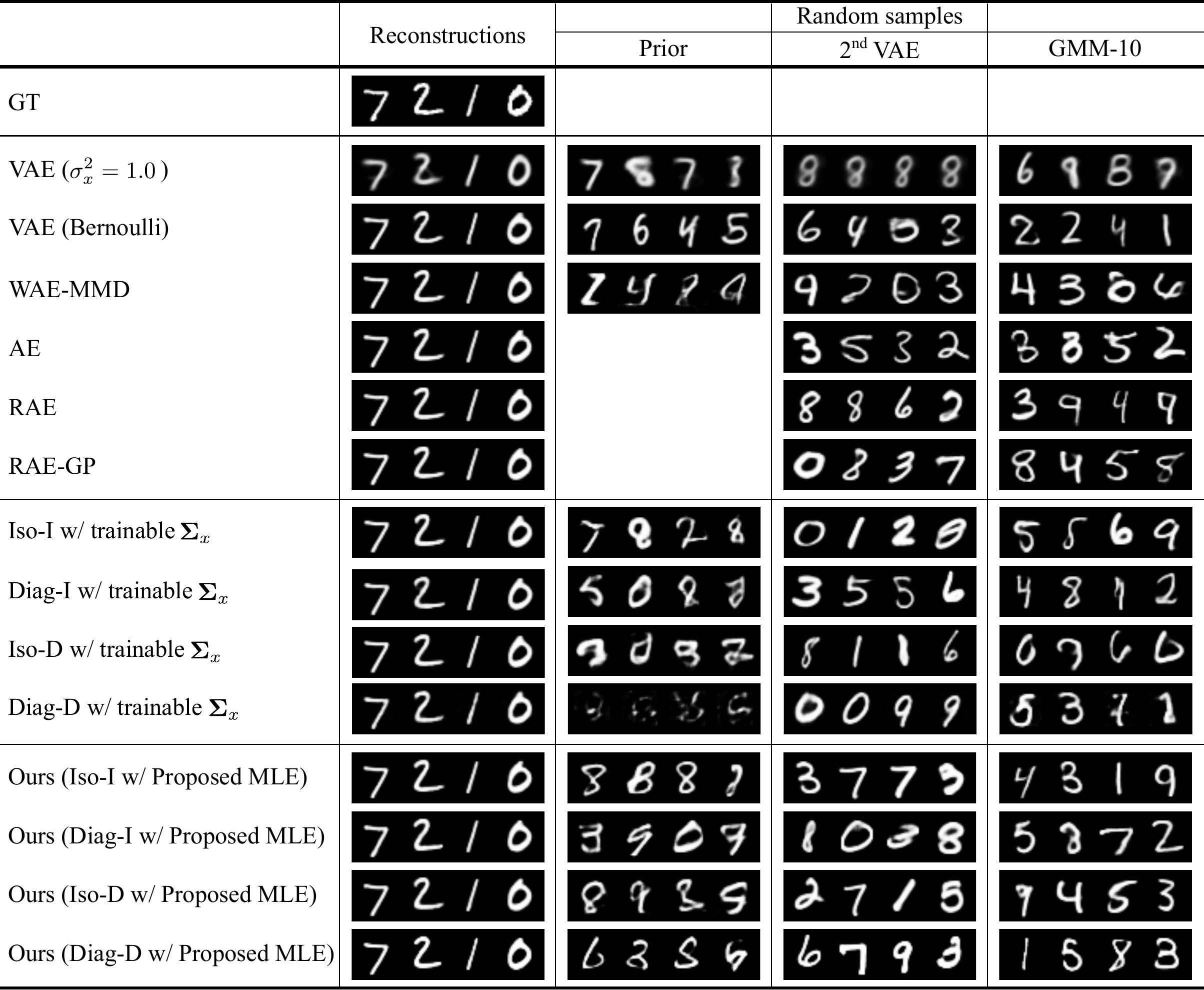}
  \caption{\color{crevise}Reconstructed images and examples of images generated from the prior and the estimated posterior on MNIST. ``GT'' stands for ground truth.}
  \label{fig:samples_mnist}
\end{figure}

\begin{figure}[h!]
  \centering
  \includegraphics[width=0.95\textwidth]{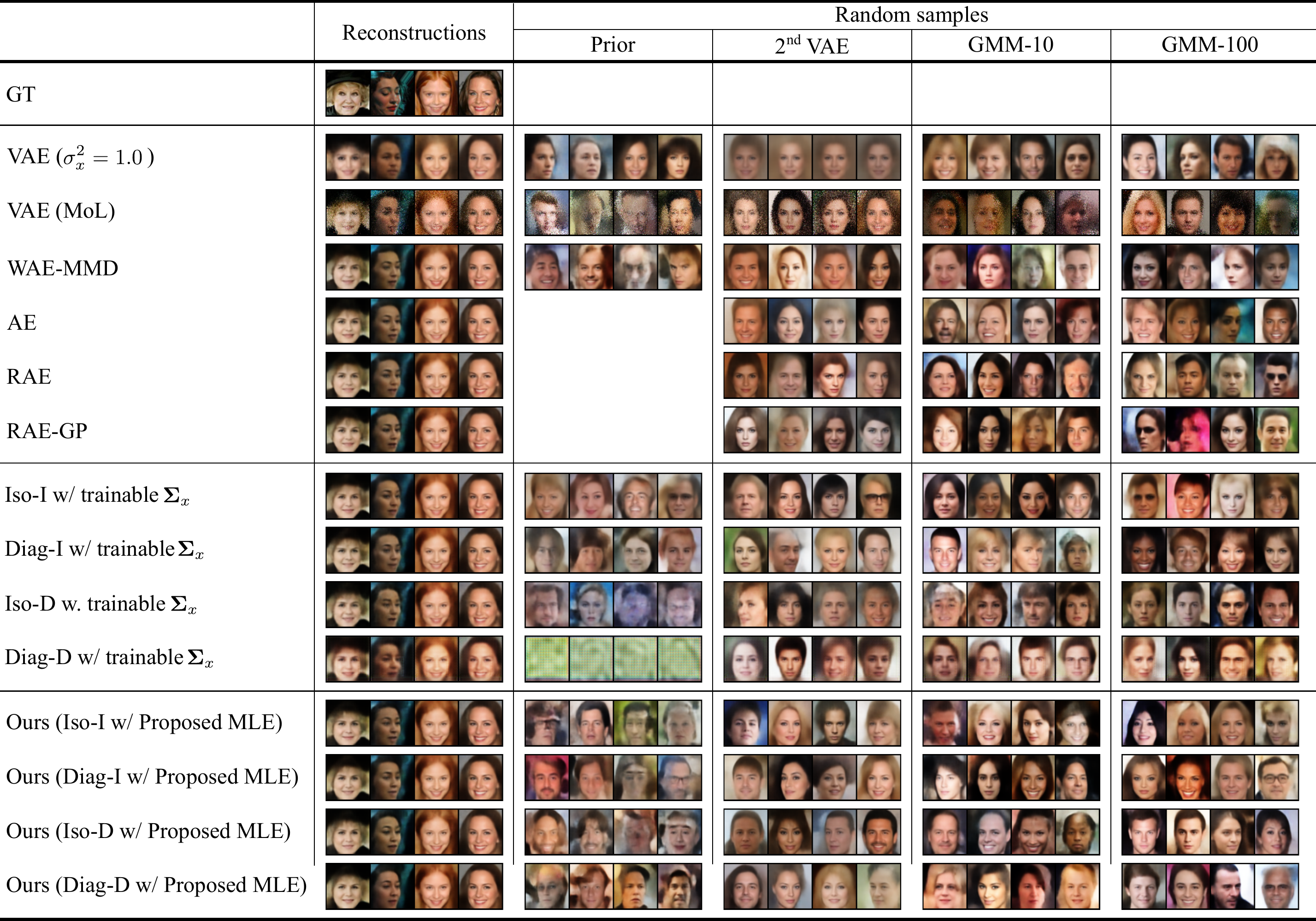}
  \caption{\color{crevise}Reconstructed images and examples of images generated from the prior and the estimated posterior on CelebA.}
  \label{fig:samples_celeba}
\end{figure}

\begin{figure}[h!]
  \centering
  \includegraphics[width=0.95\textwidth]{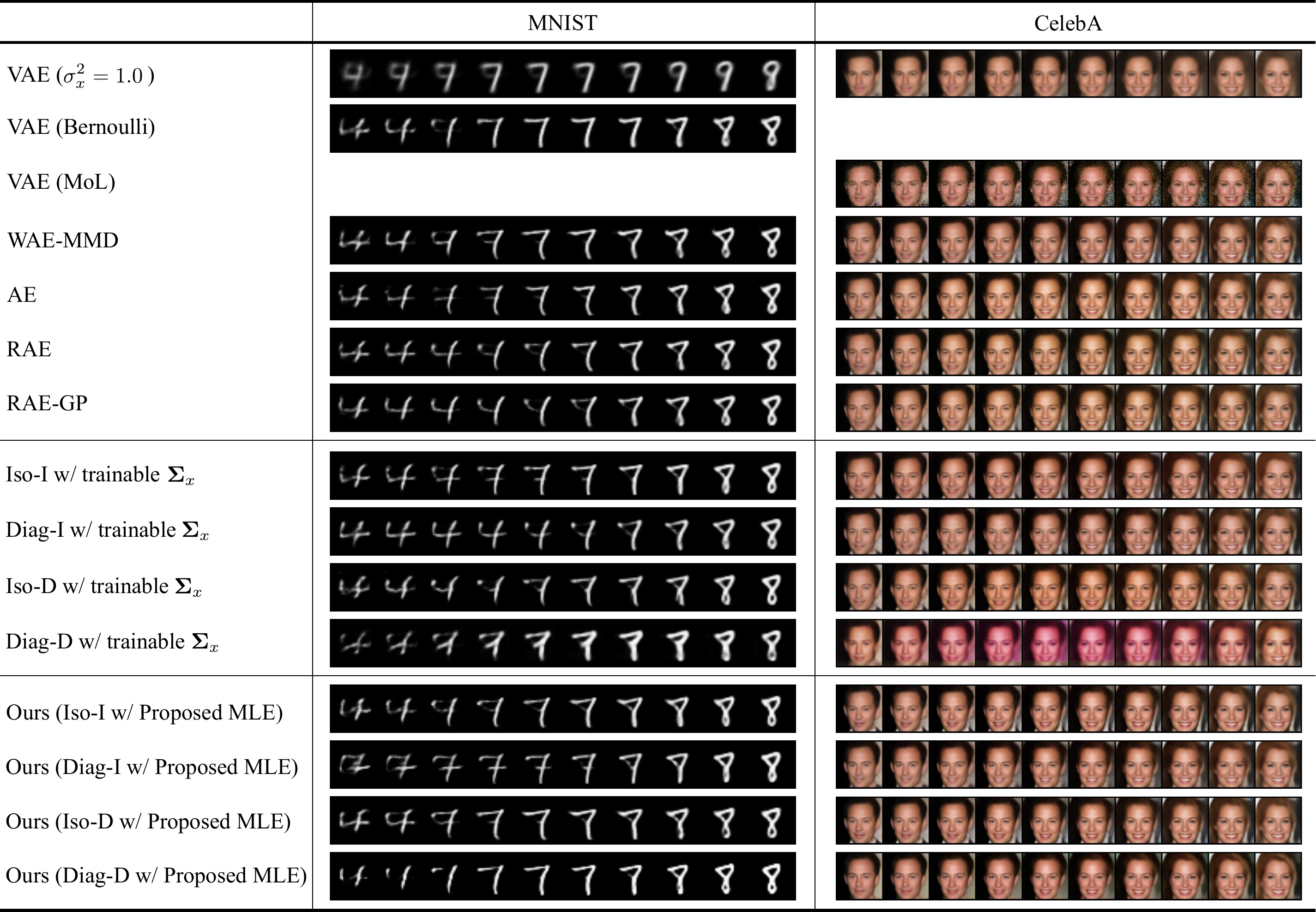}
  \caption{Examples of interpolated images.}
  \label{fig:interpolations}
\end{figure}

\begin{figure}[h!]
  \centering
  \includegraphics[width=0.95\textwidth]{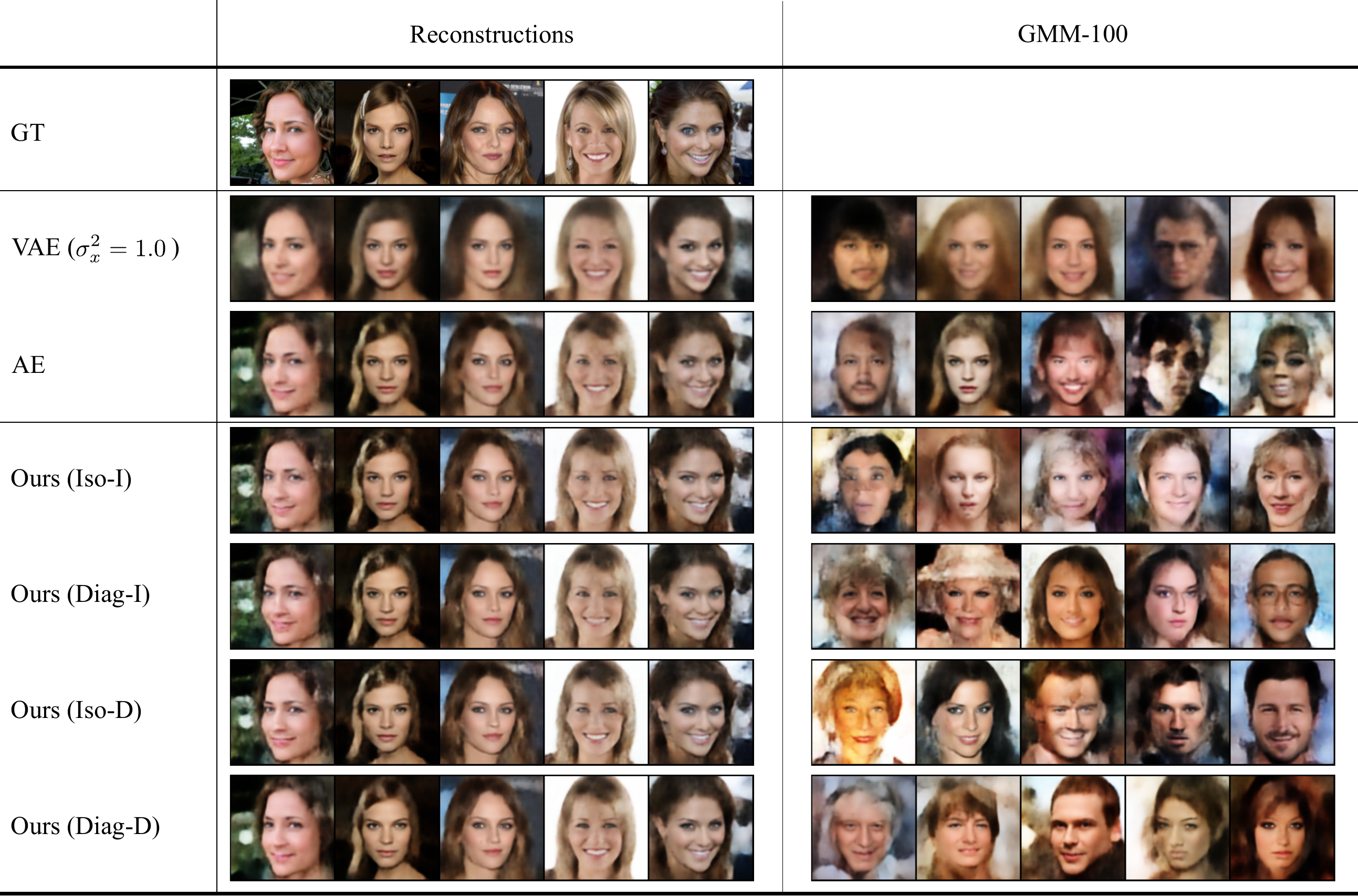}
  \caption{\color{crevise}Reconstructed images and examples of images generated from the prior and the estimated posterior on CelebAHQ.}
  \label{fig:samples_celebahq}
\end{figure}

We show examples of reconstructed images, images generated by sampling the learned approximated posterior and interpolated images from the proposed method and other works in Figures~\ref{fig:samples_mnist}, \ref{fig:samples_celeba}, \ref{fig:interpolations} and {\color{crevise}\ref{fig:samples_celebahq}}.



\clearpage





\end{document}